\documentclass{article}

\usepackage{arxiv}

\usepackage[utf8]{inputenc} 
\usepackage[T1]{fontenc}    
\usepackage{hyperref}       
\usepackage{url}            
\usepackage{booktabs}       
\usepackage{amsfonts}       
\usepackage{nicefrac}       
\usepackage{microtype}      
\usepackage{lipsum}
\usepackage{graphicx}
\graphicspath{ {./images/} }
\usepackage{float}  
\usepackage{enumitem}
\setlist[itemize]{noitemsep, topsep=0pt}
\setlist[enumerate]{noitemsep, topsep=0pt}
\usepackage{float}        
\usepackage{booktabs}     
\usepackage[skip=0pt]{caption} 
\usepackage{titlesec}     

\titlespacing*{\section}{0pt}{0.5ex plus 0.2ex minus 0.2ex}{0.5ex plus 0.2ex}
\titlespacing*{\subsection}{0pt}{0.5ex plus 0.2ex minus 0.2ex}{0.5ex plus 0.2ex}

\usepackage{rotating}  
\usepackage{multirow}  
\usepackage{array}     
\usepackage{graphicx}  
\usepackage{tabularx}  
\usepackage{float}     
\usepackage{changepage} 
\usepackage[numbers]{natbib}
\usepackage{caption}
\usepackage{listings}
\usepackage[table,xcdraw]{xcolor} 
\usepackage{graphicx} 

\title{The Future of MLLM Prompting is Adaptive: A Comprehensive Experimental Evaluation of Prompt Engineering Methods for Robust Multimodal Performance}

\author{
 Anwesha Mohanty, Venkatesh Balavadhani Parthasarathy, and Arsalan Shahid \\ \\
  CeADAR: Ireland's Centre for AI \\
  University College Dublin\\
  Belfield, Dublin 4, Ireland\\
  \texttt{\{anwesha.mohanty, venkatesh.parthasarathy, arsalan.shahid\}@ ucd.ie}
  }
 
\begin{document}
\maketitle

\begin{abstract}
Multimodal Large Language Models (MLLMs) are set to transform how machines process and generate human-like responses by integrating diverse modalities such as text, images, and code. Yet, effectively harnessing their capabilities hinges on optimal prompt engineering. In this study, we present a comprehensive experimental evaluation of seven prompt engineering methods applied to 13 open‐source MLLMs over 24 tasks spanning Reasoning and Compositionality, Multimodal Understanding and Alignment, Complex Code Generation and Execution, and Knowledge Retrieval and Integration. Our approach stratifies models by parameter count into Small (\(<4\)B), Medium (4B–10B), and Large (\(>10\)B) categories and compares prompting techniques including Zero-Shot, One-Shot, Few-Shot, Chain-of-Thought, Analogical, Generated Knowledge, and Tree-of-Thought. Our experiments reveal that while Large MLLMs excel in structured tasks such as code generation and execution, achieving accuracies as high as 96.88\% under Few-Shot prompting. In multimodal understanding and alignment (with relevance scores reaching 100\% using Zero-Shot prompting), all models struggle with complex reasoning and abstract model understanding, often yielding accuracies below 60\% and high hallucination rates. Notably, structured reasoning prompts (Chain-of-Thought, Analogical, Generated Knowledge and Tree-of-Thought) frequently increased hallucination up to 75\% in small models and led to longer response times (exceeding 20 seconds in Large MLLMs), while simpler prompting methods (One-Shot and Few-Shot) provided more concise and efficient outputs. Our findings underscore that no single prompting method uniformly optimizes all task types. Instead, adaptive prompting strategies that combine the strengths of example-based guidance with selective structured reasoning are essential to enhance robustness, efficiency, and factual accuracy in MLLMs. Our work provides critical insights and actionable recommendations for optimizing prompt engineering, paving the way for more reliable deployment of MLLMs in real-world applications ranging from AI-assisted coding and knowledge retrieval to multimodal content understanding.
\end{abstract}


\section{Introduction}

The rapid evolution of Multimodal Large Language Models (MLLMs) has catalyzed a paradigm shift in bridging visual representation learning with natural language understanding, thereby enabling sophisticated multimodal reasoning and broader real-world applicability. Although conventional Large Language Models (LLMs) have demonstrated impressive scaling behaviors \cite{zeng2025scaling}, the leap toward multimodality is primarily driven by the advent of increasingly capable LLM backbones \cite{liu2024llavanext}. Nonetheless, a critical gap persists in the seamless integration of visual processing components with language models. Many current MLLMs employ vision transformer-based architectures, most notably CLIP \cite{radford2021learning, zhai2023sigmoid} as feature extractors, often relegating visual understanding to a secondary role rather than embedding it integrally within the reasoning pipeline. Although alternative approaches, such as self-supervised learning methods exemplified by DINO \cite{oquab2023dinov2}, have shown promise, systematic studies that jointly address architectural design and contextual prompting strategies remain scarce \cite{lu2023reference, liu2023pre}.

MLLMs are inherently designed to process heterogeneous data modalities, thereby expanding their task coverage significantly. However, their effective instruction-following and context comprehension are hindered by suboptimal integration of vision and language modules, inadequate prompt design, and inconsistencies in input representation \cite{liu2023pre, arif2025fixing}. While many evaluations focus on benchmarking MLLM performance across diverse tasks, they frequently overlook critical factors such as:
\begin{itemize}
    \item The compatibility between MLLM architectures and the evaluation dimensions pertinent to specific tasks.
    \item The alignment between training datasets and evaluation benchmarks, ensuring models are optimally calibrated for the tasks they are assessed on.
    \item The effectiveness of prompt engineering techniques in enhancing multimodal understanding and robust instruction-following.
\end{itemize}

To address these shortcomings, our study explores an evaluation-centric and prompt-based framework. We begin by surveying prevalent use cases and delineating the task requirements and evaluation aspects essential for effective multimodal reasoning. This analysis informs our selection of MLLMs, highlighting both architectural diversity and contextual considerations. Subsequently, we rigorously investigate a range of prompt engineering methodologies, examining their impact on enhancing multimodal context integration and overall instruction adherence. In doing so, we propose a comprehensive evaluation framework that elevates multimodal instruction-following as a pivotal performance metric for MLLMs.

\subsection{MLLM Architecture and Applications}

\begin{figure}[ht]
    \centering
    \includegraphics[width=1\textwidth, angle=0]{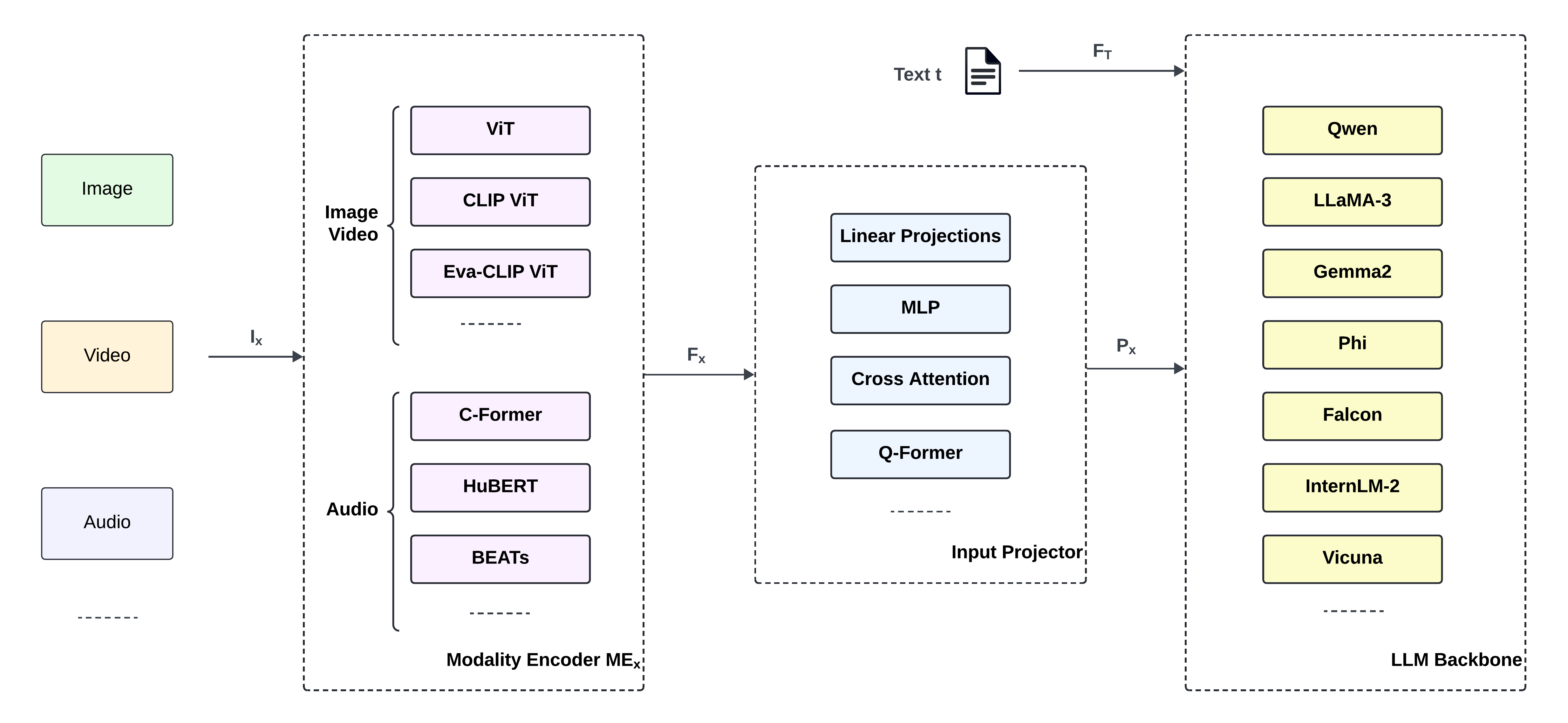}
    \caption{A high-level overview of a typical MLLM pipeline. Multiple input modalities (e.g., images, video, audio) are first processed by dedicated modality encoders (e.g., ViT \cite{dosovitskiy2020image}, CLIP-ViT \cite{radford2021learning}, BEiT \cite{bao2021beit}). The encoded features are then projected or transformed via components such as linear projections, MLPs, or cross-attention to align with the text embedding space. Finally, the LLM backbone (e.g., Qwen, LLaMA, Falcon) integrates these multimodal features for unified reasoning and generation.}
    \label{mllm_architecture}
\end{figure}

Although LLMs are optimized for text-based inputs and outputs, MLLMs extend these capabilities to incorporate images, videos, and audio, necessitating more complex architectural integrations. Typically, an MLLM comprises three primary components: a Modality Encoder, a Transformation Layer, and an LLM backbone, as depicted in Figure \ref{mllm_architecture} \cite{vaswani2023attentionneed, zhang2024mmllmsrecentadvancesmultimodal}.

MLLMs, especially those with the capability to process images and videos, have seen rapid development, as underscored by the HuggingFace VLM Leaderboard \cite{duan2024vlmevalkit}, which now lists more than 200 models introduced since 2022. Notable contributions in this space include the categorization of vision language models (VLMs) by Ghosh et al. \cite{ghosh2024exploringfrontiervisionlanguagemodels} and Yin et al. \cite{Yin_2024}, which analyze architectures, training methodologies, and evaluation metrics.

Specialized applications further highlight the potential of MLLMs. Niu et al. \cite{niu2024textmultimodalityexploringevolution} demonstrate the integration of multimodal data in healthcare for clinical decision-making and medical imaging analysis. Wang et al. \cite{wang2023finvisgptmultimodallargelanguage} introduce FinVis-GPT for financial graph analysis through custom datasets and instruction-based annotations, while Liang et al. \cite{Liang2024.09.29.615524} present DrugChat for the prediction of drug molecule properties. Furthermore, Bewersdorff et al. \cite{Bewersdorff_2025} propose a framework for integrating MLLMs into science education, and Yang et al. \cite{yang2024seedstorymultimodallongstory} develop SEED-Story for generating multimodal narratives using novel attention mechanisms.

A common architectural trend among these models is the use of a vision encoder coupled with an LLM, often linked via a transformation layer \cite{zhang2024mmllmsrecentadvancesmultimodal}. While CLIP-based models excel in zero-shot prompting scenarios \cite{kojima2022large}, they frequently fall short on fine-grained tasks. In contrast, ViT-based architectures \cite{dosovitskiy2020image, wu2020visual, xiao2021early} offer robust spatial attention but incur higher computational costs and reduced interpretability.

\subsection{Evaluation Challenges and Recent Advances}

The HuggingFace VLM Leaderboard \cite{duan2024vlmevalkit} reveals heterogeneous performance across benchmarks; ranging from Vision Q\&A and OCR to RealWorldQA, ChartQA, and MathQA. Notably, leading models such as Step-1o \cite{duan2024vlmevalkit} underperform on benchmarks like HallusionBench \cite{guan2024hallusionbench}, highlighting nuanced trade-offs in current evaluation schemes. Most benchmarks follow a zero-shot evaluation approach with their own set of metrics, typically presenting models with a question and multiple-choice options based on an image. The use of simple and uniform prompts for all samples can lead to an underestimation of model capabilities. The sensitivity of both LLMs and MLLMs to prompt variations, as noted by Xie et al. \cite{xie2024tpevaltapmultimodalllms}, further suggests that uniform prompt designs may fail to capture a model's true potential.

Recent studies have attempted to overcome these limitations through heuristic and ensemble prompting techniques in zero-shot and few-shot settings \cite{brown2020language, sivarajkumar2024empirical}. Enhanced evaluation frameworks, such as those proposed by Hao et al. \cite{hao2025mllmsreasonmultimodalityemma}, indicate that even state-of-the-art models like GPT-o1 \cite{jaech2024openai} struggle to exceed 50\% accuracy in multimodal reasoning tasks despite employing Chain-of-Thought (CoT) prompting \cite{ge2023chain}. Structured CoT (SCoT) approaches have demonstrated improvements of up to 13.79\% over traditional CoT methods \cite{brown2020language, deepseek2025deepseek}.

Complementing these evaluations, Jiang et al. \cite{Jiang_2024} provide a holistic assessment of Large Vision-Language Models (LVLMs) across both specialized tasks (e.g., object detection and medical diagnosis) and general tasks (e.g., object counting and spatial reasoning). Their evaluation, which includes models such as MiniGPT-v2 \cite{chen2023minigpt}, LLaVA-1.5 \cite{liu2024improved}, and Shikra \cite{chen2023shikra}, along with assessments via GPT-4V \cite{OpenAI2025Models} highlights ongoing challenges including limited cognition, object hallucination, and robustness issues. Additionally, research by Li et al. \cite{li2025structured} explores Structured CoT for improving code generation by incorporating programming principles (sequential, branch, and loop) to guide reasoning steps before coding; however, code generation remains underexplored within the MLLM context.

Different models exhibit varying sensitivities to the same prompt changes, and existing evaluation frameworks often fail to address prompt-induced bias, leading to unfair comparisons. These architectural limitations, inconsistencies in reported benchmark accuracies, and current evaluation methodologies raise significant questions about the true potential of these models.

\subsection{Our Approach and Contributions}

To address these challenges, we developed a comprehensive experimental framework to rigorously evaluate 13 open-source MLLMs across four key aspects: Multimodal Reasoning, Model Understanding, Knowledge Retrieval, and Code Generation. The selected models represent a stratified sample based on parameter sizes and are paired with diverse LLM backbones to facilitate a robust evaluation of both performance and real-world applicability. The main contributions of this paper include: 

\begin{enumerate}
    \item A comparative study of seven prompt engineering methods applied to 13 open-source MLLMs, evaluating their performance on 24 tasks across four evaluation aspects. In doing so, we design a diverse set of tasks, create standardized prompt templates, and share both the datasets and templates to ensure reproducibility.
    \item An in-depth analysis of how prompt engineering strategies interact with task types and model scales, providing insights into model performance, reliability, and resource requirements, as well as offering best practices and actionable recommendations for optimizing performance in real-world applications.
\end{enumerate}

\section{Methods}

To understand the impact of prompt engineering techniques across diverse tasks and evaluation metrics, we employed a four-staged experimental design and evaluation framework including:

\begin{itemize}
    \item \textbf{Stage 1 – Defining Core Evaluation Aspects:} We define four Evaluation Aspects (EAs) to provide a comprehensive analysis of model performance across reasoning, multimodal interpretation, code generation, and knowledge integration. Detailed discussion of these aspects is provided in Section~\ref{modelEAs}. For each EA, we curated a set of tasks designed to challenge the models’ ability to integrate and process multimodal inputs (primarily images) across real-world scenarios. The corresponding tasks are listed in Tables~\ref{tab:ea1_tasks}, \ref{tab:ea2_tasks}, \ref{tab:ea3_tasks}, and \ref{tab:ea4_tasks}.
    
    \item \textbf{Stage 2 – Review and Selection of MLLMs:} A diverse set of 13 open-source MLLMs were chosen based on their architecture, parameter size, and availability (as discussed in Section~\ref{modelselection}). These models showcase the breadth of current MLLM capabilities by combining different image and text models, each built upon a distinct text encoder.
    
    \item \textbf{Stage 3 – Selection of Prompt Engineering Methods:} We apply seven prompting methods including Zero-Shot, One-Shot, Few-Shot, Chain-of-Thought, Analogical, Generated Knowledge, and Tree-of-Thought prompting. Section~\ref{PromptEnggTechs} provides further discussions on selected methods.
    
    \item \textbf{Stage 4 – Evaluation Framework:} Our experimental setup was designed to ensure consistency and reliability across multiple tasks and prompting techniques. Model outputs were evaluated along two primary dimensions. The first dimension focuses on task performance, using four key metrics: accuracy, relevancy, conciseness, and hallucination (detailed in Section~\ref{manualevaluationcriteria}). The second dimension assesses resource consumption, including inference time and memory consumption. A comprehensive manual review process is devised to analyse MLLM outputs to assess aforementioned metrics.
\end{itemize}

\subsection{Stage 1: Defining Core Evaluation Aspects} \label{modelEAs}

Evaluating MLLMs is crucial to understanding their capabilities, limitations, and applicability across diverse domains. Prior studies \cite{li2024llava, wu2025controlmllm, yu2023mmvet} have explored various evaluation aspects, from perceptual understanding and compositional reasoning to multimodal alignment and task-specific problem solving. Unlike traditional unimodal models, MLLMs require frameworks that assess their ability to process, integrate, and generate outputs from multiple modalities (e.g., text, images, and in some cases audio and video). Reviews such as \cite{nwae403} typically categorize evaluations into general multimodal understanding, task-specific assessments, and trustworthiness metrics, ensuring that models transition from broad reasoning to reliable, specialized real-world applications. Motivated by these insights and based on our review of current applications and evaluations in MLLM landscape (see Appendix~\ref{rationale-for-selected-EAs}), we select four core Evaluation Aspects (EAs) for their broad impact and practical relevance:

\begin{enumerate}
    \item \textbf{Reasoning and Compositionality (EA1):} This aspect centers on the model’s ability to process textual and visual cues, perform logical deductions, and synthesizing disparate information into coherent outputs. It tests capabilities such as Multi-Step Problem Solving, Pattern Recognition, Logical Deduction, and Compositional Synthesis (see Table~\ref{tab:ea1_tasks}).
    \item \textbf{Multimodal Understanding and Alignment (EA2):} Evaluates how well the model aligns, integrates, and interprets information across modalities. It is critical for tasks that demand accurate cross-referencing between text and visual data. Key capabilities to evaluate include cross-modal referencing and alignment image interpretation and description, and consistency in visual–textual context (see Table~\ref{tab:ea2_tasks}).
    \item \textbf{Complex Code Generation and Execution (EA3):} This aspect evaluates the model’s ability to interpret instructions, extract data from visual inputs, and generate executable code. It is essential for programming tasks where code must be both syntactically correct and logically coherent. The key evaluation capabilities include data extraction from visual input, programmatic transformation and logic construction, and accurate code synthesis, debugging, and explanation (see Table~\ref{tab:ea3_tasks}).
    \item \textbf{Knowledge Retrieval and Integration (EA4):} Focuses on the model’s ability to recall, verify, and merge factual information from textual and visual sources into coherent responses. This is critical for domains such as research, journalism, medicine, and data science. The key capabilities to evaluate include factual recall and verification, domain-specific context and explanation, and cross-modal knowledge synthesis (see Table~\ref{tab:ea4_tasks}).
\end{enumerate}

For each EA, we curated tasks to challenge models in realistic, application-oriented scenarios. Task objectives, and key challenges are summarized in the following tables and design rationales are elaborated in Appendices~\ref{apdx_ea1}, \ref{apdx_ea2}, \ref{apdx_ea3}, and \ref{apdx_ea4}. Full task descriptions and expected outputs are available in the supplementary material \ref{supp_EA1} to \ref{supp_EA4}.

\begin{table}[ht]
\caption{Overview of Evaluation Aspect 1 (EA1): Reasoning and Compositionality, comprising four tasks (T1 to T4) focused on visual pattern recognition, logical deduction, mathematical reasoning, and narrative synthesis.}
\label{tab:ea1_tasks}
\begin{minipage}{\textwidth}
\renewcommand{\arraystretch}{1.0}
\setlength{\tabcolsep}{8pt}
\begin{tabularx}{\textwidth}{p{3.0cm} X X}
\toprule
\textbf{Task(s)} & \textbf{Objective(s)} & \textbf{Key Challenges}\\
\midrule
\textbf{EA1\_T1:}\\
Pattern Recognition in Visual Sequences 
&
Test the model’s ability to detect and generalise patterns in a sequence of related images or diagrams.
&
Identifying logical/visual patterns and extrapolating rules from limited data.
\\
\midrule
\textbf{EA1\_T2:}\\
Logical Deduction from Text and Simplified Diagram
&
Evaluate the model’s capacity to interpret textual instructions alongside a simple diagram to reach a correct conclusion.
&
Integrating textual clues with diagrammatic cues and ensuring consistency in multi-step reasoning.
\\
\midrule
\textbf{EA1\_T3:}\\
Mathematical Puzzle with Visual Data
&
Assess how the model handles numeric computations and interprets simple visual representations (e.g., shapes or charts).
&
Bridging quantitative reasoning with visual elements and avoiding arithmetic errors.
\\
\midrule
\textbf{EA1\_T4:}\\
Story Synthesis from Text and Image
&
Check the model’s ability to create a coherent narrative by merging textual descriptions and a relevant image.
&
Maintaining logical flow in narrative form and blending visual context with text.
\\
\bottomrule
\end{tabularx}
\end{minipage}
\end{table}

\begin{table}[ht]
\caption{Overview of Evaluation Aspect 2 (EA2): Multimodal Understanding and Alignment, including four tasks (T1 to T4) that test the integration and interpretation of information across text, images, and charts.}
\label{tab:ea2_tasks}
\begin{minipage}{\textwidth}
\renewcommand{\arraystretch}{1.0}
\setlength{\tabcolsep}{8pt}
\begin{tabularx}{\textwidth}{p{3.0cm} X X}
\toprule
\textbf{Task(s)} & \textbf{Objective(s)} & \textbf{Key Challenge(s)}\\
\midrule
\textbf{EA2\_T1:}\\
Image-Text Matching and Explanation 
&
Verify the model’s capacity to match an image with a corresponding text description and explain the match.
&
Correctly identifying key features and ensuring textual alignment with visual elements.
\\
\midrule
\textbf{EA2\_T2:}\\
Inferring Context from Combined Modalities
&
Determine how the model integrates separate text and image inputs to deduce higher-level context.
&
Seamlessly fusing diverse information sources and handling ambiguous or incomplete data.
\\
\midrule
\textbf{EA2\_T3:}\\
Cross-Modal Translation
&
Evaluate the model’s ability to translate visual information (e.g., symbols or icons) into meaningful text.
&
Accurately handling symbolic representations and capturing fine-grained details.
\\
\midrule
\textbf{EA2\_T4:}\\
Aligning Data from Charts and Text
&
Assess how well the model interprets and aligns quantitative data from a chart with textual analysis.
&
Avoiding misinterpretation of graphical data and integrating numerical details with text.
\\
\bottomrule
\end{tabularx}
\end{minipage}
\end{table}

\begin{table}[ht]
\caption{Overview of Evaluation Aspect 3 (EA3): Complex Code Generation and Execution, comprising eight tasks (T1 to T8) that evaluate a model’s ability to generate executable code from visual and textual inputs across a range of structured reasoning and programming challenges.}
\label{tab:ea3_tasks}
\begin{minipage}{\textwidth}
\renewcommand{\arraystretch}{1.0}
\setlength{\tabcolsep}{8pt}
\begin{tabularx}{\textwidth}{p{3.0cm} X X}
\toprule
\textbf{Task(s)} & \textbf{Objective(s)} & \textbf{Key Challenges}\\
\midrule
\textbf{EA3\_T1:}\\
Data Visualization from an Image of a Table
&
Generate a script that converts table data in an image into a visualization (e.g., bar chart).
&
Handling OCR-like interpretation, mapping image data to structured format, and producing valid code.
\\
\midrule
\textbf{EA3\_T2:}\\
Drawing a Shape Based on an Image
&
Produce code that programmatically draws a shape using visual hints from an image.
&
Translating visual references into geometric coordinates and ensuring syntax correctness.
\\
\midrule
\textbf{EA3\_T3:}\\
Calculating a Sum from Text in an Image
&
Write a function to parse textual or numeric data from an image and compute a sum.
&
Accurate text extraction, handling parsing errors, and verifying arithmetic accuracy.
\\
\midrule
\textbf{EA3\_T4:}\\
Creating a Dictionary from an Image of a Chart
&
Convert labels and values in a chart image into a dictionary or key-value structure.
&
Extracting structured data, ensuring correct type conversion, and code clarity.
\\
\midrule
\textbf{EA3\_T5:}\\
Summing Prices from a Shopping List Image
&
Generate code to read item prices from an image of a shopping list and calculate the total.
&
Handling varied text formats, summing accurately, and managing currency symbols or decimals.
\\
\midrule
\textbf{EA3\_T6:}\\
Parsing a Simple CSV Structure from an Image
&
Build a script that interprets an image containing CSV-like text and converts it into a data table.
&
Accurate extraction of rows/columns and handling formatting inconsistencies.
\\
\midrule
\textbf{EA3\_T7:}\\
Generating Fibonacci Sequence Based on Image Instruction
&
Produce code to generate a Fibonacci sequence following instructions specified in an image.
&
Interpreting visual instructions accurately and ensuring logical code correctness.
\\
\midrule
\textbf{EA3\_T8:}\\
Responding to a Flowchart Image
&
Interpret a flowchart diagram and output code or logic to implement the described process.
&
Translating flowchart nodes into algorithmic steps and ensuring overall coherence.
\\
\bottomrule
\end{tabularx}
\end{minipage}
\end{table}

\begin{table}[ht]
\caption{Overview of Evaluation Aspect 4 (EA4): Knowledge Retrieval and Integration, comprising eight tasks (T1 to T8) that assess how effectively a model combines visual cues and textual context to retrieve, interpret, and explain domain-specific knowledge.}
\label{tab:ea4_tasks}
\begin{minipage}{\textwidth}
\renewcommand{\arraystretch}{1.0}
\setlength{\tabcolsep}{8pt}
\begin{tabularx}{\textwidth}{p{3.0cm} X X}
\toprule
\textbf{Task(s)} & \textbf{Objective(s)} & \textbf{Key Challenges}\\
\midrule
\textbf{EA4\_T1:}\\
Historical Monument Identification and Explanation
&
Identify a famous monument from an image and provide its historical context.
&
Handling historical facts, distinguishing similar monuments, and accurately explaining cultural significance.
\\
\midrule
\textbf{EA4\_T2:}\\
Scientific Data Interpretation from Graph and Text
&
Integrate textual and graphical data to answer a scientific question and summarize key findings.
&
Extracting relevant data from graphs, merging with textual context, and ensuring scientific accuracy.
\\
\midrule
\textbf{EA4\_T3:}\\
Medical Image Analysis with Knowledge Integration
&
Provide a brief medical interpretation from an image (e.g., an X-ray) along with a textual description of symptoms.
&
Understanding medical terminology, ensuring factual accuracy, and integrating visual and textual cues.
\\
\midrule
\textbf{EA4\_T4:}\\
Cultural Artifact Interpretation
&
Identify an artifact from an image and explain its cultural/historical background using textual clues.
&
Combining historical and cultural knowledge accurately and presenting a coherent explanation.
\\
\midrule
\textbf{EA4\_T5:}\\
Integrating Knowledge from a Map and Text Description
&
Combine visual map data with textual instructions to address location-based or geographical queries.
&
Accurately interpreting map symbols and reconciling textual directions with visual references.
\\
\midrule
\textbf{EA4\_T6:}\\
Integrating Information from a Chart and Article
&
Merge insights from a chart (e.g., population growth) with an accompanying article to produce a synthesized summary.
&
Ensuring correct numerical interpretation, linking data points to textual arguments, and forming a comprehensive summary.
\\
\midrule
\textbf{EA4\_T7:}\\
Multimodal Fact Checking
&
Verify the factual accuracy of a statement by cross-referencing an image (e.g., a photograph) with textual sources.
&
Cross-validating visual evidence with textual claims and identifying potential inconsistencies.
\\
\midrule
\textbf{EA4\_T8:}\\
Integrating Visual Art and Historical Context
&
Explain an artwork shown in an image, including its historical context and cultural details.
&
Recognizing artistic styles, contextualizing the piece historically, and referencing relevant artistic movements.
\\
\bottomrule
\end{tabularx}
\end{minipage}
\end{table}

\subsection{Stage 2: Review and Selection of MLLMs} \label{modelselection}

This stage consists of two parts. First, we review key models current MLLM landscape, encompassing both proprietary and open‐source developments. Second, we detail the selection criteria and present the 13 open‐source MLLMs chosen for our evaluation.

Proprietary models such as OpenAI’s GPT-4o, GPT-4.5 Preview \cite{OpenAI2025Models}, Anthropic’s Claude 3 \cite{anthropic2024claude}, and Google’s Gemini series \cite{DeepMind2025Gemini} demonstrate strong multimodal reasoning, particularly in instruction-following and contextual understanding. For instance, GPT-4 introduced multimodal inputs to process both text and images, while Claude 3 (including its variants Claude 3 Opus, Sonnet, and Haiku) has been optimized for text–image reasoning tasks. Despite their robust performance, the closed-source nature of these models limits customization, transparency, and independent research. In contrast, the open-source ecosystem has rapidly evolved, providing competitive alternatives with enhanced accessibility and community-driven improvements. Models such as Meta’s LLaMA series \cite{meta2024llama3}, Mistral \cite{mistral2024pixtral}, Falcon \cite{malartic2024falcon2}, and DeepSeek \cite{deepseek2025deepseek} offer greater flexibility in fine-tuning and deployment. Nonetheless, open-source MLLMs still face challenges in matching proprietary models’ instruction-following, contextual learning, and multimodal alignment. A comprehensive discussion of the various distinct MLLMs including their architectural designs, multimodal processing capabilities, and evaluation suitability is provided in Appendix~\ref{apdx_review_mllm}.

After conducting an in-depth review of various models, several factors influenced the exclusion of specific models from our selection:
\begin{itemize}
    \item Modality Focus – Models explicitly designed for video, audio, or long-sequence temporal reasoning (e.g., VITA \cite{fu2024vita}, Long-VITA \cite{shen2025long}, mPLUG-Owl3 \cite{ye2024mplug}) were not included, as our evaluation focuses on text-image multimodal reasoning rather than temporal or multi-frame processing.
    \item Task-Specific Specialization – Some models, such as MoAI (OCR-centric) \cite{lee2024moai}, ChatRex (object detection) \cite{jiang2024chatrex}, and ViP-LLaVA (region-aware multimodal interaction) \cite{cai2024vip}, are highly optimized for domain-specific tasks rather than general-purpose multimodal reasoning, making them less suitable for a structured comparative analysis.
    \item Reliance on External Modules – Models like Molmo \cite{deitke2024molmo} and Cambrian-1 \cite{tong2025cambrian}, which incorporate external retrieval mechanisms or specialized visual processing modules, introduce dependencies that complicate direct performance comparisons across standardized benchmarks.
    \item Performance-Compute Tradeoff – While Falcon2-11B \cite{malartic2024falcon2} and MiniCPM \cite{yao2024minicpm} are highly efficient, they are optimized for lightweight multimodal interactions rather than advanced vision-language compositionality, making them less aligned with our focus on deeper reasoning and complex multimodal integration.
    \item Maturity and Benchmarking Limitations – Some recently released models, such as Meteor \cite{lee2025meteor} and Cambrian-1 \cite{tong2025cambrian} lack comprehensive benchmarking on widely used multimodal datasets, making it difficult to systematically compare their performance against well-established counterparts.
\end{itemize}

Taking these factors into account, we structured our model selection process to ensure a balanced, computationally feasible, and diverse evaluation of open-source MLLMs. 

We selected 13 open-source MLLMs for detailed evaluation, focusing on models that offer a balanced combination of scalability, architectural diversity, and multimodal reasoning capabilities. Our selection criteria include:
\begin{itemize}
    \item Parameter Scale and Architectural Diversity - Models are categorized as Small ($<$4B), Medium (4B–10B), and Large ($>$10B) to capture variations in computational demands and performance.
    \item Multimodal Integration - Preference was given to models integrating different vision encoders (e.g., ViT, SigLIP, CLIP, EVA) with various language models (e.g., Qwen, Gemma, Llama, Phi), enabling an exploration of how vision–language pairings influence task performance.
    \item Open-Source Availability - Only models with fully open-source code and weights were considered to ensure reproducibility and community accessibility.
    \item Practical Feasibility - Due to GPU constraints, models exceeding 15B parameters were generally excluded from direct evaluation. However, we incorporated one model above 15B parameters using quantization techniques to examine the trade-offs between model size and inference efficiency.
\end{itemize}

For each category, four models were selected (with five in the large model group due to the inclusion of a quantized model). This strategy ensures a fair comparison across small, medium, and large models while addressing key questions related to parameter scaling, vision–language integration, and computational efficiency.

Table~\ref{tab:model_comparison} summarizes the final list of models along with key specifications such as parameter count, release date, underlying language and vision models, and image input support.

\begin{table}[ht]
\caption{List of Models Finalized for Evaluation. This table includes the model name, parameter size, release date, and details of the language and vision model combination.}
\label{tab:model_comparison}
\centering
\resizebox{\textwidth}{!}{%
\begin{tabularx}{\textwidth}{
  >{\raggedright\arraybackslash}m{0.22\textwidth}  
  >{\centering\arraybackslash}m{0.14\textwidth}    
  >{\centering\arraybackslash}m{0.14\textwidth}    
  >{\raggedright\arraybackslash}m{0.23\textwidth}  
  >{\raggedright\arraybackslash}m{0.23\textwidth}  
}
\toprule
\textbf{Model} & 
\textbf{Params (B)} & 
\textbf{Release Date} & 
\textbf{Language Model} & 
\textbf{Vision Model} \\
\midrule
InternVL-2 & 1 & 8-Jul-24 & Qwen2.5-0.5B & InternViT-300M \\
\midrule
Qwen2-VL & 2 & 29-Aug-24 & Qwen2-1.5B & ViT-600M \\
\midrule
MiniMonkey & 2.2 & 9-Aug-24 & InternLM2-1.8B & InternViT-300M \\
\midrule
Paligemma-3B-mix-448 & 3 & 14-May-24 & Gemma-2B & SigLIP-400M \\
\midrule
Phi-3.5 VLM & 4 & 21-May-24 & Phi-3.5 & CLIP ViT-L/14 \\
\midrule
LLaVA OneVision-7B & 8 & 14-Sep-24 & Qwen2-7B & SigLIP-400M \\
\midrule
Ovis 1.5-Llama 3-8B & 8 & 17-Jun-24 & Llama-3-8B-Instruct & SigLIP-400M \\
\midrule
GLM-4v-9B & 9 & 30-Jul-24 & GLM-4-9B & EVA-02-5B \\
\midrule
Ovis-1.6 & 10.2 & 17-Jun-24 & Gemma2-9B-lt & SigLIP-400M \\
\midrule
Llama3.2-Vision & 11 & 25-Sep-24 & Llama 3.1 & ViT \\
\midrule
Pixtral & 12 & 17-Sep-24 & Nemo-12B & ViT-400M \\
\midrule
OmChat V2 & 13 & 6-Jul-24 & Qwen2-7B & InternViT-6B \\
\midrule
InternVL-2 & 26 & 8-Jul-24 & InternLM2-20B & InternViT-6B \\
\bottomrule
\end{tabularx}%
}
\end{table}

\subsection{Stage 3: Selection of Prompt Engineering Methods} 
\label{PromptEnggTechs}

Understanding multimodal content requires deep multimodal knowledge and models must not only grasp information within each modality but also accurately infer how these modalities interact to support effective reasoning \cite{yang2023mm}. Prompt engineering has emerged as a straightforward and efficient method for guiding LLMs and MLLMs, enabling enhanced performance in complex reasoning tasks. This approach generally involves two types of prompts: instruction-based and example-based \cite{bhattacharjya2024foundation}. Instruction-based prompts include system-level prompts that establish overarching guidelines and task-specific prompts tailored to particular objectives, while example-based prompts rely on a few illustrative examples to define desired input-output relationships.

Numerous studies \cite{jiang2022promptmaker, zamfirescu2023johnny} have identified two key challenges in prompting: crafting effective prompts and evaluating their efficacy. In particular, example-based approaches have proven effective in guiding large models \cite{mann2020language, wei2022chain, yao2024tree}. Although many interactive systems support prompt engineering, most focus predominantly on textual or limited visual inputs. This narrow focus overlooks the intricate interactions between modalities, thereby limiting the development of prompts that fully leverage the contextual richness of multimodal inputs to enhance reasoning \cite{zamfirescu2023johnny}.

Models can quickly adapt to new downstream tasks in few-shot or even zero-shot settings without requiring retraining \cite{liu2023pre}. As exemplified by the pioneering Chain-of-Thought (CoT) prompting technique, which prompts LLMs to generate intermediate reasoning steps, mirroring human cognitive processes \cite{wei2022chain, kojima2022large, zhang2022automatic}; the concept has been extended to the multimodal domain (M-CoT) in several studies \cite{rose2023visual, zhang2023multimodal, ge2023chain}. Analogical Reasoning Prompting leverages shared structural similarities between scenarios to unlock a model’s analogical reasoning abilities \cite{yasunaga2023large}, while Generated Knowledge Prompting encourages models to generate additional background knowledge to enhance reasoning \cite{liu2021generated, liu2023pre}. Tree-of-Thought (ToT) prompting further extends CoT by structuring reasoning into a decision tree that explores multiple pathways before converging on a solution \cite{yao2024tree}.

To effectively guide MLLMs across diverse tasks, we adopt and implement seven distinct prompting techniques:

\begin{enumerate}
    \item Zero-Shot Prompting \cite{radford2019language}
    \item One-Shot Prompting \cite{mann2020language}
    \item Few-Shot Prompting \cite{mann2020language}
    \item Chain-of-Thought (CoT) Prompting \cite{wei2022chain}
    \item Analogical Prompting \cite{yasunaga2023large}
    \item Generated Knowledge Prompting \cite{liu2021generated}
    \item Tree-of-Thought (ToT) Prompting \cite{yao2024tree}
\end{enumerate}

For detailed reviews, prompt templates, and usage scenarios for each prompting technique, please refer to Appendix~\ref{apdx_prompt_methods}. By systematically designing and refining these prompts, our approach aims to generate consistent, rationale-driven outputs across a wide range of multimodal tasks.

\subsection{Stage 4: Evaluation Framework} \label{manualevaluationcriteria}

Our evaluation framework is designed to rigorously assess model outputs across multiple tasks and prompting techniques. All experiments were performed on high-performance Nvidia GPUs (see Appendix~\ref{experimental_setup}). To ensure consistency, inference parameters, including temperature, maximum token length, and decoding strategies were held constant across all models.
We assessed model outputs along two primary dimensions: task performance and resource consumption. Task performance metrics include accuracy, relevancy, conciseness, and hallucination. Accuracy captures whether the response correctly addresses all components of the task. Relevancy assesses how well the response aligns with the task’s context and objectives. Conciseness evaluates the clarity and brevity of the response. Hallucination measures the extent to which responses include irrelevant, redundant, or fabricated content.
Resource consumption metrics include inference time and memory usage, recorded to evaluate model efficiency.

Table~\ref{tab:evaluation_thresholds} summarizes the empirical thresholds established for these metrics. Table~\ref{tab:detailed_evalcriteria} supplement these empirical thresholds to ensure consistent assessment of model performance.

\begin{table}[ht]
    \caption{Empirical thresholds for evaluation metrics based on industry benchmarks and prior research \cite{DeepMind2025Gemini, jiangmmad, huang2023language, zhang2024mme, adler2024gpt, meta2024llama32}. The Accuracy metric determines if all task elements are correctly addressed, while Relevancy ensures that responses remain contextually aligned. Conciseness evaluates clarity and brevity, and Hallucination flags irrelevant or repetitive content.}
    \label{tab:evaluation_thresholds}
    \centering
    \renewcommand{\arraystretch}{1.0}
    \setlength{\tabcolsep}{12pt}
    \begin{tabular}{|l|c|}
        \hline
        \textbf{Metric} & \textbf{Threshold} \\
        \hline
        Accuracy       & $\geq 80\%$ \\
        Hallucination  & $< 5\%$ \\
        Relevancy      & $\geq 90\%$ \\
        Conciseness    & $\geq 80\%$ direct, $< 10\%$ under-explained \\
        \hline
    \end{tabular}
\end{table}


\subsubsection{Inter-Annotator Agreement and Evaluation Consistency}
To ensure objectivity, two independent annotators evaluated model outputs using the criteria in Table~\ref{tab:evaluation_thresholds}. Annotator A assessed EA1 and EA4, while Annotator B evaluated EA2 and EA3. A cross-review process was implemented in which each annotator reviewed the other's scores to identify discrepancies, which were resolved via consensus. A structured guideline document was used to standardize interpretation of each metric.

\subsubsection{Empirical Basis for Evaluation Metrics}
Our evaluation metrics are quantified to objectively assess the performance of MLLMs. The 80\% accuracy threshold is based on evidence that state-of-the-art multimodal models often struggle to exceed 75\% on standard benchmarks \cite{jiangmmad}, and higher accuracy is critical in high-stakes applications \cite{huang2023language, zhang2024mme}. Similarly, a hallucination rate below 5\% is essential for maintaining factual grounding, as confirmed by recent studies \cite{vectara_hallucination_leaderboard, galileo_hallucination_index}. Relevancy and conciseness thresholds ensure outputs are both meaningful and precise.

This structured framework provides a robust foundation for comparing prompt engineering techniques across MLLMs by balancing qualitative and quantitative aspects of model outputs.

\section{Results}
\label{sec:results}

This section presents a comprehensive evaluation of model performance across multiple evaluation aspects (EAs) using a diverse set of tasks. The evaluation encompasses seven aforementioned prompting techniques and key performance indicators such as Accuracy, Hallucination control, Response Relevance, Irrelevance, and Conciseness. For Conciseness, results are divided into Under-Explained (UE) and a combined measure of Target Precision and Over-Explained (TP + OE).

Models are categorized based on their parameter count as follows:
\begin{itemize}
    \item Small MLLMs (\(<4\)B parameters),
    \item Medium MLLMs (4B--10B parameters),
    \item Large MLLMs (\(>10\)B parameters).
\end{itemize}

\subsection{Summary of Average Model Performance Across Evaluation Aspects}

Tables~\ref{tab:reasoning_results_summary}--\ref{tab:knowledge_retrieval_results_summary} report the average performance (in \%) for each model category (Small, Medium, and Large MLLMs) under the seven prompting techniques across the four evaluation aspects.

In EA1 (Reasoning and Compositionality) tasks, which assess a model’s ability to perform multi-step problem solving and integrate information across modalities, Table~\ref{tab:reasoning_results_summary} highlights several key findings.
Few-shot prompting yields the highest accuracy for large multimodal language models (MLLMs), achieving 45\% and outperforming both Chain-of-Thought and Tree-of-Thought prompting strategies. In contrast, small MLLMs exhibit notably higher hallucination rates, reaching up to 75\% when using Tree-of-Thought, while medium and large models demonstrate substantially lower hallucination levels. Relevance scores remain consistently high for large MLLMs, with performance exceeding 90\%, whereas small models tend to lag in aligning their responses with task context. Finally, in terms of conciseness, small MLLMs are more likely to produce under-explained or verbose responses, in contrast to medium and large models, which maintain clearer and more concise outputs.

In EA2 (Multimodal Understanding and Alignment) tasks, which assess how well the model aligns, integrates, and interprets information across modalities, Table~\ref{tab:model_understanding_tasks_summary_results} summarizes the performance on four tasks across three model sizes of MLLMs. Large and medium MLLMs achieve near-perfect relevance, close to 100\% when using Zero-Shot, One-Shot, and Few-Shot prompting techniques. In contrast, small MLLMs demonstrate higher hallucination rates, particularly when using Tree-of-Thought prompting. While accuracy across all model sizes remains moderate, small MLLMs tend to score lower on average. Conciseness metrics further reveal that small models often produce shorter, under-explained responses, whereas medium and large models provide more complete and detailed outputs.

For complex code generation and execution tasks under EA3, the results summarized in Table~\ref{tab:code_generation_tasks_summary_results} indicate that large MLLMs achieve the highest accuracy, reaching up to 96.88\% with Few-Shot prompting. Hallucination levels are nearly zero in medium and large MLLMs across several prompting methods, whereas small MLLMs exhibit considerably higher hallucination. Relevance scores remain uniformly high, approaching 100\% for medium and large MLLMs. In terms of response quality, small MLLMs tend to produce under-explained outputs, while medium and large models generate more balanced and detailed code responses.

In knowledge retrieval and integration tasks (EA4), Table~\ref{tab:knowledge_retrieval_results_summary} shows that large MLLMs achieve the highest accuracy, up to 87.5\%, along with near-perfect relevance (close to 100\%), particularly when using Zero-Shot prompting. In contrast, small MLLMs display higher hallucination rates, exceeding 40\% in some cases, while medium MLLMs fall between the two in terms of performance. Additionally, medium MLLMs are more likely to produce under-explained outputs compared to both small and large categories.

\begin{table}[ht]
\centering
\caption{
EA1: Reasoning and Compositionality Tasks Results Summary. This table presents the average performance (in \%) of Small (S-MLLMs, \(<4\)B), Medium (M-MLLMs, 4B--10B), and Large (L-MLLMs, \(>10\)B) models on reasoning tasks. Performance metrics include Accuracy, Hallucination, Relevance (Fully and Partially Relevant), Irrelevance, Conciseness (Under Explained - UE, Over Explained - OE). Abbreviations: ZS = Zero-Shot, OS = One-Shot, FS = Few-Shot, CoT = Chain-of-Thought, Anl = Analogical, GK = Generated Knowledge, ToT = Tree-of-Thought.
}
\label{tab:reasoning_results_summary}

\renewcommand{\arraystretch}{1.1}
\setlength{\tabcolsep}{4pt}
\begin{tabular}{|p{3.5cm}|p{1.7cm}|p{1.2cm}|p{1.2cm}|p{1.2cm}|p{1.2cm}|p{1.2cm}|p{1.2cm}|p{1.2cm}|}
\hline
\textbf{Metrics} & \textbf{Size} & ZS & OS & FS & CoT & Anl & GK & ToT \\ \hline
Accuracy 
  & S-MLLMs & 31.25 & 18.75 & 31.25 & 18.75 & 25 & 6.25 & 6.25 \\ \cline{2-9}
  & M-MLLMs & 31.25 & 37.5 & 31.25 & 18.75 & 37.5 & 25 & 18.75 \\ \cline{2-9}
  & L-MLLMs & 30 & 30 & 45 & 25 & 25 & 20 & 35 \\ \hline
Hallucination 
  & S-MLLMs & 37.5 & 50 & 37.5 & 31.25 & 37.5 & 50 & 75 \\ \cline{2-9}
  & M-MLLMs & 6.25 & 12.5 & 0 & 12.5 & 0 & 12.5 & 43.75 \\ \cline{2-9}
  & L-MLLMs & 0 & 0 & 5 & 15 & 5 & 15 & 30 \\ \hline
Relevance (F + P)
  & S-MLLMs & 75 & 56.25 & 62.5 & 75 & 68.75 & 56.25 & 56.25 \\ \cline{2-9}
  & M-MLLMs & 93.75 & 87.5 & 81.25 & 100 & 100 & 100 & 75 \\ \cline{2-9}
  & L-MLLMs & 95 & 90 & 90 & 90 & 95 & 90 & 95 \\ \hline
Irrelevance 
  & S-MLLMs & 25 & 43.75 & 37.5 & 25 & 31.25 & 43.75 & 43.75 \\ \cline{2-9}
  & M-MLLMs & 6.25 & 12.5 & 18.75 & 0 & 0 & 0 & 25 \\ \cline{2-9}
  & L-MLLMs & 5 & 10 & 10 & 10 & 5 & 10 & 5 \\ \hline
Conciseness (UE)
  & S-MLLMs & 25 & 12.5 & 18.75 & 25 & 31.25 & 6.25 & 0 \\ \cline{2-9}
  & M-MLLMs & 50 & 50 & 43.75 & 43.75 & 62.5 & 43.75 & 43.75 \\ \cline{2-9}
  & L-MLLMs & 50 & 75 & 70 & 50 & 60 & 45 & 50 \\ \hline
Conciseness (TP + OE)
  & S-MLLMs & 75 & 87.5 & 81.25 & 75 & 68.75 & 93.75 & 100 \\ \cline{2-9}
  & M-MLLMs & 50 & 50 & 56.25 & 56.25 & 37.5 & 56.25 & 56.25 \\ \cline{2-9}
  & L-MLLMs & 50 & 25 & 30 & 50 & 40 & 55 & 50 \\ \hline
\end{tabular}
\end{table}


\begin{table}[ht]
\centering
\caption{
EA2: Multimodal Understanding and Alignment Tasks Results Summary. This table displays the average performance (in \%) of Small (S-MLLMs, \(<4\)B), Medium (M-MLLMs, 4B--10B), and Large (L-MLLMs, \(>10\)B) models on tasks requiring multimodal understanding. Performance metrics include Accuracy, Hallucination, Relevance (Fully and Partially Relevant), Irrelevance, Conciseness (Under Explained - UE, Over Explained - OE). Abbreviations: ZS = Zero-Shot, OS = One-Shot, FS = Few-Shot, CoT = Chain-of-Thought, Anl = Analogical, GK = Generated Knowledge, ToT = Tree-of-Thought.
\label{tab:model_understanding_tasks_summary_results}
}
\renewcommand{\arraystretch}{1.1}
\setlength{\tabcolsep}{4pt}
\begin{tabular}{|p{3.5cm}|p{1.7cm}|p{1.2cm}|p{1.2cm}|p{1.2cm}|p{1.2cm}|p{1.2cm}|p{1.2cm}|p{1.2cm}|}
\hline
\textbf{Metrics} & \textbf{Size} & ZS & OS & FS & CoT & Anl & GK & ToT \\ \hline
Accuracy 
  & S-MLLMs & 31.25 & 6.25 & 12.5 & 12.5 & 18.75 & 6.25 & 0 \\ \cline{2-9}
  & M-MLLMs & 43.75 & 37.5 & 37.5 & 37.5 & 56.25 & 43.75 & 43.75 \\ \cline{2-9}
  & L-MLLMs & 56.25 & 43.75 & 37.5 & 43.75 & 37.5 & 37.5 & 43.75 \\ \hline
Hallucination 
  & S-MLLMs & 12.5 & 25 & 37.5 & 25 & 37.5 & 43.75 & 50 \\ \cline{2-9}
  & M-MLLMs & 0 & 0 & 0 & 0 & 6.25 & 6.25 & 6.25 \\ \cline{2-9}
  & L-MLLMs & 0 & 6.25 & 0 & 0 & 25 & 6.25 & 12.5 \\ \hline
Relevance (F + P)
  & S-MLLMs & 87.5 & 75 & 75 & 81.25 & 75 & 81.25 & 62.5 \\ \cline{2-9}
  & M-MLLMs & 100 & 100 & 100 & 100 & 100 & 100 & 100 \\ \cline{2-9}
  & L-MLLMs & 100 & 100 & 100 & 100 & 93.75 & 93.75 & 93.75 \\ \hline
Irrelevance 
  & S-MLLMs & 12.5 & 25 & 25 & 18.75 & 25 & 18.75 & 37.5 \\ \cline{2-9}
  & M-MLLMs & 0 & 0 & 0 & 0 & 0 & 0 & 0 \\ \cline{2-9}
  & L-MLLMs & 0 & 0 & 0 & 0 & 6.25 & 6.25 & 6.25 \\ \hline
Conciseness (UE)
  & S-MLLMs & 50 & 50 & 50 & 37.5 & 37.5 & 31.25 & 31.25 \\ \cline{2-9}
  & M-MLLMs & 81.25 & 93.75 & 87.5 & 75 & 56.25 & 50 & 56.25 \\ \cline{2-9}
  & L-MLLMs & 68.75 & 68.75 & 75 & 56.25 & 50 & 56.25 & 68.75 \\ \hline
Conciseness (TP + OE)
  & S-MLLMs & 50 & 50 & 50 & 62.5 & 62.5 & 68.75 & 68.75 \\ \cline{2-9}
  & M-MLLMs & 18.75 & 6.25 & 12.5 & 25 & 43.75 & 50 & 43.75 \\ \cline{2-9}
  & L-MLLMs & 31.25 & 31.25 & 25 & 43.75 & 50 & 43.75 & 31.25 \\ \hline
\end{tabular}
\end{table}


\begin{table}[ht]
\centering
\caption{
EA3: Complex Code Generation and Execution Tasks Results Summary. This table reports the average performance (in \%) of Small, Medium, and Large MLLMs on code generation tasks. Performance metrics include Accuracy, Hallucination, Relevance (Fully and Partially Relevant), Irrelevance, Conciseness (Under Explained - UE, Over Explained - OE). Abbreviations: ZS = Zero-Shot, OS = One-Shot, FS = Few-Shot, CoT = Chain-of-Thought, Anl = Analogical, GK = Generated Knowledge, ToT = Tree-of-Thought.}
\label{tab:code_generation_tasks_summary_results}
\renewcommand{\arraystretch}{1.1}
\setlength{\tabcolsep}{4pt}
\begin{tabular}{|p{3.5cm}|p{1.7cm}|p{1.2cm}|p{1.2cm}|p{1.2cm}|p{1.2cm}|p{1.2cm}|p{1.2cm}|p{1.2cm}|}
\hline
Metrics & Size & ZS & OS & FS & CoT & Anl & GK & ToT \\ \hline
Accuracy 
  & S-MLLMs & 46.88 & 56.25 & 53.12 & 50 & 25 & 40.62 & 31.25 \\ \cline{2-9}
  & M-MLLMs & 78.12 & 87.50 & 90.62 & 84.38 & 65.62 & 78.12 & 78.12 \\ \cline{2-9}
  & L-MLLMs & 84.38 & 87.50 & 96.88 & 93.75 & 78.12 & 78.12 & 84.38 \\ \hline
Hallucination 
  & S-MLLMs & 37.5 & 28.12 & 31.25 & 31.25 & 40.62 & 37.5 & 46.88 \\ \cline{2-9}
  & M-MLLMs & 3.12 & 0 & 0 & 0 & 12.5 & 6.25 & 3.12 \\ \cline{2-9}
  & L-MLLMs & 0 & 0 & 0 & 0 & 15.62 & 6.25 & 6.25 \\ \hline
Relevance (F + P)
  & S-MLLMs & 81.25 & 78.12 & 84.37 & 87.50 & 71.88 & 71.88 & 75 \\ \cline{2-9}
  & M-MLLMs & 100 & 100 & 100 & 100 & 100 & 100 & 100 \\ \cline{2-9}
  & L-MLLMs & 100 & 100 & 100 & 100 & 96.88 & 100 & 100 \\ \hline
Irrelevance 
  & S-MLLMs & 18.75 & 21.88 & 15.63 & 12.5 & 28.12 & 28.12 & 25 \\ \cline{2-9}
  & M-MLLMs & 0 & 0 & 0 & 0 & 0 & 0 & 0 \\ \cline{2-9}
  & L-MLLMs & 0 & 0 & 0 & 0 & 3.12 & 0 & 0 \\ \hline
Conciseness (UE)
  & S-MLLMs & 31.25 & 56.25 & 37.5 & 37.5 & 21.88 & 34.38 & 21.88 \\ \cline{2-9}
  & M-MLLMs & 21.88 & 31.25 & 31.25 & 18.75 & 3.12 & 12.25 & 3.12 \\ \cline{2-9}
  & L-MLLMs & 9.38 & 15.62 & 21.88 & 12.5 & 6.25 & 12.25 & 3.12 \\ \hline
Conciseness (TP + OE)
  & S-MLLMs & 68.75 & 43.75 & 62.5 & 62.5 & 78.12 & 65.62 & 78.12 \\ \cline{2-9}
  & M-MLLMs & 78.12 & 68.75 & 68.75 & 81.25 & 96.88 & 87.5 & 96.88 \\ \cline{2-9}
  & L-MLLMs & 90.62 & 84.38 & 78.13 & 87.5 & 93.75 & 87.5 & 96.88 \\ \hline
\end{tabular}
\end{table}


\begin{table}[ht]
\centering
\caption{
EA4: Knowledge Retrieval and Integration Tasks Results Summary. This table presents the average performance (in \%) of Small, Medium, and Large MLLMs on knowledge retrieval tasks. Performance metrics include Accuracy, Hallucination, Relevance (Fully and Partially Relevant), Irrelevance, Conciseness (Under Explained - UE, Over Explained - OE). The prompting techniques assessed include Zero-Shot (ZS), One-Shot (OS), Few-Shot (FS), Chain-of-Thought (CoT), Analogical (Anl), Generated Knowledge (GK), and Tree of Thought (ToT).
}
\label{tab:knowledge_retrieval_results_summary}
\renewcommand{\arraystretch}{1.1}
\setlength{\tabcolsep}{4pt}
\begin{tabular}{|p{3.3cm}|p{1.7cm}|p{1.3cm}|p{1.3cm}|p{1.3cm}|p{1.3cm}|p{1.3cm}|p{1.3cm}|p{1.3cm}|}
\hline
Metrics & Size & ZS & OS & FS & CoT & Anl & GK & ToT \\ \hline
Accuracy 
  & S-MLLMs & 43.75 & 34.38 & 37.5 & 37.5 & 21.88 & 28.12 & 32.26 \\ \cline{2-9}
  & M-MLLMs & 71.88 & 59.38 & 62.5 & 62.5 & 53.12 & 53.12 & 62.5 \\ \cline{2-9}
  & L-MLLMs & 87.5 & 77.5 & 75 & 77.5 & 75 & 53.12 & 64.1 \\ \hline
Hallucination 
  & S-MLLMs & 25 & 31.25 & 34.38 & 25 & 40.62 & 43.75 & 51.61 \\ \cline{2-9}
  & M-MLLMs & 0 & 0 & 3.12 & 0 & 6.25 & 0 & 6.25 \\ \cline{2-9}
  & L-MLLMs & 0 & 0 & 0 & 0 & 7.5 & 0 & 2.56 \\ \hline
Relevance (F + P)
  & S-MLLMs & 81.25 & 78.12 & 71.88 & 78.12 & 68.75 & 68.75 & 67.74 \\ \cline{2-9}
  & M-MLLMs & 100 & 100 & 96.88 & 100 & 100 & 96.88 & 93.75 \\ \cline{2-9}
  & L-MLLMs & 97.5 & 100 & 100 & 100 & 100 & 96.88 & 94.87 \\ \hline
Irrelevance 
  & S-MLLMs & 18.75 & 21.88 & 28.12 & 21.88 & 31.25 & 31.25 & 32.26 \\ \cline{2-9}
  & M-MLLMs & 0 & 0 & 0 & 0 & 0 & 0 & 0 \\ \cline{2-9}
  & L-MLLMs & 2.5 & 0 & 0 & 0 & 0 & 3.12 & 5.13 \\ \hline
Conciseness (UE)
  & S-MLLMs & 53.12 & 31.25 & 37.5 & 28.12 & 15.62 & 21.88 & 19.35 \\ \cline{2-9}
  & M-MLLMs & 28.12 & 62.5 & 62.5 & 43.75 & 53.12 & 46.88 & 34.38 \\ \cline{2-9}
  & L-MLLMs & 82.5 & 87.5 & 82.5 & 55 & 55 & 46.88 & 41.03 \\ \hline
Conciseness (TP + OE)
  & S-MLLMs & 46.88 & 68.75 & 62.5 & 71.88 & 84.38 & 78.12 & 80.65 \\ \cline{2-9}
  & M-MLLMs & 71.88 & 37.5 & 37.5 & 56.25 & 46.88 & 53.12 & 65.42 \\ \cline{2-9}
  & L-MLLMs & 17.5 & 12.5 & 17.5 & 45 & 45 & 53.12 & 58.97 \\ \hline
\end{tabular}
\end{table}

\subsection{Performance Profiling of MLLMs}
To better understand the computational efficiency and generative behavior of the evaluated models, we analysed response time, output length, and memory utilization across all four evaluation aspects. Descriptive statistics, including average (AVG), standard deviation (STD), median (MEDIAN), minimum (MIN), and maximum (MAX) were computed for each metric and are presented in Tables~\ref{tab:time_statistics_for_reasoning} -- \ref{tab:char_statistics_for_knowledge_retrieval}.

The results are discussed in three parts. First, we examine model-specific trends in response time across prompting methods and task types, highlighting differences across small, medium, and large MLLMs. Next, we present evaluation aspect-specific insights, comparing how models behave across reasoning, multimodal understanding, code generation, and knowledge retrieval tasks. Finally, we report the memory footprint of each model, noting significant variation based on architecture and scale.

\subsubsection{Model-Specific Observations:}
For small MLLMs, CoT prompting exhibited stable processing times across tasks, ranging approximately from 5.5 to 6.2 seconds. Among all techniques, Analogical and ToT prompting resulted in the longest processing times, with knowledge retrieval tasks requiring the most time and code generation tasks the least. 

Medium MLLMs showed similar trends, where Analogical prompting remained the slowest across tasks. Few-Shot prompting was generally the fastest, although in model understanding and code generation tasks, One-Shot prompting was marginally quicker.

For large MLLMs, Analogical and ToT prompting consistently led to the highest processing times. Few-Shot prompting was typically faster across tasks, except in reasoning and code generation, where the time differences were minimal. CoT prompting was notably stable for large models, with processing durations ranging between 13.5 and 16 seconds.

\subsubsection{Evaluation Aspect Specific Obeservations}
For EA1 (Reasoning and Compositionality) tasks, 
Table~\ref{tab:time_statistics_for_reasoning} presents the response time statistics for reasoning tasks, and Table~\ref{tab:char_statistics_for_reasoning} shows the corresponding output lengths. For Small MLLMs, response times are relatively short (AVG $\approx$ 5.3--6.2 s) and outputs are concise, while Large MLLMs exhibit higher variance in response time (AVG up to 22.97 s) and produce more verbose outputs.

For EA2 (Multimodal Understanding and Alignment) tasks,   
Tables~\ref{tab:time_statistics_for_model_understanding} and \ref{tab:char_statistics_for_model_understanding} provide the response time and output length statistics for tasks requiring multimodal understanding. Small MLLMs respond faster (AVG $\approx$ 5.91 s with ZS) and generate shorter outputs (AVG $\approx$ 917 characters with ZS), whereas Large MLLMs require significantly more time (AVG $\approx$ 24.67 s with ZS) and produce longer responses (AVG $\approx$ 1417 characters with ZS).

For EA3 (Complex Code Generation and Execution) tasks,  
Tables~\ref{tab:time_statistics_for_code_generation} and \ref{tab:char_statistics_for_code_generation} report the response time and output length for code generation tasks. Large MLLMs show the highest accuracy with Few-Shot prompting and require longer response times and output lengths compared to smaller models.

For EA4 (Knowledge Retrieval and Integration) tasks,  Tables~\ref{tab:time_statistics_for_knowledge_retrieval} and \ref{tab:char_statistics_for_knowledge_retrieval} detail the response time and output length for knowledge retrieval tasks. In these tasks, Large MLLMs achieve high accuracy with minimal hallucination and produce longer, more detailed responses than Small and Medium models.

\begin{table}[ht]
\centering
\caption{
Response Time Statistics for EA1 (Reasoning and Compositionality) Tasks. Descriptive statistics (in seconds: AVG, STD, MEDIAN, MIN, MAX) are provided for each prompting technique for Small, Medium, and Large MLLMs. Abbreviations: ZS = Zero-Shot, OS = One-Shot, FS = Few-Shot, CoT = Chain-of-Thought, Anl = Analogical, GK = Generated Knowledge, ToT = Tree-of-Thought.
\label{tab:time_statistics_for_reasoning}
}
\renewcommand{\arraystretch}{1.1}
\setlength{\tabcolsep}{4pt}
\begin{tabular}{|c|l|l|l|l|l|l|}
\hline
\textbf{Model Category} & \textbf{Prompt Technique} & \textbf{AVG} & \textbf{STD} & \textbf{MEDIAN} & \textbf{MIN} & \textbf{MAX} \\ \hline
Small MLLM  & ZS  & 6.24  & 5.24  & 5.17  & 0.49  & 14.51 \\ \cline{2-7} 
            & OS  & 5.55  & 4.73  & 4.80  & 0.49  & 13.95 \\ \cline{2-7} 
            & FS  & 5.30  & 4.69  & 4.45  & 0.48  & 14.52 \\ \cline{2-7} 
            & CoT & 6.51  & 3.91  & 6.15  & 0.87  & 13.91 \\ \cline{2-7} 
            & Anl & 7.47  & 4.68  & 7.51  & 0.81  & 14.48 \\ \cline{2-7} 
            & GK  & 4.83  & 3.93  & 3.36  & 0.55  & 13.95 \\ \cline{2-7} 
            & ToT & 9.51  & 6.26  & 10.33 & 0.47  & 19.22 \\ \hline
Medium MLLM & ZS  & 9.30  & 4.62  & 8.09  & 1.30  & 17.10 \\ \cline{2-7} 
            & OS  & 9.52  & 5.74  & 8.69  & 1.32  & 24.29 \\ \cline{2-7} 
            & FS  & 9.00  & 5.73  & 7.33  & 1.27  & 23.12 \\ \cline{2-7} 
            & CoT & 10.86 & 5.89  & 8.63  & 4.74  & 25.75 \\ \cline{2-7} 
            & Anl & 12.35 & 5.87  & 11.80 & 3.34  & 25.38 \\ \cline{2-7} 
            & GK  & 12.02 & 7.39  & 9.49  & 4.00  & 29.60 \\ \cline{2-7} 
            & ToT & 13.52 & 5.74  & 11.88 & 7.49  & 30.02 \\ \hline
Large MLLM  & ZS  & 22.97 & 26.38 & 12.74 & 1.90  & 101.69 \\ \cline{2-7} 
            & OS  & 19.57 & 25.01 & 10.90 & 0.40  & 101.08 \\ \cline{2-7} 
            & FS  & 21.28 & 26.46 & 10.73 & 0.44  & 101.24 \\ \cline{2-7} 
            & CoT & 26.07 & 30.67 & 13.70 & 0.48  & 101.71 \\ \cline{2-7} 
            & Anl & 29.29 & 35.76 & 16.72 & 0.39  & 120.92 \\ \cline{2-7} 
            & GK  & 20.95 & 24.55 & 13.24 & 0.67  & 101.56 \\ \cline{2-7} 
            & ToT & 26.67 & 26.80 & 16.12 & 0.55  & 101.85 \\ \hline
\end{tabular}

\end{table}


\begin{table}[H]
\centering
\caption{
Response Length (Character) Statistics for EA1 (Reasoning and Compositionality) Tasks. This table reports the average output lengths (in characters) along with STD, MEDIAN, MIN, and MAX values for Small, Medium, and Large MLLMs across the prompting techniques. Abbreviations: ZS = Zero-Shot, OS = One-Shot, FS = Few-Shot, CoT = Chain-of-Thought, Anl = Analogical, GK = Generated Knowledge, ToT = Tree-of-Thought.
\label{tab:char_statistics_for_reasoning}
}
\renewcommand{\arraystretch}{1.1}
\setlength{\tabcolsep}{4pt}
\begin{tabular}{|c|l|l|l|l|l|l|}
\hline
Model Category & Prompt Technique & AVG   & STD   & MEDIAN & MIN  & MAX  \\ \hline
Small MLLM  & ZS  & 878  & 819  & 751  & 11   & 2294 \\ \cline{2-7} 
            & OS  & 763  & 656  & 839    & 11   & 1712 \\ \cline{2-7} 
            & FS  & 785  & 753  & 658  & 11   & 2699 \\ \cline{2-7} 
            & CoT & 960  & 589  & 1069 & 17   & 1869 \\ \cline{2-7} 
            & Anl & 1147 & 823  & 1157   & 67   & 2800 \\ \cline{2-7} 
            & GK  & 670  & 535  & 607    & 1    & 1946 \\ \cline{2-7} 
            & ToT & 1475 & 1008 & 1550 & 1    & 2744 \\ \hline
Medium MLLM & ZS  & 1082 & 596  & 1105 & 82   & 2294 \\ \cline{2-7} 
            & OS  & 1094 & 645  & 873    & 81   & 2174 \\ \cline{2-7} 
            & FS  & 1022 & 677  & 889    & 73   & 2124 \\ \cline{2-7} 
            & CoT & 1151 & 496  & 1074 & 452  & 2402 \\ \cline{2-7} 
            & Anl & 1482 & 663  & 1406   & 397  & 3190 \\ \cline{2-7} 
            & GK  & 1339 & 707  & 1193   & 409  & 3343 \\ \cline{2-7} 
            & ToT & 1645 & 705  & 1573 & 691  & 3845 \\ \hline
Large MLLM  & ZS  & 1417 & 780  & 1313   & 81   & 3111 \\ \cline{2-7} 
            & OS  & 1179 & 630  & 1164 & 86   & 2437 \\ \cline{2-7} 
            & FS  & 1206 & 708  & 1083 & 79   & 2580 \\ \cline{2-7} 
            & CoT & 1459 & 642  & 1410 & 438  & 2810 \\ \cline{2-7} 
            & Anl & 1630 & 660  & 1595   & 392  & 2725 \\ \cline{2-7} 
            & GK  & 1435 & 574  & 1439   & 453  & 2457 \\ \cline{2-7} 
            & ToT & 1768 & 905  & 1602   & 770  & 4432 \\ \hline
\end{tabular}
\end{table}

\begin{table}[H]
\centering
\caption{Response Time Statistics for EA2 (Model Understanding and Alignment) Tasks. This table presents the response time (in seconds) descriptive statistics for each prompting technique for Small, Medium, and Large MLLMs. Abbreviations: ZS = Zero-Shot, OS = One-Shot, FS = Few-Shot, CoT = Chain-of-Thought, Anl = Analogical, GK = Generated Knowledge, ToT = Tree-of-Thought.
}
\label{tab:time_statistics_for_model_understanding}
\renewcommand{\arraystretch}{1.1}
\setlength{\tabcolsep}{4pt}
\begin{tabular}{|c|l|l|l|l|l|l|}
\hline
Model Category & Prompt Technique & AVG  & STD  & MEDIAN & MIN  & MAX  \\ \hline
Small MLLM  & ZS  & 5.91  & 4.40  & 4.76  & 0.42  & 15.01 \\ \cline{2-7} 
            & OS  & 3.60  & 2.08  & 3.84  & 0.78  & 7.67  \\ \cline{2-7} 
            & FS  & 3.80  & 2.32  & 3.96  & 0.51  & 8.55  \\ \cline{2-7} 
            & CoT & 5.90  & 3.80  & 5.64  & 0.61  & 14.24 \\ \cline{2-7} 
            & Anl & 8.54  & 5.46  & 10.26 & 0.65  & 16.08 \\ \cline{2-7} 
            & GK  & 6.21  & 4.54  & 5.12  & 0.61  & 14.53 \\ \cline{2-7} 
            & ToT & 6.95  & 4.97  & 6.68  & 0.62  & 18.72 \\ \hline
Medium MLLM & ZS  & 8.69  & 3.25  & 8.32  & 2.64  & 14.68 \\ \cline{2-7} 
            & OS  & 9.05  & 3.69  & 7.71  & 4.37  & 15.40 \\ \cline{2-7} 
            & FS  & 9.25  & 3.90  & 7.91  & 4.09  & 17.42 \\ \cline{2-7} 
            & CoT & 11.98 & 3.02  & 12.01 & 7.73  & 19.84 \\ \cline{2-7} 
            & Anl & 14.05 & 4.89  & 14.94 & 7.96  & 25.33 \\ \cline{2-7} 
            & GK  & 11.29 & 4.98  & 11.24 & 3.72  & 21.71 \\ \cline{2-7} 
            & ToT & 13.85 & 7.07  & 12.18 & 3.63  & 32.82 \\ \hline
Large MLLM  & ZS  & 24.67 & 27.87 & 15.15 & 0.95  & 102.57 \\ \cline{2-7} 
            & OS  & 19.82 & 21.79 & 13.15 & 0.60  & 77.78  \\ \cline{2-7} 
            & FS  & 19.70 & 24.63 & 14.13 & 0.76  & 113.42 \\ \cline{2-7} 
            & CoT & 25.06 & 27.28 & 15.86 & 0.87  & 103.77 \\ \cline{2-7} 
            & Anl & 31.67 & 34.42 & 20.38 & 0.68  & 103.31 \\ \cline{2-7} 
            & GK  & 25.67 & 29.44 & 13.93 & 0.90  & 99.00  \\ \cline{2-7} 
            & ToT & 28.89 & 29.90 & 19.34 & 0.94  & 119.25 \\ \hline
\end{tabular}
\end{table}

\begin{table}[H]
\centering
\caption{
Response Length (Character) Statistics for EA2 (Model Understanding and Alignment) Tasks. This table lists the output length (in characters) descriptive statistics for Small, Medium, and Large MLLMs. Abbreviations: ZS = Zero-Shot, OS = One-Shot, FS = Few-Shot, CoT = Chain-of-Thought, Anl = Analogical, GK = Generated Knowledge, ToT = Tree-of-Thought.
}
\label{tab:char_statistics_for_model_understanding}
\renewcommand{\arraystretch}{1.1}
\setlength{\tabcolsep}{4pt}
\begin{tabular}{|c|l|l|l|l|l|l|}
\hline
Model Category & Prompt Technique & AVG   & STD   & MEDIAN & MIN  & MAX  \\ \hline
Small MLLM  & ZS  & 917  & 751  & 706  & 4    & 2100 \\ \cline{2-7} 
            & OS  & 565  & 419  & 522  & 62   & 1558 \\ \cline{2-7} 
            & FS  & 608  & 492  & 617  & 4    & 1790 \\ \cline{2-7} 
            & CoT & 894  & 614  & 799  & 20   & 1979 \\ \cline{2-7} 
            & Anl & 1319 & 841  & 1640 & 11   & 2380 \\ \cline{2-7} 
            & GK  & 1008 & 895  & 620  & 7    & 2612 \\ \cline{2-7} 
            & ToT & 1087 & 802  & 1113 & 8    & 2553 \\ \hline
Medium MLLM & ZS  & 1007 & 452  & 967  & 202  & 1844 \\ \cline{2-7} 
            & OS  & 985  & 416  & 931  & 280  & 1857 \\ \cline{2-7} 
            & FS  & 1030 & 444  & 1048 & 406  & 1899 \\ \cline{2-7} 
            & CoT & 1151 & 496  & 1074 & 452  & 2402 \\ \cline{2-7} 
            & Anl & 1482 & 663  & 1406   & 397  & 3190 \\ \cline{2-7} 
            & GK  & 1339 & 707  & 1193   & 409  & 3343 \\ \cline{2-7} 
            & ToT & 1645 & 705  & 1573 & 691  & 3845 \\ \hline
Large MLLM  & ZS  & 1417 & 780  & 1313   & 81   & 3111 \\ \cline{2-7} 
            & OS  & 1179 & 630  & 1164 & 86   & 2437 \\ \cline{2-7} 
            & FS  & 1206 & 708  & 1083 & 79   & 2580 \\ \cline{2-7} 
            & CoT & 1459 & 642  & 1410 & 438  & 2810 \\ \cline{2-7} 
            & Anl & 1630 & 660  & 1595   & 392  & 2725 \\ \cline{2-7} 
            & GK  & 1435 & 574  & 1439   & 453  & 2457 \\ \cline{2-7} 
            & ToT & 1768 & 905  & 1602   & 770  & 4432 \\ \hline
\end{tabular}
\end{table}

\begin{table}[H]
\centering
\caption{
Response Time Statistics for EA3 (Code Generation and Execution) Tasks. Descriptive statistics (in seconds: AVG, STD, MEDIAN, MIN, MAX) for response time across prompting techniques for Small, Medium, and Large MLLMs. Abbreviations: ZS = Zero-Shot, OS = One-Shot, FS = Few-Shot, CoT = Chain-of-Thought, Anl = Analogical, GK = Generated Knowledge, ToT = Tree-of-Thought.
}
\label{tab:time_statistics_for_code_generation}
\renewcommand{\arraystretch}{1.1}
\setlength{\tabcolsep}{4pt}
\begin{tabular}{|c|l|l|l|l|l|l|}
\hline
Model Category & Prompt Technique & AVG  & STD  & MEDIAN & MIN  & MAX  \\ \hline
Small MLLM  & ZS  & 5.35  & 3.95  & 4.97  & 0.62  & 15.46 \\ \cline{2-7} 
            & OS  & 3.47  & 2.69  & 2.98  & 0.47  & 14.68 \\ \cline{2-7} 
            & FS  & 4.64  & 3.50  & 3.82  & 0.65  & 14.63 \\ \cline{2-7} 
            & CoT & 6.28  & 4.67  & 6.05  & 0.64  & 15.03 \\ \cline{2-7} 
            & Anl & 7.23  & 4.16  & 7.13  & 0.85  & 14.67 \\ \cline{2-7} 
            & GK  & 5.18  & 3.94  & 4.63  & 0.57  & 14.68 \\ \cline{2-7} 
            & ToT & 8.50  & 5.64  & 8.28  & 0.63  & 20.91 \\ \hline
Medium MLLM & ZS  & 6.77  & 3.55  & 6.63  & 1.53  & 17.50 \\ \cline{2-7} 
            & OS  & 4.48  & 1.38  & 4.26  & 2.15  & 7.73  \\ \cline{2-7} 
            & FS  & 4.60  & 1.44  & 4.32  & 1.96  & 7.21  \\ \cline{2-7} 
            & CoT & 7.40  & 4.70  & 6.43  & 1.58  & 24.34 \\ \cline{2-7} 
            & Anl & 13.08 & 6.02  & 11.37 & 5.65  & 31.50 \\ \cline{2-7} 
            & GK  & 7.46  & 3.29  & 7.61  & 1.98  & 15.04 \\ \cline{2-7} 
            & ToT & 12.67 & 6.21  & 11.02 & 3.31  & 30.59 \\ \hline
Large MLLM  & ZS  & 16.48 & 15.63 & 11.41 & 1.90  & 68.60 \\ \cline{2-7} 
            & OS  & 12.63 & 11.82 & 11.37 & 0.31  & 50.79 \\ \cline{2-7} 
            & FS  & 12.02 & 10.90 & 10.82 & 0.37  & 50.76 \\ \cline{2-7} 
            & CoT & 19.70 & 23.71 & 13.89 & 0.53  & 97.30 \\ \cline{2-7} 
            & Anl & 34.01 & 38.85 & 20.91 & 0.44  & 126.27\\ \cline{2-7} 
            & GK  & 17.78 & 20.62 & 13.34 & 0.65  & 88.48 \\ \cline{2-7} 
            & ToT & 29.46 & 33.48 & 20.44 & 0.62  & 124.55 \\ \hline
\end{tabular}
\end{table}

\begin{table}[H]
\centering
\caption{
Response Length (Character) Statistics for EA3 (Code Generation and Execution) Tasks. This table reports the output length (in characters) descriptive statistics for each prompting technique for Small, Medium, and Large MLLMs. Abbreviations: ZS = Zero-Shot, OS = One-Shot, FS = Few-Shot, CoT = Chain-of-Thought, Anl = Analogical, GK = Generated Knowledge, ToT = Tree-of-Thought.
}
\label{tab:char_statistics_for_code_generation}
\renewcommand{\arraystretch}{1.1}
\setlength{\tabcolsep}{4pt}
\begin{tabular}{|c|l|l|l|l|l|l|}
\hline
Model Category & Prompt Technique & AVG   & STD   & MEDIAN & MIN  & MAX  \\ \hline
Small MLLM  & ZS  & 577  & 417   & 562  & 41   & 1621 \\ \cline{2-7} 
            & OS  & 364  & 340   & 256  & 9    & 1690 \\ \cline{2-7} 
            & FS  & 565  & 494   & 392  & 26   & 2054 \\ \cline{2-7} 
            & CoT & 736  & 619   & 807    & 10   & 2189 \\ \cline{2-7} 
            & Anl & 1009   & 592   & 1134 & 61   & 2321 \\ \cline{2-7} 
            & GK  & 637  & 555   & 396    & 4    & 1880 \\ \cline{2-7} 
            & ToT & 1129   & 772   & 1192 & 2    & 2468 \\ \hline
Medium MLLM & ZS  & 559  & 345   & 467    & 157  & 1426 \\ \cline{2-7} 
            & OS  & 353    & 177   & 316  & 176  & 1039 \\ \cline{2-7} 
            & FS  & 374  & 201   & 341  & 176  & 899  \\ \cline{2-7} 
            & CoT & 671  & 461   & 513    & 131  & 1571 \\ \cline{2-7} 
            & Anl & 1304   & 550   & 1176   & 478  & 2946 \\ \cline{2-7} 
            & GK  & 691  & 365   & 655  & 163  & 1383 \\ \cline{2-7} 
            & ToT & 1292   & 599   & 1311 & 161  & 2827 \\ \hline
Large MLLM  & ZS  & 995  & 591   & 875    & 112  & 2799 \\ \cline{2-7} 
            & OS  & 980    & 813   & 771    & 170  & 3585 \\ \cline{2-7} 
            & FS  & 780  & 489   & 748  & 112  & 2143 \\ \cline{2-7} 
            & CoT & 1130   & 654   & 1074 & 142  & 2451 \\ \cline{2-7} 
            & Anl & 1692   & 672   & 1770 & 485  & 3246 \\ \cline{2-7} 
            & GK  & 1016   & 485   & 1080 & 117  & 1911 \\ \cline{2-7} 
            & ToT & 1527   & 657   & 1520 & 122  & 2807 \\ \hline
\end{tabular}
\end{table}

\begin{table}[H]
\centering
\caption{
Response Time Statistics for EA4 (Knowledge Retrieval and Integration) Tasks. This table provides descriptive statistics (in seconds) for response time across the seven prompting techniques for Small, Medium, and Large MLLMs. Abbreviations: ZS = Zero-Shot, OS = One-Shot, FS = Few-Shot, CoT = Chain-of-Thought, Anl = Analogical, GK = Generated Knowledge, ToT = Tree-of-Thought.
}
\label{tab:time_statistics_for_knowledge_retrieval}
\renewcommand{\arraystretch}{1.1}
\setlength{\tabcolsep}{4pt}
\begin{tabular}{|c|l|l|l|l|l|l|}
\hline
Model Category & Prompt Technique & AVG  & STD  & MEDIAN & MIN  & MAX  \\ \hline
Small MLLM  & ZS  & 8.08  & 5.11  & 8.56  & 0.74  & 18.01 \\ \cline{2-7} 
            & OS  & 5.54  & 3.17  & 6.02  & 0.49  & 12.51 \\ \cline{2-7} 
            & FS  & 6.24  & 3.84  & 6.21  & 0.60  & 14.19 \\ \cline{2-7} 
            & CoT & 6.19  & 3.62  & 6.14  & 0.62  & 13.41 \\ \cline{2-7} 
            & Anl & 9.97  & 5.59  & 11.71 & 0.86  & 18.20 \\ \cline{2-7} 
            & GK  & 5.99  & 4.01  & 6.37  & 0.65  & 14.59 \\ \cline{2-7} 
            & ToT & 8.17  & 5.64  & 8.01  & 0.82  & 19.09 \\ \hline
Medium MLLM & ZS  & 9.95  & 3.19  & 9.54  & 4.72  & 17.92 \\ \cline{2-7} 
            & OS  & 9.31  & 4.29  & 8.00  & 4.31  & 21.60 \\ \cline{2-7} 
            & FS  & 8.42  & 3.70  & 7.44  & 2.81  & 20.88 \\ \cline{2-7} 
            & CoT & 11.48 & 4.68  & 9.97  & 5.85  & 24.76 \\ \cline{2-7} 
            & Anl & 14.83 & 6.26  & 12.80 & 4.34  & 26.58 \\ \cline{2-7} 
            & GK  & 12.49 & 6.39  & 10.63 & 0.95  & 29.66 \\ \cline{2-7} 
            & ToT & 13.44 & 5.80  & 11.99 & 0.76  & 30.51 \\ \hline
Large MLLM  & ZS  & 29.54 & 36.69 & 14.72 & 2.22  & 122.41 \\ \cline{2-7} 
            & OS  & 21.42 & 24.86 & 13.45 & 0.62  & 117.56 \\ \cline{2-7} 
            & FS  & 21.69 & 24.82 & 13.33 & 0.75  & 89.18  \\ \cline{2-7} 
            & CoT & 27.32 & 32.82 & 15.00 & 0.90  & 122.06 \\ \cline{2-7} 
            & Anl & 34.61 & 39.30 & 19.70 & 0.78  & 123.03 \\ \cline{2-7} 
            & GK  & 26.50 & 32.38 & 14.46 & 0.95  & 123.01 \\ \cline{2-7} 
            & ToT & 31.71 & 36.70 & 19.34 & 0.95  & 123.46 \\ \hline
\end{tabular}
\end{table}

\begin{table}[H]
\centering
\caption{
Response Length (Character) Statistics for EA4 (Knowledge Retrieval and Integration) Tasks. This table provides output length (in characters) descriptive statistics for Small, Medium, and Large MLLMs across the seven prompting techniques. Abbreviations: ZS = Zero-Shot, OS = One-Shot, FS = Few-Shot, CoT = Chain-of-Thought, Anl = Analogical, GK = Generated Knowledge, ToT = Tree-of-Thought.
}
\label{tab:char_statistics_for_knowledge_retrieval}
\renewcommand{\arraystretch}{1.1}
\setlength{\tabcolsep}{4pt}
\begin{tabular}{|c|l|l|l|l|l|l|}
\hline
\textbf{Model Category} & \textbf{Prompt} & \textbf{AVG} & \textbf{STD} & \textbf{MEDIAN} & \textbf{MIN} & \textbf{MAX} \\ \hline
Small MLLM  & ZS  & 1236  & 780  & 1186 & 39   & 2649 \\ \cline{2-7}
            & OS  & 859 & 500  & 987 & 9    & 1642 \\ \cline{2-7}
            & FS  & 984 & 664  & 987  & 10   & 2483 \\ \cline{2-7}
            & CoT & 956 & 597  & 980  & 9    & 2371 \\ \cline{2-7}
            & Anl & 1572  & 892  & 2018 & 62   & 2509 \\ \cline{2-7}
            & GK  & 939.5 & 710  & 935  & 7    & 2680 \\ \cline{2-7}
            & ToT & 1266  & 890  & 1266 & 29   & 2588 \\ \hline
Medium MLLM & ZS  & 1161  & 325  & 1203 & 449  & 1725 \\ \cline{2-7}
            & OS  & 1042  & 389  & 1012 & 453  & 2111 \\ \cline{2-7}
            & FS  & 974 & 430  & 837  & 212  & 2332 \\ \cline{2-7}
            & CoT & 1281  & 398  & 1253 & 673  & 2375 \\ \cline{2-7}
            & Anl & 1742  & 488  & 1772 & 676  & 2743 \\ \cline{2-7}
            & GK  & 1415  & 734  & 1314 & 588  & 4174 \\ \cline{2-7}
            & ToT & 1520  & 584  & 1404 & 624  & 3439 \\ \hline
Large MLLM  & ZS  & 1655  & 599  & 1726 & 607  & 3879 \\ \cline{2-7}
            & OS  & 1436  & 552  & 1321 & 612  & 2719 \\ \cline{2-7}
            & FS  & 1310  & 508  & 1217 & 613  & 2597 \\ \cline{2-7}
            & CoT & 1590  & 663  & 1469 & 448  & 3175 \\ \cline{2-7}
            & Anl & 2103  & 1263 & 1855 & 542  & 6551 \\ \cline{2-7}
            & GK  & 1523  & 691  & 1295 & 549  & 3074 \\ \cline{2-7}
            & ToT & 1892  & 940  & 1747 & 579  & 4965 \\ \hline
\end{tabular}
\end{table}

\subsubsection{Model Memory Allocation}

Table~\ref{tab:model_memory_size_by_category} lists the selected MLLMs along with their categorization (Small, Medium, or Large), model size (in billions), and allocated GPU memory (in GB). For example, the Pixtral 12B model uses 35\,GB, while the InternVL2-1B model uses only 0.05\,GB. These variations highlight the impact of architecture-specific optimizations in addition to model size.

In summary, our analysis reveals consistent patterns across prompting techniques and model sizes. Analogical prompting typically resulted in the longest response times and the most verbose outputs, followed by Tree-of-Thought (ToT). In contrast, Few-Shot and One-Shot prompting were generally faster and produced more concise outputs. Among task types, code generation was the fastest to process, while response time tended to increase with model size. 

\begin{table}[H]
\centering
\normalsize  
\caption{
Model Memory Allocation and Categorization. This table lists individual MLLMs along with their category (Small, Medium, or Large), model size (in billions), and allocated GPU memory (in GB).
}
\label{tab:model_memory_size_by_category}
\renewcommand{\arraystretch}{1.1}
\setlength{\tabcolsep}{4pt}
\begin{tabular}{|c|l|l|l|}
\hline
\textbf{Model Category} & \textbf{Model Name}    & \textbf{Model Size (B)} & \textbf{Allocated Memory (GB)} \\ \hline
\multirow{4}{*}{Small MLLM} & InternVL2-1B         & 1    & 0.05 \\ \cline{2-4} 
                           & Qwen2-vl             & 2    & 4.16 \\ \cline{2-4} 
                           & MiniMonkey           & 2.2  & 4.49 \\ \cline{2-4} 
                           & Paligemma            & 3    & 10.96 \\ \hline
\multirow{4}{*}{Medium MLLM} & Phi3.5               & 4    & 7.77  \\ \cline{2-4} 
                           & Llava-one-vision     & 8    & 15.05 \\ \cline{2-4} 
                           & Ovis1.5-8B           & 8    & 17.33 \\ \cline{2-4} 
                           & Glm-4v               & 9    & 25.94 \\ \hline
\multirow{5}{*}{Large MLLM}  & Ovis1.6              & 10.2 & 19.02 \\ \cline{2-4} 
                           & Llama3.2-vision      & 11   & 11.54 \\ \cline{2-4} 
                           & Pixtral              & 12   & 35.30 \\ \cline{2-4} 
                           & Omchat               & 13   & 24.62 \\ \cline{2-4} 
                           & InternVL2-26B        & 26   & 25.27 \\ \hline
\end{tabular}
\end{table}

\section{Discussion}
\label{sec:discussion}

Our evaluation reveals significant variations in MLLM performance across task types and prompting techniques. Overall, larger models consistently outperform medium and small models, especially in tasks such as knowledge retrieval and code generation, yet tasks requiring complex reasoning and nuanced understanding still yield relatively low accuracies (often below 60\%). This disparity indicates that while scaling improves certain capabilities, even the largest models struggle with multi-step reasoning and abstract deduction.

For tasks that require multi-step reasoning and compositional problem solving, our results show that providing multiple examples (Few-Shot prompting) enhances accuracy in large models. However, more structured prompting approaches (such as Chain-of-Thought, Analogical, and Tree-of-Thought) tend to increase hallucination rates, particularly in smaller models. This suggests that, although structured prompts are designed to guide logical inference by encouraging intermediate reasoning steps, they can sometimes introduce extraneous or confabulated details that ultimately undermine output quality. 

In tasks demanding multimodal understanding and alignment (EA2), large models achieve near-perfect relevance scores when using simpler prompting strategies (e.g., Zero-Shot and One-Shot). This implies that pre-trained multimodal embeddings in these models are highly effective at integrating text and visual inputs. 

Zero-Shot prompting emerged as the most effective technique in EA2, achieving the highest accuracy and lowest hallucination rates across model sizes. This suggests that MLLMs may rely heavily on pre-trained multimodal embeddings rather than explicit reasoning across different modalities. In contrast, complex reasoning-based prompts, such as Analogical and Tree-of-Thought, degraded performance, indicating that multimodal models struggle when required to interpret and synthesize abstract relationships between text and images. These results highlight limitations in current MLLMs’ spatial and contextual awareness, which are critical for applications such as visual question answering (VQA), AI-generated content moderation, and automated medical image interpretation. While MLLMs can extract information from multimodal inputs, they lack deep semantic alignment; a challenge that must be addressed before deploying these models in high-risk environments. Given these limitations, current open-source MLLMs cannot be relied upon for reasoning-intensive tasks without human oversight or external validation mechanisms.

Code generation tasks exhibited the highest accuracies across all model sizes. The structured nature of programming tasks appears to benefit from Few-Shot prompting, which provides clear examples that guide both syntactic and semantic generation. Nonetheless, hallucination remains an issue, especially in smaller models; which is critical in contexts such as software development and cybersecurity where errors can have severe consequences.

Knowledge retrieval tasks also demonstrate the advantage of scaling: large MLLMs achieve the highest accuracy and relevance, particularly with Zero-Shot prompting. However, these models sometimes present outputs with unwarranted confidence, even when portions of the retrieved information are incorrect. This lack of reliable verification is problematic in domains that demand high factual accuracy, such as legal, medical, and scientific applications.

Hallucination remained a fundamental challenge across all models and prompting strategies, particularly in tasks requiring abstract reasoning. Analogical, General Knowledge, and Tree-of-Thought prompting exhibited the highest hallucination rates, suggesting that the current implementations of structured reasoning within MLLMs remain unreliable. This is especially concerning in safety-critical applications where factual correctness is imperative, such as AI-generated medical reports, legal document drafting, and financial risk assessments.

Our analysis of response times and output lengths further underscores the trade-offs inherent in different prompting techniques. More complex methods like Analogical and Tree-of-Thought prompting require longer processing times and produce more verbose outputs, while One-Shot and Few-Shot prompting yield faster and more concise responses. Although larger models generally incur higher computational costs and longer response times, the improvements in accuracy and relevance, particularly for multimodal understanding and knowledge retrieval often justify these trade-offs.

In summary, no single prompting method optimally addresses every task. The effectiveness of a prompting strategy is highly dependent on the nature of the task and the model scale. For instance, Few-Shot prompting appears best suited for structured tasks like code generation, while simpler prompting techniques are more effective for multimodal alignment. These findings suggest that hybrid approaches, combining example-based prompts with selective structured reasoning, may offer a promising path toward more reliable and contextually aware multimodal reasoning.

\subsection{Implications and Use Cases}

These findings highlight critical implications for deploying MLLMs in real-world scenarios. While large MLLMs demonstrate strong retrieval and structured output generation, their failure in logical reasoning and multimodal alignment indicates that they are currently unsuitable for fully autonomous decision-making systems in healthcare, finance, or legal domains. Instead, their most effective applications lie in AI-assisted software development, where few-shot prompting can improve code generation workflows while integrating human validation to mitigate errors. Automated knowledge retrieval systems, where large MLLMs can assist in search and summarization tasks but require additional verification mechanisms to ensure factual accuracy. AI-powered tutoring systems, where structured output generation can support educational applications, but deeper logical reasoning capabilities must be further refined. Visual question answering and multimodal content moderation, where large MLLMs can be used to process images and text but require improvements in contextual alignment. Conversely, caution must be exercised when integrating MLLMs into fields where reasoning-based accuracy is paramount. Current models struggle to maintain logical consistency in long-form reasoning tasks, limiting their utility in legal contract analysis, autonomous robotic planning, and financial forecasting.

\subsection{Future Research Directions}

The findings of this study highlight several critical areas for future research to enhance the reliability and effectiveness of MLLMs. One promising direction is the development of hybrid prompting strategies, where a combination of few-shot examples and explicit logical structuring could improve reasoning-intensive tasks by guiding models through step-by-step inference. Additionally, memory-augmented models could be explored to enable MLLMs to reference factual information more effectively, reducing the likelihood of hallucinations and improving long-term contextual understanding. Another essential avenue is the advancement of explainability and verification frameworks, particularly for high-stakes applications in legal, medical, and financial domains, where the factual consistency of generated content is crucial.

Furthermore, integrating neurosymbolic AI approaches, which combine deep learning with symbolic reasoning could enhance logical inference capabilities, especially in tasks requiring structured decision-making. In the context of multimodal alignment, research should focus on improving spatial awareness, cross-modal dependencies, and semantic consistency, ensuring that MLLMs can effectively interpret and synthesize diverse inputs. Finally, bias and robustness studies remain indispensable, as the failure of reasoning-based prompting techniques suggests that current architectures may not generalize well to unseen logical structures. Investigating dataset biases and refining training methodologies will be critical to ensuring that MLLMs become more reliable, fair, and interpretable across a wider range of real-world applications. Addressing these challenges through targeted research will be crucial in advancing MLLMs beyond pattern recognition, enabling them to perform more consistent, factually grounded, and contextually aware reasoning in complex decision-making tasks.

Additionally, exploring adaptive prompting strategies and self-correcting mechanisms could provide a pathway toward enhancing MLLMs' generalizability and reliability across diverse domains. This study is a motivation for advancing AI systems from reactive models to proactive, agentic entities \cite{russell2020artificial} capable of sustained, goal-oriented reasoning. As Agentic AI continues to evolve, robust evaluation frameworks such as the one presented in our work will be essential for ensuring that MLLMs are not only technically proficient but also trustworthy, interpretable, and capable of autonomous knowledge synthesis in complex real-world scenarios.

\section{Conclusion}
This study systematically evaluated open-source MLLMs across a diverse scale of model sizes using a structured benchmarking framework to assess their performance across four key evaluation aspects spanning 24 tasks. By employing diverse prompting techniques, including Zero-shot, One-shot, Few-shot, Chain-of-Thought, Analogical reasoning, Generated Knowledge, Tree-of-Thought; the evaluation provided insights into how these models process multimodal inputs and generate outputs aligned with expected task solutions. Our findings highlight the varying effectiveness of different prompting strategies in enhancing MLLMs' interpretability, consistency, and reasoning depth. While some models demonstrated strong performance in certain multimodal translation and cross-modal reasoning tasks, challenges persist in areas requiring deeper contextual understanding, abstraction, and nuanced interpretation of complex inputs. The evaluation underscores the necessity for improved prompt engineering methodologies and more robust benchmarks tailored for multimodal AI. Future work will focus on refining the evaluation criteria, expanding the dataset scope, and integrating real-world application scenarios to further stress-test MLLMs. This includes expanding upon this framework by integrating additional modalities such as video comprehension, auditory processing, and multi-turn interactions to build a more comprehensive evaluation paradigm. Ultimately, these efforts aim to inform the design of next-generation MLLMs that are not only technically proficient but also adaptable, context-aware, and aligned with practical, human-centric applications.

\newpage

\bibliographystyle{unsrt}  
\bibliography{template}  

\begin{thebibliography}{100}

\bibitem{zeng2025scaling}
Hansi Zeng, Julian Killingback, and Hamed Zamani.
\newblock Scaling sparse and dense retrieval in decoder-only llms.
\newblock {\em arXiv preprint arXiv:2502.15526}, 2025.

\bibitem{liu2024llavanext}
Haotian Liu, Chunyuan Li, Yuheng Li, Bo~Li, Yuanhan Zhang, Sheng Shen, and Yong~Jae Lee.
\newblock Llavanext: Improved reasoning, ocr, and world knowledge, 2024.

\bibitem{radford2021learning}
A~Radford, J.W Kim, C~Hallacy, A~Ramesh, G~Goh, S~Agarwal, G~Sastry, A~Askell, P~Mishkin, J~Clark, et~al.
\newblock Learning transferable visual models from natural language supervision.
\newblock {\em Proceedings of the International Conference on Machine Learning}, pages 8748--8763, 2021.

\bibitem{zhai2023sigmoid}
X~Zhai, B~Mustafa, A~Kolesnikov, and L~Beyer.
\newblock Sigmoid loss for language image pre-training.
\newblock {\em Proceedings of the 2023 IEEE/CVF International Conference on Computer Vision (ICCV)}, pages 11941--11952, 2023.

\bibitem{oquab2023dinov2}
Maxime Oquab, Timoth{\'e}e Darcet, Th{\'e}o Moutakanni, Huy Vo, Marc Szafraniec, Vasil Khalidov, Pierre Fernandez, Daniel Haziza, Francisco Massa, Alaaeldin El-Nouby, et~al.
\newblock Dinov2: Learning robust visual features without supervision.
\newblock {\em arXiv preprint arXiv:2304.07193}, 2023.

\bibitem{lu2023reference}
Qinghua Lu, Liming Zhu, Xiwei Xu, Zhenchang Xing, and Jon Whittle.
\newblock A reference architecture for designing foundation model based systems.
\newblock {\em arXiv preprint arXiv:2304.11090}, 2023.

\bibitem{liu2023pre}
Pengfei Liu, Weizhe Yuan, Jinlan Fu, Zhengbao Jiang, Hiroaki Hayashi, and Graham Neubig.
\newblock Pre-train, prompt, and predict: A systematic survey of prompting methods in natural language processing.
\newblock {\em ACM computing surveys}, 55(9):1--35, 2023.

\bibitem{arif2025fixing}
Kazi Hasan~Ibn Arif, Sajib~Acharjee Dip, Khizar Hussain, Lang Zhang, and Chris Thomas.
\newblock Fixing imbalanced attention to mitigate in-context hallucination of large vision-language model.
\newblock {\em arXiv preprint arXiv:2501.12206}, 2025.

\bibitem{vaswani2023attentionneed}
Ashish Vaswani, Noam Shazeer, Niki Parmar, Jakob Uszkoreit, Llion Jones, Aidan~N. Gomez, Lukasz Kaiser, and Illia Polosukhin.
\newblock Attention is all you need, 2023.

\bibitem{zhang2024mmllmsrecentadvancesmultimodal}
Duzhen Zhang, Yahan Yu, Jiahua Dong, Chenxing Li, Dan Su, Chenhui Chu, and Dong Yu.
\newblock Mm-llms: Recent advances in multimodal large language models, 2024.

\bibitem{dosovitskiy2020image}
Alexey Dosovitskiy, Lucas Beyer, Alexander Kolesnikov, Dirk Weissenborn, Xiaohua Zhai, Thomas Unterthiner, Mostafa Dehghani, Matthias Minderer, Georg Heigold, Sylvain Gelly, et~al.
\newblock An image is worth 16x16 words: Transformers for image recognition at scale.
\newblock {\em arXiv preprint arXiv:2010.11929}, 2020.

\bibitem{bao2021beit}
Hangbo Bao, Li~Dong, Songhao Piao, and Furu Wei.
\newblock Beit: Bert pre-training of image transformers.
\newblock {\em arXiv preprint arXiv:2106.08254}, 2021.

\bibitem{duan2024vlmevalkit}
Haodong Duan, Junming Yang, Yuxuan Qiao, Xinyu Fang, Lin Chen, Yuan Liu, Xiaoyi Dong, Yuhang Zang, Pan Zhang, Jiaqi Wang, et~al.
\newblock Vlmevalkit: An open-source toolkit for evaluating large multi-modality models.
\newblock In {\em Proceedings of the 32nd ACM International Conference on Multimedia}, pages 11198--11201, 2024.

\bibitem{ghosh2024exploringfrontiervisionlanguagemodels}
Akash Ghosh, Arkadeep Acharya, Sriparna Saha, Vinija Jain, and Aman Chadha.
\newblock Exploring the frontier of vision-language models: A survey of current methodologies and future directions, 2024.

\bibitem{Yin_2024}
Shukang Yin, Chaoyou Fu, Sirui Zhao, Ke~Li, Xing Sun, Tong Xu, and Enhong Chen.
\newblock A survey on multimodal large language models.
\newblock {\em National Science Review}, 11(12), November 2024.

\bibitem{niu2024textmultimodalityexploringevolution}
Qian Niu, Keyu Chen, Ming Li, Pohsun Feng, Ziqian Bi, Lawrence~KQ Yan, Yichao Zhang, Caitlyn~Heqi Yin, Cheng Fei, Junyu Liu, Benji Peng, Tianyang Wang, Yunze Wang, Silin Chen, and Ming Liu.
\newblock From text to multimodality: Exploring the evolution and impact of large language models in medical practice, 2024.

\bibitem{wang2023finvisgptmultimodallargelanguage}
Ziao Wang, Yuhang Li, Junda Wu, Jaehyeon Soon, and Xiaofeng Zhang.
\newblock Finvis-gpt: A multimodal large language model for financial chart analysis, 2023.

\bibitem{Liang2024.09.29.615524}
Youwei Liang, Ruiyi Zhang, Yongce Li, Mingjia Huo, Zinnia Ma, Digvijay Singh, Chengzhan Gao, Hamidreza Rahmani, Satvik Bandi, Li~Zhang, Robert Weinreb, Atul Malhotra, Danielle~A. Grotjahn, Linda Awdishu, Trey Ideker, Michael Gilson, and Pengtao Xie.
\newblock Multi-modal large language model enables all-purpose prediction of drug mechanisms and properties.
\newblock {\em bioRxiv}, 2024.

\bibitem{Bewersdorff_2025}
Arne Bewersdorff, Christian Hartmann, Marie Hornberger, Kathrin Seßler, Maria Bannert, Enkelejda Kasneci, Gjergji Kasneci, Xiaoming Zhai, and Claudia Nerdel.
\newblock Taking the next step with generative artificial intelligence: The transformative role of multimodal large language models in science education.
\newblock {\em Learning and Individual Differences}, 118:102601, February 2025.

\bibitem{yang2024seedstorymultimodallongstory}
Shuai Yang, Yuying Ge, Yang Li, Yukang Chen, Yixiao Ge, Ying Shan, and Yingcong Chen.
\newblock Seed-story: Multimodal long story generation with large language model, 2024.

\bibitem{kojima2022large}
Takeshi Kojima, Shixiang~Shane Gu, Machel Reid, Yutaka Matsuo, and Yusuke Iwasawa.
\newblock Large language models are zero-shot reasoners.
\newblock {\em Advances in neural information processing systems}, 35:22199--22213, 2022.

\bibitem{wu2020visual}
Bichen Wu, Chenfeng Xu, Xiaoliang Dai, Alvin Wan, Peizhao Zhang, Zhicheng Yan, Masayoshi Tomizuka, Joseph Gonzalez, Kurt Keutzer, and Peter Vajda.
\newblock Visual transformers: Token-based image representation and processing for computer vision.
\newblock {\em arXiv preprint arXiv:2006.03677}, 2020.

\bibitem{xiao2021early}
Tete Xiao, Mannat Singh, Eric Mintun, Trevor Darrell, Piotr Doll{\'a}r, and Ross Girshick.
\newblock Early convolutions help transformers see better.
\newblock {\em Advances in neural information processing systems}, 34:30392--30400, 2021.

\bibitem{guan2024hallusionbench}
Tianrui Guan, Fuxiao Liu, Xiyang Wu, Ruiqi Xian, Zongxia Li, Xiaoyu Liu, Xijun Wang, Lichang Chen, Furong Huang, Yaser Yacoob, et~al.
\newblock Hallusionbench: an advanced diagnostic suite for entangled language hallucination and visual illusion in large vision-language models.
\newblock In {\em Proceedings of the IEEE/CVF Conference on Computer Vision and Pattern Recognition}, pages 14375--14385, 2024.

\bibitem{xie2024tpevaltapmultimodalllms}
Yuxuan Xie, Tianhua Li, Wenqi Shao, and Kaipeng Zhang.
\newblock Tp-eval: Tap multimodal llms' potential in evaluation by customizing prompts, 2024.

\bibitem{brown2020language}
Tom Brown, Benjamin Mann, Nick Ryder, Melanie Subbiah, Jared~D Kaplan, Prafulla Dhariwal, Arvind Neelakantan, Pranav Shyam, Girish Sastry, Amanda Askell, et~al.
\newblock Language models are few-shot learners.
\newblock {\em Advances in neural information processing systems}, 33:1877--1901, 2020.

\bibitem{sivarajkumar2024empirical}
Sonish Sivarajkumar, Mark Kelley, Alyssa Samolyk-Mazzanti, Shyam Visweswaran, and Yanshan Wang.
\newblock An empirical evaluation of prompting strategies for large language models in zero-shot clinical natural language processing: algorithm development and validation study.
\newblock {\em JMIR Medical Informatics}, 12:e55318, 2024.

\bibitem{hao2025mllmsreasonmultimodalityemma}
Yunzhuo Hao, Jiawei Gu, Huichen~Will Wang, Linjie Li, Zhengyuan Yang, Lijuan Wang, and Yu~Cheng.
\newblock Can mllms reason in multimodality? emma: An enhanced multimodal reasoning benchmark, 2025.

\bibitem{jaech2024openai}
Aaron Jaech, Adam Kalai, Adam Lerer, Adam Richardson, Ahmed El-Kishky, Aiden Low, Alec Helyar, Aleksander Madry, Alex Beutel, Alex Carney, et~al.
\newblock Openai o1 system card.
\newblock {\em arXiv preprint arXiv:2412.16720}, 2024.

\bibitem{ge2023chain}
Jiaxin Ge, Hongyin Luo, Siyuan Qian, Yulu Gan, Jie Fu, and Shanghang Zhang.
\newblock Chain of thought prompt tuning in vision language models.
\newblock {\em arXiv preprint arXiv:2304.07919}, 2023.

\bibitem{deepseek2025deepseek}
Daya~Guo DeepSeek-AI, Dejian Yang, Haowei Zhang, Junxiao Song, Ruoyu Zhang, Runxin Xu, Qihao Zhu, Shirong Ma, Peiyi Wang, Xiao Bi, et~al.
\newblock Deepseek-r1: Incentivizing reasoning capability in llms via reinforcement learning.
\newblock {\em arXiv preprint arXiv:2501.12948}, 2025.

\bibitem{Jiang_2024}
Yao Jiang, Xinyu Yan, Ge-Peng Ji, Keren Fu, Meijun Sun, Huan Xiong, Deng-Ping Fan, and Fahad~Shahbaz Khan.
\newblock Effectiveness assessment of recent large vision-language models.
\newblock {\em Visual Intelligence}, 2(1), June 2024.

\bibitem{chen2023minigpt}
Jun Chen, Deyao Zhu, Xiaoqian Shen, Xiang Li, Zechun Liu, Pengchuan Zhang, Raghuraman Krishnamoorthi, Vikas Chandra, Yunyang Xiong, and Mohamed Elhoseiny.
\newblock Minigpt-v2: large language model as a unified interface for vision-language multi-task learning.
\newblock {\em arXiv preprint arXiv:2310.09478}, 2023.

\bibitem{liu2024improved}
Haotian Liu, Chunyuan Li, Yuheng Li, and Yong~Jae Lee.
\newblock Improved baselines with visual instruction tuning.
\newblock In {\em Proceedings of the IEEE/CVF Conference on Computer Vision and Pattern Recognition}, pages 26296--26306, 2024.

\bibitem{chen2023shikra}
Keqin Chen, Zhao Zhang, Weili Zeng, Richong Zhang, Feng Zhu, and Rui Zhao.
\newblock Shikra: Unleashing multimodal llm's referential dialogue magic.
\newblock {\em arXiv preprint arXiv:2306.15195}, 2023.

\bibitem{OpenAI2025Models}
OpenAI.
\newblock Openai models documentation, 2025.
\newblock Accessed: 25 February 2025.

\bibitem{li2025structured}
Jia Li, Ge~Li, Yongmin Li, and Zhi Jin.
\newblock Structured chain-of-thought prompting for code generation.
\newblock {\em ACM Transactions on Software Engineering and Methodology}, 34(2):1--23, 2025.

\bibitem{li2024llava}
B~Li, Y~Zhang, D~Guo, R~Zhang, F~Li, H~Zhang, K~Zhang, P~Zhang, Y~Li, Z~Liu, et~al.
\newblock Llava-onevision: Easy visual task transfer.
\newblock {\em arXiv preprint arXiv:2408.03326}, 2024.

\bibitem{wu2025controlmllm}
Mingrui Wu, Xinyue Cai, Jiayi Ji, Jiale Li, Oucheng Huang, Gen Luo, Hao Fei, Guannan Jiang, Xiaoshuai Sun, and Rongrong Ji.
\newblock Controlmllm: Training-free visual prompt learning for multimodal large language models.
\newblock {\em Advances in Neural Information Processing Systems}, 37:45206--45234, 2025.

\bibitem{yu2023mmvet}
Weihao Yu, Zhengyuan Yang, Linjie Li, Jianfeng Wang, Kevin Lin, Zicheng Liu, Xinchao Wang, and Lijuan Wang.
\newblock Mm-vet: Evaluating large multimodal models for integrated capabilities.
\newblock {\em arXiv preprint arXiv:2308.02490}, 2023.

\bibitem{nwae403}
Shukang Yin, Chaoyou Fu, Sirui Zhao, Ke~Li, Xing Sun, Tong Xu, and Enhong Chen.
\newblock A survey on multimodal large language models.
\newblock {\em National Science Review}, 11(12):nwae403, 11 2024.

\bibitem{anthropic2024claude}
AI~Anthropic.
\newblock The claude 3 model family: Opus, sonnet, haiku.
\newblock {\em Claude-3 Model Card}, 1:1, 2024.

\bibitem{DeepMind2025Gemini}
DeepMind.
\newblock Gemini: Multimodal ai by deepmind, 2025.
\newblock Accessed: 25 February 2025.

\bibitem{meta2024llama3}
Meta.
\newblock Build the future of ai with meta llama 3.
\newblock {\em Technical Report}, 2024.
\newblock accessed on 13 February 2025.

\bibitem{mistral2024pixtral}
MistralAI.
\newblock Pixtral-12b-2409.
\newblock {\em Available online}, 2024.
\newblock accessed on 13 February 2025.

\bibitem{malartic2024falcon2}
Quentin Malartic, Nilabhra~Roy Chowdhury, Ruxandra Cojocaru, Mugariya Farooq, Giulia Campesan, Yasser Abdelaziz~Dahou Djilali, Sanath Narayan, Ankit Singh, Maksim Velikanov, Basma El~Amel Boussaha, et~al.
\newblock Falcon2-11b technical report.
\newblock {\em arXiv preprint arXiv:2407.14885}, 2024.

\bibitem{fu2024vita}
Chaoyou Fu, Haojia Lin, Zuwei Long, Yunhang Shen, Meng Zhao, Yifan Zhang, Shaoqi Dong, Xiong Wang, Di~Yin, Long Ma, et~al.
\newblock Vita: Towards open-source interactive omni multimodal llm.
\newblock {\em arXiv preprint arXiv:2408.05211}, 2024.

\bibitem{shen2025long}
Yunhang Shen, Chaoyou Fu, Shaoqi Dong, Xiong Wang, Peixian Chen, Mengdan Zhang, Haoyu Cao, Ke~Li, Xiawu Zheng, Yan Zhang, et~al.
\newblock Long-vita: Scaling large multi-modal models to 1 million tokens with leading short-context accuray.
\newblock {\em arXiv preprint arXiv:2502.05177}, 2025.

\bibitem{ye2024mplug}
Jiabo Ye, Haiyang Xu, Haowei Liu, Anwen Hu, Ming Yan, Qi~Qian, Ji~Zhang, Fei Huang, and Jingren Zhou.
\newblock mplug-owl3: Towards long image-sequence understanding in multi-modal large language models.
\newblock In {\em The Thirteenth International Conference on Learning Representations}, 2024.

\bibitem{lee2024moai}
Byung-Kwan Lee, Beomchan Park, Chae Won~Kim, and Yong Man~Ro.
\newblock Moai: Mixture of all intelligence for large language and vision models.
\newblock In {\em European Conference on Computer Vision}, pages 273--302. Springer, 2024.

\bibitem{jiang2024chatrex}
Qing Jiang, Yuqin Yang, Yuda Xiong, Yihao Chen, Zhaoyang Zeng, Tianhe Ren, Lei Zhang, et~al.
\newblock Chatrex: Taming multimodal llm for joint perception and understanding.
\newblock {\em arXiv preprint arXiv:2411.18363}, 2024.

\bibitem{cai2024vip}
Mu~Cai, Haotian Liu, Siva~Karthik Mustikovela, Gregory~P Meyer, Yuning Chai, Dennis Park, and Yong~Jae Lee.
\newblock Vip-llava: Making large multimodal models understand arbitrary visual prompts.
\newblock In {\em Proceedings of the IEEE/CVF Conference on Computer Vision and Pattern Recognition}, pages 12914--12923, 2024.

\bibitem{deitke2024molmo}
Matt Deitke, Christopher Clark, Sangho Lee, Rohun Tripathi, Yue Yang, Jae~Sung Park, Mohammadreza Salehi, Niklas Muennighoff, Kyle Lo, Luca Soldaini, et~al.
\newblock Molmo and pixmo: Open weights and open data for state-of-the-art multimodal models.
\newblock {\em arXiv preprint arXiv:2409.17146}, 2024.

\bibitem{tong2025cambrian}
Peter Tong, Ellis Brown, Penghao Wu, Sanghyun Woo, Adithya Jairam~Vedagiri IYER, Sai~Charitha Akula, Shusheng Yang, Jihan Yang, Manoj Middepogu, Ziteng Wang, et~al.
\newblock Cambrian-1: A fully open, vision-centric exploration of multimodal llms.
\newblock {\em Advances in Neural Information Processing Systems}, 37:87310--87356, 2025.

\bibitem{yao2024minicpm}
Yuan Yao, Tianyu Yu, Ao~Zhang, Chongyi Wang, Junbo Cui, Hongji Zhu, Tianchi Cai, Haoyu Li, Weilin Zhao, Zhihui He, et~al.
\newblock Minicpm-v: A gpt-4v level mllm on your phone.
\newblock {\em arXiv preprint arXiv:2408.01800}, 2024.

\bibitem{lee2025meteor}
Byung-Kwan Lee, Chae~Won Kim, Beomchan Park, and Yong~Man Ro.
\newblock Meteor: Mamba-based traversal of rationale for large language and vision models.
\newblock {\em Advances in Neural Information Processing Systems}, 37:40278--40315, 2025.

\bibitem{yang2023mm}
Xiaocui Yang, Wenfang Wu, Shi Feng, Ming Wang, Daling Wang, Yang Li, Qi~Sun, Yifei Zhang, Xiaoming Fu, and Soujanya Poria.
\newblock Mm-bigbench: Evaluating multimodal models on multimodal content comprehension tasks.
\newblock {\em arXiv preprint arXiv:2310.09036}, 2023.

\bibitem{bhattacharjya2024foundation}
D~Bhattacharjya, J~Lee, D.J Agravante, B~Ganesan, and R~Marinescu.
\newblock Foundation model sherpas: Guiding foundation models through knowledge and reasoning.
\newblock {\em arXiv preprint arXiv:2402.01602}, 2024.

\bibitem{jiang2022promptmaker}
E~Jiang, K~Olson, E~Toh, A~Molina, A~Donsbach, M~Terry, and C.J Cai.
\newblock Promptmaker: Prompt-based prototyping with large language models.
\newblock {\em CHI Conference on Human Factors in Computing Systems Extended Abstracts}, pages 1--8, 2022.

\bibitem{zamfirescu2023johnny}
J.D Zamfirescu-Pereira, R.Y Wong, B~Hartmann, and Q~Yang.
\newblock Why johnny can’t prompt: How non-ai experts try (and fail) to design llm prompts.
\newblock {\em Proceedings of the 2023 CHI Conference on Human Factors in Computing Systems}, pages 1--21, 2023.

\bibitem{mann2020language}
Ben Mann, N~Ryder, M~Subbiah, J~Kaplan, P~Dhariwal, A~Neelakantan, P~Shyam, G~Sastry, A~Askell, S~Agarwal, et~al.
\newblock Language models are few-shot learners.
\newblock {\em arXiv preprint arXiv:2005.14165}, 1, 2020.

\bibitem{wei2022chain}
Jason Wei, Xuezhi Wang, Dale Schuurmans, Maarten Bosma, Fei Xia, Ed~Chi, Quoc~V Le, Denny Zhou, et~al.
\newblock Chain-of-thought prompting elicits reasoning in large language models.
\newblock {\em Advances in neural information processing systems}, 35:24824--24837, 2022.

\bibitem{yao2024tree}
Shunyu Yao, Dian Yu, Jeffrey Zhao, Izhak Shafran, Tom Griffiths, Yuan Cao, and Karthik Narasimhan.
\newblock Tree of thoughts: Deliberate problem solving with large language models.
\newblock {\em Advances in Neural Information Processing Systems}, 36, 2024.

\bibitem{zhang2022automatic}
Zhuosheng Zhang, Aston Zhang, Mu~Li, and Alex Smola.
\newblock Automatic chain of thought prompting in large language models.
\newblock {\em arXiv preprint arXiv:2210.03493}, 2022.

\bibitem{rose2023visual}
Daniel Rose, Vaishnavi Himakunthala, Andy Ouyang, Ryan He, Alex Mei, Yujie Lu, Michael Saxon, Chinmay Sonar, Diba Mirza, and William~Yang Wang.
\newblock Visual chain of thought: bridging logical gaps with multimodal infillings.
\newblock {\em arXiv preprint arXiv:2305.02317}, 2023.

\bibitem{zhang2023multimodal}
Zhuosheng Zhang, Aston Zhang, Mu~Li, Hai Zhao, George Karypis, and Alex Smola.
\newblock Multimodal chain-of-thought reasoning in language models.
\newblock {\em arXiv preprint arXiv:2302.00923}, 2023.

\bibitem{yasunaga2023large}
Michihiro Yasunaga, Xinyun Chen, Yujia Li, Panupong Pasupat, Jure Leskovec, Percy Liang, Ed~H Chi, and Denny Zhou.
\newblock Large language models as analogical reasoners.
\newblock {\em arXiv preprint arXiv:2310.01714}, 2023.

\bibitem{liu2021generated}
Jiacheng Liu, Alisa Liu, Ximing Lu, Sean Welleck, Peter West, Ronan~Le Bras, Yejin Choi, and Hannaneh Hajishirzi.
\newblock Generated knowledge prompting for commonsense reasoning.
\newblock {\em arXiv preprint arXiv:2110.08387}, 2021.

\bibitem{radford2019language}
Alec Radford, Jeffrey Wu, Rewon Child, David Luan, Dario Amodei, Ilya Sutskever, et~al.
\newblock Language models are unsupervised multitask learners.
\newblock {\em OpenAI blog}, 1(8):9, 2019.

\bibitem{jiangmmad}
Xi~Jiang, Jian Li, Hanqiu Deng, Yong Liu, Bin-Bin Gao, Yifeng Zhou, Jialin Li, Chengjie Wang, and Feng Zheng.
\newblock Mmad: A comprehensive benchmark for multimodal large language models in industrial anomaly detection.
\newblock In {\em The Thirteenth International Conference on Learning Representations}, 2025.

\bibitem{huang2023language}
Shaohan Huang, Li~Dong, Wenhui Wang, Yaru Hao, Saksham Singhal, Shuming Ma, Tengchao Lv, Lei Cui, Owais~Khan Mohammed, Barun Patra, et~al.
\newblock Language is not all you need: Aligning perception with language models.
\newblock {\em Advances in Neural Information Processing Systems}, 36:72096--72109, 2023.

\bibitem{zhang2024mme}
Yi-Fan Zhang, Huanyu Zhang, Haochen Tian, Chaoyou Fu, Shuangqing Zhang, Junfei Wu, Feng Li, Kun Wang, Qingsong Wen, Zhang Zhang, et~al.
\newblock Mme-realworld: Could your multimodal llm challenge high-resolution real-world scenarios that are difficult for humans?
\newblock {\em arXiv preprint arXiv:2408.13257}, 2024.

\bibitem{adler2024gpt}
Steven Adler, Sandhini Agarwal, Lama Ahmad, Ilge Akkaya, Florencia~Leoni Aleman, Diogo Almeida, Janko Altenschmidt, Sam Altman, Shyamal Anadkat, Red Avila, et~al.
\newblock Gpt-4 technical report, 2024.
\newblock {\em URL https://arxiv. org/abs/2303.08774}, 2:6, 2024.

\bibitem{meta2024llama32}
Meta.
\newblock Llama 3.2-11b vision.
\newblock {\em Available online}, 2024.
\newblock accessed on 13 February 2025.

\bibitem{vectara_hallucination_leaderboard}
Vectara.
\newblock Hallucination leaderboard.
\newblock GitHub repository, 2024.

\bibitem{galileo_hallucination_index}
Galileo AI.
\newblock The hallucination index: A new paradigm for evaluating ai generated content.
\newblock Galileo AI Blog, 2024.

\bibitem{russell2020artificial}
Stuart Russell and Peter Norvig.
\newblock Artificial intelligence: a modern approach. hoboken, 2020.

\bibitem{chen2024internvl}
Z~Chen, J~Wu, W~Wang, W~Su, G~Chen, S~Xing, M~Zhong, Q~Zhang, X~Zhu, L~Lu, et~al.
\newblock Internvl: Scaling up vision foundation models and aligning for generic visual-linguistic tasks.
\newblock {\em Proceedings of the IEEE/CVF Conference on Computer Vision and Pattern Recognition}, pages 24185--24198, 2024.

\bibitem{fu2024mme}
Chaoyou Fu, Yi-Fan Zhang, Shukang Yin, Bo~Li, Xinyu Fang, Sirui Zhao, Haodong Duan, Xing Sun, Ziwei Liu, Liang Wang, et~al.
\newblock Mme-survey: A comprehensive survey on evaluation of multimodal llms.
\newblock {\em arXiv preprint arXiv:2411.15296}, 2024.

\bibitem{wang2024exploring}
Yiqi Wang, Wentao Chen, Xiaotian Han, Xudong Lin, Haiteng Zhao, Yongfei Liu, Bohan Zhai, Jianbo Yuan, Quanzeng You, and Hongxia Yang.
\newblock Exploring the reasoning abilities of multimodal large language models (mllms): A comprehensive survey on emerging trends in multimodal reasoning.
\newblock {\em arXiv preprint arXiv:2401.06805}, 2024.

\bibitem{tong2024eyes}
Shengbang Tong, Zhuang Liu, Yuexiang Zhai, Yi~Ma, Yann LeCun, and Saining Xie.
\newblock Eyes wide shut? exploring the visual shortcomings of multimodal llms.
\newblock In {\em Proceedings of the IEEE/CVF Conference on Computer Vision and Pattern Recognition}, pages 9568--9578, 2024.

\bibitem{zhou2025aretheythesame}
Yikang Zhou, Tao Zhang, Shilin Xu, Shihao Chen, Qianyu Zhou, Yunhai Tong, Shunping Ji, Jiangning Zhang, Xiangtai Li, and Lu~Qi.
\newblock Are they the same? exploring visual correspondence shortcomings of multimodal llms.
\newblock {\em arXiv preprint arXiv:2501.04670}, 2025.

\bibitem{li2024surveying}
Ming Li, Keyu Chen, Ziqian Bi, Ming Liu, Benji Peng, Qian Niu, Junyu Liu, Jinlang Wang, Sen Zhang, Xuanhe Pan, et~al.
\newblock Surveying the mllm landscape: A meta-review of current surveys.
\newblock {\em arXiv preprint arXiv:2409.18991}, 2024.

\bibitem{wang2024comprehensive}
Jiaqi Wang, Hanqi Jiang, Yiheng Liu, Chong Ma, Xu~Zhang, Yi~Pan, Mengyuan Liu, Peiran Gu, Sichen Xia, Wenjun Li, et~al.
\newblock A comprehensive review of multimodal large language models: Performance and challenges across different tasks.
\newblock {\em arXiv preprint arXiv:2408.01319}, 2024.

\bibitem{zhang2024generalist}
Kai Zhang, Rong Zhou, Eashan Adhikarla, Zhiling Yan, Yixin Liu, Jun Yu, Zhengliang Liu, Xun Chen, Brian~D Davison, Hui Ren, et~al.
\newblock A generalist vision--language foundation model for diverse biomedical tasks.
\newblock {\em Nature Medicine}, pages 1--13, 2024.

\bibitem{royer2024multimedeval}
Corentin Royer, Bjoern Menze, and Anjany Sekuboyina.
\newblock Multimedeval: A benchmark and a toolkit for evaluating medical vision-language models.
\newblock {\em arXiv preprint arXiv:2402.09262}, 2024.

\bibitem{li2023llava}
Chunyuan Li, Cliff Wong, Sheng Zhang, Naoto Usuyama, Haotian Liu, Jianwei Yang, Tristan Naumann, Hoifung Poon, and Jianfeng Gao.
\newblock Llava-med: Training a large language-and-vision assistant for biomedicine in one day.
\newblock {\em Advances in Neural Information Processing Systems}, 36:28541--28564, 2023.

\bibitem{shi2023exploring}
Yongxin Shi, Dezhi Peng, Wenhui Liao, Zening Lin, Xinhong Chen, Chongyu Liu, Yuyi Zhang, and Lianwen Jin.
\newblock Exploring ocr capabilities of gpt-4v (ision): A quantitative and in-depth evaluation.
\newblock {\em arXiv preprint arXiv:2310.16809}, 2023.

\bibitem{sima2024drivelm}
Chonghao Sima, Katrin Renz, Kashyap Chitta, Li~Chen, Hanxue Zhang, Chengen Xie, Jens Bei{\ss}wenger, Ping Luo, Andreas Geiger, and Hongyang Li.
\newblock Drivelm: Driving with graph visual question answering.
\newblock In {\em European Conference on Computer Vision}, pages 256--274. Springer, 2024.

\bibitem{qian2024nuscenes}
Tianwen Qian, Jingjing Chen, Linhai Zhuo, Yang Jiao, and Yu-Gang Jiang.
\newblock Nuscenes-qa: A multi-modal visual question answering benchmark for autonomous driving scenario.
\newblock In {\em Proceedings of the AAAI Conference on Artificial Intelligence}, volume~38, pages 4542--4550, 2024.

\bibitem{kim2018textual}
Jinkyu Kim, Anna Rohrbach, Trevor Darrell, John Canny, and Zeynep Akata.
\newblock Textual explanations for self-driving vehicles.
\newblock In {\em Proceedings of the European conference on computer vision (ECCV)}, pages 563--578, 2018.

\bibitem{zhang2024benchmarking}
Yichi Zhang, Yao Huang, Yitong Sun, Chang Liu, Zhe Zhao, Zhengwei Fang, Yifan Wang, Huanran Chen, Xiao Yang, Xingxing Wei, et~al.
\newblock Benchmarking trustworthiness of multimodal large language models: A comprehensive study.
\newblock {\em arXiv preprint arXiv:2406.07057}, 2024.

\bibitem{li2024red}
Mukai Li, Lei Li, Yuwei Yin, Masood Ahmed, Zhenguang Liu, and Qi~Liu.
\newblock Red teaming visual language models.
\newblock {\em arXiv preprint arXiv:2401.12915}, 2024.

\bibitem{wang2024vlbiasbench}
Sibo Wang, Xiangkui Cao, Jie Zhang, Zheng Yuan, Shiguang Shan, Xilin Chen, and Wen Gao.
\newblock Vlbiasbench: A comprehensive benchmark for evaluating bias in large vision-language model.
\newblock {\em arXiv preprint arXiv:2406.14194}, 2024.

\bibitem{dang2024exploring}
Yunkai Dang, Mengxi Gao, Yibo Yan, Xin Zou, Yanggan Gu, Aiwei Liu, and Xuming Hu.
\newblock Exploring response uncertainty in mllms: An empirical evaluation under misleading scenarios.
\newblock {\em arXiv preprint arXiv:2411.02708}, 2024.

\bibitem{bai2024hallucination}
Zechen Bai, Pichao Wang, Tianjun Xiao, Tong He, Zongbo Han, Zheng Zhang, and Mike~Zheng Shou.
\newblock Hallucination of multimodal large language models: A survey.
\newblock {\em arXiv preprint arXiv:2404.18930}, 2024.

\bibitem{yin2024woodpecker}
Shukang Yin, Chaoyou Fu, Sirui Zhao, Tong Xu, Hao Wang, Dianbo Sui, Yunhang Shen, Ke~Li, Xing Sun, and Enhong Chen.
\newblock Woodpecker: Hallucination correction for multimodal large language models.
\newblock {\em Science China Information Sciences}, 67(12):220105, 2024.

\bibitem{yang2024qwen2}
A~Yang, B~Yang, B~Zhang, B~Hui, B~Zheng, B~Yu, C~Li, D~Liu, F~Huang, H~Wei, et~al.
\newblock Qwen2.5 technical report.
\newblock {\em arXiv preprint arXiv:2412.15115}, 2024.

\bibitem{huang2024mini}
M~Huang, Y~Liu, D~Liang, L~Jin, and X~Bai.
\newblock Mini-monkey: Multi-scale adaptive cropping for multimodal large language models.
\newblock {\em CoRR}, 2024.

\bibitem{beyer2024paligemma}
L~Beyer, A~Steiner, A.S Pinto, A~Kolesnikov, X~Wang, D~Salz, M~Neumann, I~Alabdulmohsin, M~Tschannen, E~Bugliarello, et~al.
\newblock Paligemma: A versatile 3b vlm for transfer.
\newblock {\em arXiv preprint arXiv:2407.07726}, 2024.

\bibitem{abdin2024phi}
M~Abdin, J~Aneja, H~Awadalla, A~Awadallah, A.A Awan, N~Bach, A~Bahree, A~Bakhtiari, J~Bao, H~Behl, et~al.
\newblock Phi-3 technical report: A highly capable language model locally on your phone.
\newblock {\em arXiv preprint arXiv:2404.14219}, 2024.

\bibitem{lu2024ovis}
S~Lu, Y~Li, Q.G Chen, Z~Xu, W~Luo, K~Zhang, and H.J Ye.
\newblock Ovis: Structural embedding alignment for multimodal large language model.
\newblock {\em arXiv preprint arXiv:2405.20797}, 2024.

\bibitem{glm2024chatglm}
T~GLM, A~Zeng, B~Xu, B~Wang, C~Zhang, D~Yin, D~Zhang, D~Rojas, G~Feng, H~Zhao, et~al.
\newblock Chatglm: A family of large language models from glm-130b to glm-4 all tools.
\newblock {\em arXiv preprint arXiv:2406.12793}, 2024.

\bibitem{agrawal2024pixtral}
P~Agrawal, S~Antoniak, E.B Hanna, B~Bout, D~Chaplot, J~Chudnovsky, D~Costa, B~De~Monicault, S~Garg, T~Gervet, et~al.
\newblock Pixtral 12b.
\newblock {\em arXiv preprint arXiv:2410.07073}, 2024.

\bibitem{omchat2024v2}
OmLab.
\newblock Omchat v2.0-13b single beta.
\newblock {\em Available online}, 2024.
\newblock accessed on 13 February 2025.

\bibitem{chen2024far}
Z~Chen, W~Wang, H~Tian, S~Ye, Z~Gao, E~Cui, W~Tong, K~Hu, J~Luo, Z~Ma, et~al.
\newblock How far are we to gpt-4v? closing the gap to commercial multimodal models with open-source suites.
\newblock {\em Science China Information Sciences}, 67:220101, 2024.

\bibitem{internvl2024}
InternVL.
\newblock Internvl 2.0 blog post.
\newblock {\em Available online}, 2024.
\newblock accessed on [insert date].

\bibitem{lu2021fantastically}
Yao Lu, Max Bartolo, Alastair Moore, Sebastian Riedel, and Pontus Stenetorp.
\newblock Fantastically ordered prompts and where to find them: Overcoming few-shot prompt order sensitivity.
\newblock {\em arXiv preprint arXiv:2104.08786}, 2021.

\bibitem{wu2022self}
Zhiyong Wu, Yaoxiang Wang, Jiacheng Ye, and Lingpeng Kong.
\newblock Self-adaptive in-context learning: An information compression perspective for in-context example selection and ordering.
\newblock {\em arXiv preprint arXiv:2212.10375}, 2022.

\bibitem{guo2024can}
Diandian Guo, Cong Cao, Fangfang Yuan, Dakui Wang, Wei Ma, Yanbing Liu, and Jianhui Fu.
\newblock Can multimodal large language model think analogically?
\newblock {\em arXiv preprint arXiv:2411.01307}, 2024.

\end{thebibliography}

\appendix
\appendix
\newpage

\section*{Appendix}

\section{Literature Review on MLLM Evaluation and Rationale for Selected Evaluation Aspects} \label{rationale-for-selected-EAs}

Evaluating MLLMs is a critical task that determines their capabilities, limitations, and applicability across various domains. Various studies \cite{li2024llava, wu2025controlmllm, yu2023mmvet} have employed a wide range of evaluation aspects depending on their focus, whether it be perceptual understanding, compositional reasoning, multimodal alignment, or task-specific problem-solving. Unlike traditional unimodal models, MLLMs require evaluation frameworks that consider their ability to process, integrate, and generate outputs from different modalities, such as text, images, and, in some cases, audio and video. A review of existing studies \cite{nwae403} shows that evaluations are primarily categorized into general multimodal understanding, task-specific assessments, and trustworthiness metrics. These evaluations are particularly important for ensuring that MLLMs can transition from general-purpose reasoning to task-specific applications, making them more reliable for real-world deployments. This section discusses commonly used evaluation aspects in MLLM research and motivated us to select the evaluation aspect those have wider impact in various applications in MLLM models. 

Various studies \cite{cai2024vip, chen2024internvl, li2024llava, wu2025controlmllm, yu2023mmvet} have approached evaluation through distinct lenses, focusing on foundational multimodal capabilities and task-specific applications. One of the primary evaluation aspects in existing research is perception and object recognition, where models are tested on their ability to identify objects, classify images, and interpret visual attributes. Benchmarks such as VQA v2, COCO, and ImageNet have been widely used to assess how well models recognize and describe elements within an image, ensuring that they correctly map visual features to textual outputs \cite{fu2024mme}. Similarly, scene understanding is another crucial dimension that determines whether models can accurately describe complex environments, capturing spatial relationships and multiple interacting objects, as tested in datasets such as GQA and VizWiz \cite{nwae403}.

Another key area of evaluation in existing studies is multimodal reasoning and alignment, which assesses how well MLLMs integrate information across different modalities to derive meaningful conclusions \cite{wang2024exploring}. The ability to perform multimodal chain-of-thought (CoT) reasoning has gained increasing attention, as models must generate structured responses by logically sequencing multiple inputs from different modalities\cite{nwae403}. Studies \cite{tong2024eyes, zhou2025aretheythesame} have found that current MLLMs, despite their strong textual reasoning abilities, still struggle to connect visual cues with textual prompts in a coherent manner, limiting their reliability in domains requiring step-by-step logical processing. Additionally, multimodal reasoning is closely tied to vision-language alignment, which evaluates how effectively models fuse textual and visual data to generate accurate and context-aware outputs. Research \cite{fu2024mme} indicates that poor alignment often leads to failures in high-level reasoning tasks, as models may incorrectly interpret visual information when translating it into text-based responses. 

Beyond fundamental perception and reasoning, some studies have investigated task-specific and real-world applications of MLLMs \cite{li2024surveying, wang2024comprehensive}. One of the most prominent applications is medical image analysis, where models are tested on their ability to assist in diagnosing conditions from X-rays, MRIs, and other medical scans \cite{zhang2024generalist, royer2024multimedeval, li2023llava}. In this domain, specialized benchmarks such as RadBench and PMC-VQA have been used to evaluate the accuracy of MLLM-generated medical insights \cite{nwae403}. Another critical application is autonomous driving and environmental scene interpretation \cite{shi2023exploring, sima2024drivelm}, where models process real-world driving scenarios by integrating road signs, obstacles, and pedestrian movements into decision-making processes. Datasets such as NuScenes-QA \cite{qian2024nuscenes} and BDD-X \cite{kim2018textual} serve as standard benchmarks in this space, ensuring that models can reliably analyze traffic conditions and provide accurate assessments of dynamic environments \cite{nwae403}. 

Some studies have explored multi-round QA and instruction following \cite{fu2024mme}, where models are tested on how well they can retain contextual information across conversational turns. Research indicates that existing MLLMs often exhibit context drift, failing to maintain coherence when responding to sequential queries, which is a key challenge in developing effective AI-powered assistants \cite{nwae403}. Furthermore, aspects related to bias and fairness have also been examined in some studies \cite{zhang2024benchmarking, li2024red}, as concerns grow about MLLMs inheriting and amplifying societal biases from their training data. Studies \cite{adler2024gpt, anthropic2024claude} stand out as the most ethically aligned models, showing high accuracy and a strong ability to refuse ethically questionable prompts. Fairness evaluation often involves testing for gender, racial, and regional biases, using datasets such as Multi-Trust \cite{zhang2024benchmarking} and VLBiasBench \cite{wang2024vlbiasbench} to assess whether models generate discriminatory or skewed responses \cite{nwae403}.

Another critical aspect in evaluating MLLMs is their robustness and reliability \cite{dang2024exploring}, particularly in minimizing hallucinations \cite{bai2024hallucination}, a phenomenon where models generate factually incorrect or fabricated outputs. Studies assessing hallucination rates have shown that MLLMs, particularly in complex reasoning tasks, tend to produce confident but incorrect statements, making them unreliable for high-stakes applications such as medical diagnosis and financial forecasting \cite{bai2024hallucination}. Efforts to measure and mitigate hallucinations have led to the development of evaluation frameworks that focus on trustworthiness and factual consistency, ensuring that models generate responses that are not only contextually relevant but also grounded in accurate information \cite{yin2024woodpecker}.

Overall, the breadth of evaluation aspects considered in MLLM studies reflects the increasing complexity of these models and the growing need for rigorous assessment frameworks. While some studies focus on fundamental perception and reasoning abilities, others prioritize task-specific applications, conversational capabilities, and fairness concerns. However, despite the diverse range of evaluation methodologies, challenges remain in developing standardized benchmarks that comprehensively measure MLLM performance across all modalities and tasks. This highlights the importance of selecting well-defined evaluation aspects that balance theoretical rigor and real-world applicability, which is a key motivation behind the selection of four core evaluation aspects in our study.

Despite these advancements, researchers still face challenges in defining standardized, reproducible evaluation aspects. Many existing methods rely on task-specific datasets or closed-set evaluations, limiting the generalizability of their findings. There is an increasing need for comprehensive, multi-domain evaluation frameworks that can assess models across perception, reasoning, task execution, and ethical considerations, ensuring that MLLMs are robust, interpretable, and trustworthy for real-world applications \cite{fu2024mme}.

While audio-visual evaluations are critical for advancing MLLMs, current models still exhibit substantial limitations in handling high-dimensional temporal data, making standardized evaluations difficult\cite{fu2024mme}. Audio and video-based evaluation are not considered in our study, driven by the need for a controlled evaluation framework that minimizes confounding factors related to temporal dependencies, data preprocessing, and model specialization \cite{nwae403}. 

Given the wide range of evaluation aspects discussed in existing research, our study selects four core aspects \ref{modelEAs}: Reasoning and Compositionality, Multimodal Understanding and Alignment, Complex Code Generation and Execution, and Knowledge Retrieval and Integration, due to their direct relevance in assessing task-specific performance and a wide range of real-world applicability and use-cases of MLLMs. 

\section{Detailed Review of Key MLLMs}\label{apdx_review_mllm}

This section provides a comprehensive review of recent advancements in Multimodal Large Language Models (MLLMs), covering both proprietary and open-source developments. The discussion highlights their architectural designs, multimodal processing capabilities, and evaluation suitability, which informed our selection of models for this study.

\subsection{Proprietary MLLMs}

Proprietary models such as OpenAI's GPT-4o and GPT-4.5 Preview \cite{OpenAI2025Models}, Anthropic’s Claude 3 (including variants like Claude 3 Opus, Sonnet, and Haiku) \cite{anthropic2024claude}, and Google's Gemini series \cite{DeepMind2025Gemini} have set high standards in multimodal reasoning. For example, GPT-4 introduced multimodal inputs that enable the processing of both text and images, thereby improving instruction-following and contextual understanding. Similarly, the Claude 3 series is optimized for text–image reasoning tasks, leveraging large-scale datasets and fine-tuned architectures to achieve superior generalization and task flexibility. However, the closed-source nature of these models limits customization, transparency, and independent research adaptability.

\subsection{Open-Source MLLMs}

In response to the limitations of proprietary systems, the open-source ecosystem has rapidly evolved, providing models with enhanced accessibility and community-driven improvements. Several high-performance models have emerged:

\paragraph{ChatRex \cite{jiang2024chatrex}}  
ChatRex is designed as a decoupled MLLM that bridges perception and understanding in vision-language tasks. By employing a retrieval-based approach with a Universal Proposal Network (UPN), it improves visual grounding for tasks like object detection and region-based question answering. While it excels in object-level perception and spatial awareness, its specialization limits its adaptability for abstract multimodal reasoning tasks.

\paragraph{VITA \cite{fu2024vita}}  
VITA is a multimodal interactive model that handles video, image, text, and audio modalities in real time. It extends Mixtral 8×7B to enhance bilingual capabilities and integrates specialized encoders for vision and audio. Its duplex interaction pipeline enables continuous environmental monitoring and multi-turn conversations, though it is less optimized for structured vision-language reasoning such as code synthesis.

\paragraph{Long-VITA \cite{shen2025long}}  
Long-VITA targets long-context visual-language understanding, capable of processing up to 1 million tokens. It uses a multi-stage training process that includes long-sequence fine-tuning to maintain coherence over extended interactions. While it excels in video comprehension and document-level understanding, its lack of strict data filtering can impact response consistency.

\paragraph{Meteor \cite{lee2025meteor}}  
Meteor employs a rationale traversal approach to enhance multimodal reasoning without significantly increasing model size. By leveraging multifaceted rationale embeddings, it improves vision-language understanding and step-by-step reasoning. However, its performance is sensitive to the availability of rationale-enhanced pretraining datasets and may struggle with tasks that require high-resolution image processing.

\paragraph{MoAI \cite{lee2024moai}}  
MoAI utilizes a Mixture of Experts (MoE) approach by integrating external computer vision models for panoptic segmentation, object detection, and OCR. Its two-stage pipeline, comprising a MoAI-Compressor and MoAI-Mixer allows for detailed object detection and structured text reasoning. While effective in high-precision tasks, the dependency on external modules increases computational overhead.

\paragraph{ViP-LLaVA \cite{cai2024vip}}  
ViP-LLaVA enhances visual reasoning by overlaying visual markers (e.g., bounding boxes, arrows) directly onto input images, facilitating region-specific interactions. Built on a Vicuna v1.5 language backbone and CLIP-336px vision encoder, it excels in object localization and context-aware reasoning. Its training process involves multiple stages, including BLIP-captioned image-text pairs and GPT-4V instruction data.

\paragraph{Falcon2-11B \cite{malartic2024falcon2}}  
Falcon2-11B is an efficient foundation model with 11 billion parameters, optimized for text-based tasks and extended to multimodal functions via a CLIP-based vision encoder. Although it exhibits strong long-context reasoning, its vision-language variant may be less competitive in tasks requiring fine-grained visual comprehension.

\paragraph{Cambrian-1 \cite{tong2025cambrian}}  
Cambrian-1 is designed as a vision-centric MLLM using multiple vision encoders (e.g., OpenAI CLIP ViT-L/14@336, SigLIP ViT-SO400M, DINOv2 ViT-L/14) integrated via a Spatial Vision Aggregator (SVA). Available in various parameter scales (8B, 13B, 34B), it excels in OCR and chart-based tasks while maintaining efficiency through reduced visual token usage.

\paragraph{MiniCPM \cite{yao2024minicpm}}  
MiniCPM is a lightweight model focused on efficient processing of vision-language tasks. Despite its compact architecture, it performs competitively on fundamental multimodal reasoning tasks, though its capacity for complex, long-form content is limited.

\paragraph{mPLUG-Owl3 \cite{ye2024mplug}}  
mPLUG-Owl3 specializes in long image-sequence understanding and video-based multimodal processing using Hyper Attention Transformer Blocks (HATB). It is tailored for tasks requiring temporal coherence but is less suited for structured code generation and fine-tuned knowledge retrieval.

\paragraph{Molmo \cite{deitke2024molmo}}  
Molmo emphasizes transparency by being trained on the open PixMo dataset. It excels in fine-grained vision-language understanding and visual grounding tasks, though it is not optimized for structured tasks such as code execution or prompt engineering.

\paragraph{Additional Open-Source Models}  
Other notable models include InternVL2-1B \cite{chen2024internvl}, Qwen2-VL-2B-Instruct \cite{yang2024qwen2}, MiniMonkey \cite{huang2024mini}, Paligemma-3B-mix-448 \cite{beyer2024paligemma}, Phi-3.5 VLM \cite{abdin2024phi}, LLaVA OneVision-7B \cite{li2024llava}, Ovis 1.5-Llama 3-8B \cite{lu2024ovis}, GLM-4v-9B \cite{glm2024chatglm}, Ovis1.6 \cite{lu2024ovis}, Llama3.2-Vision \cite{meta2024llama32}, Pixtral \cite{agrawal2024pixtral}, OmChat-V2 \cite{omchat2024v2}, and InternVL2-26B \cite{chen2024far, internvl2024}. Each of these models is characterized by distinct combinations of language and vision encoders (e.g., Qwen, Gemma, Llama, GLM; ViT, SigLIP, CLIP, EVA) and varying parameter counts, offering a wide spectrum of capabilities from lightweight, efficient inference to high-precision multimodal reasoning. This comprehensive review highlights the diverse strategies in multimodal model design and evaluation, which underpin the criteria for our model selection. 

\section{Detailed Review of Prompt Engineering Methods}
\label{apdx_prompt_methods}

This appendix provides an in-depth review of the seven prompting techniques employed in this study. For each method, we discuss the underlying principles, key features, and provide example templates along with usage scenarios in both unimodal and multimodal contexts.

\subsection{Zero-Shot Prompting}
Zero-shot prompting involves providing the model with only the task description, relying entirely on its pre-trained knowledge \cite{radford2019language}. This method simplifies prompt design and reduces computational overhead, making it ideal for general-purpose tasks.
\begin{figure}[H]
\centering
\includegraphics[width=0.5\textwidth, angle=0]{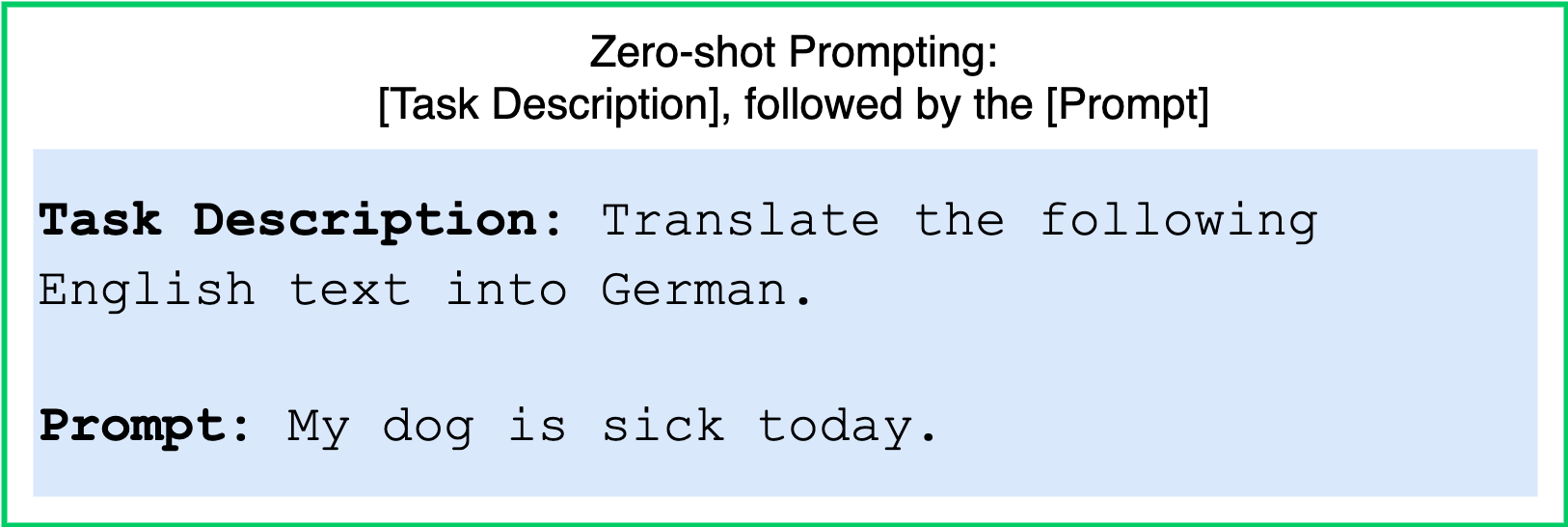}
\caption{Zero-shot Prompting Syntax}
\label{fig_zsp_app}
\end{figure}

\subsection{One-Shot Prompting}
One-shot prompting includes a single example alongside the task description to direct the model toward the desired output \cite{mann2020language}. This method provides minimal contextual guidance, balancing efficiency and accuracy for moderately complex tasks.
\begin{figure}[H]
\centering
\includegraphics[width=0.5\textwidth, angle=0]{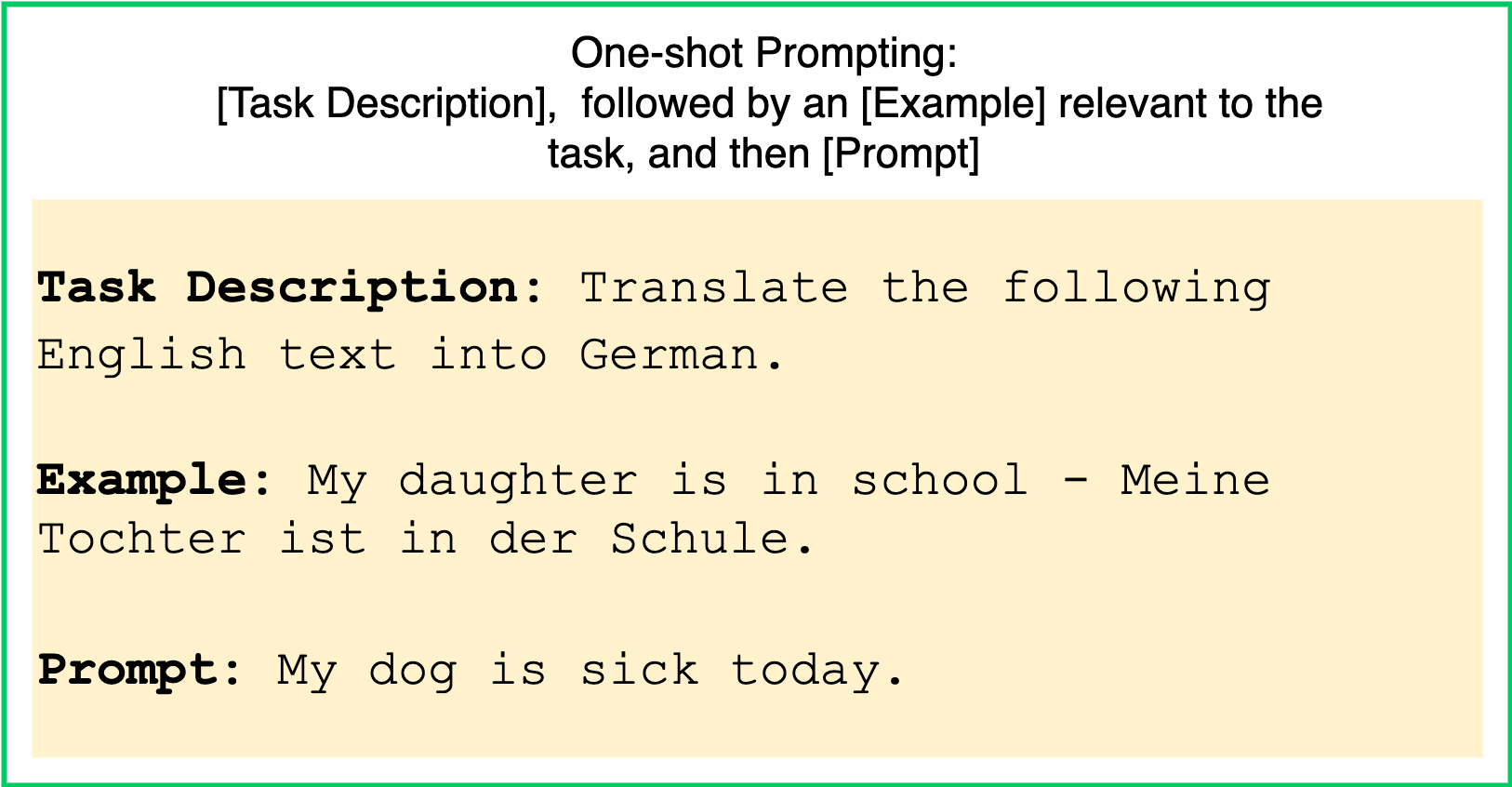}
\caption{One-shot Prompting Syntax}
\label{fig_osp_app}
\end{figure}

\subsection{Few-Shot Prompting}
Few-shot prompting incorporates multiple examples to establish clear input-output patterns, which is particularly useful for tasks that require structured responses \cite{mann2020language}.
\begin{figure}[H]
\centering
\includegraphics[width=0.5\textwidth, angle=0]{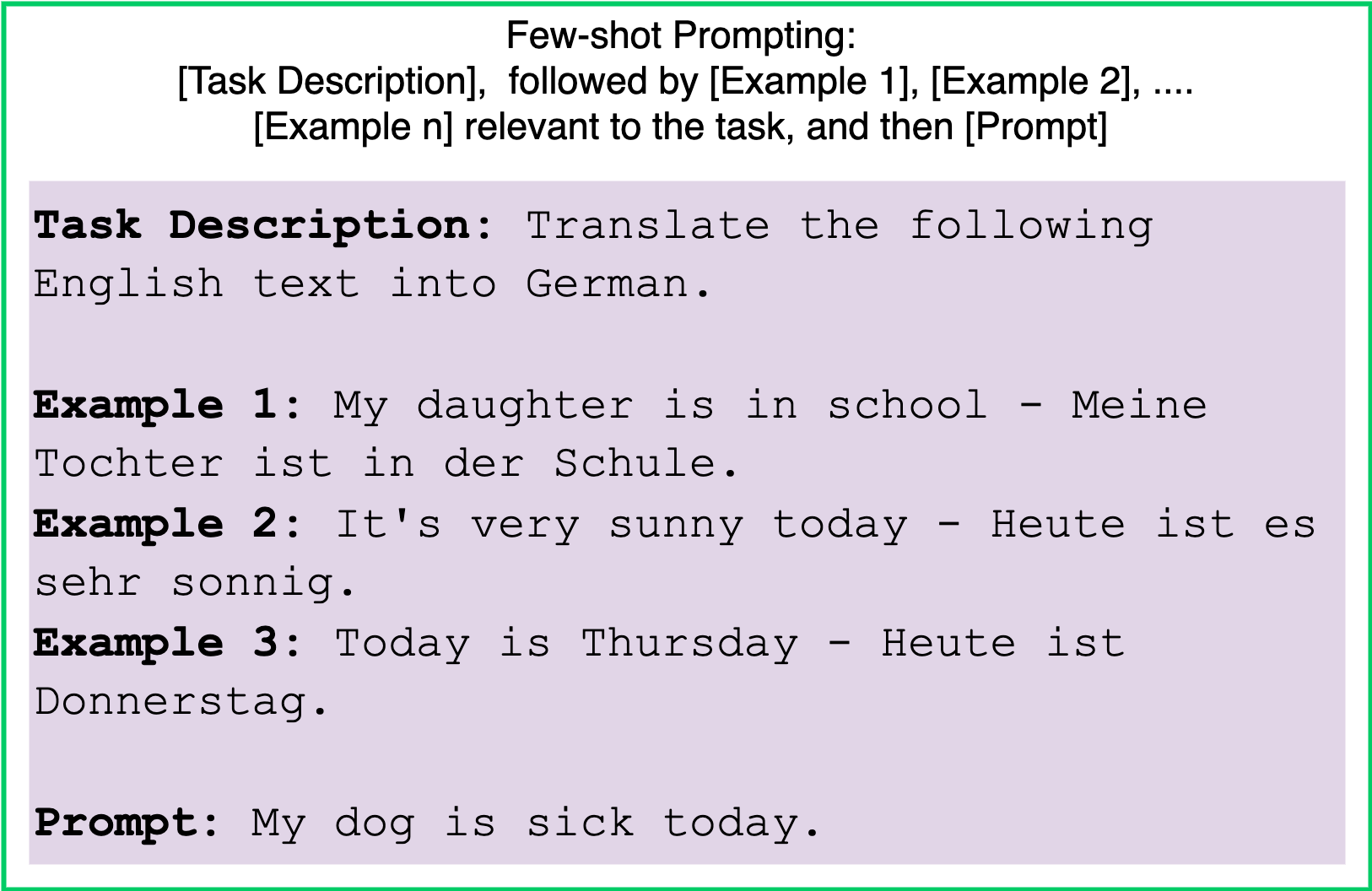}
\caption{Few-shot Prompting Syntax}
\label{fig_fsp_app}
\end{figure}

\subsection{Chain-of-Thought (CoT) Prompting}
Chain of Thought prompting encourages models to decompose problems into intermediate reasoning steps, thereby improving logical progression and accuracy in complex tasks \cite{wei2022chain, kojima2022large, zhang2022automatic}.
\begin{figure}[H]
\centering
\includegraphics[width=0.5\textwidth, angle=0]{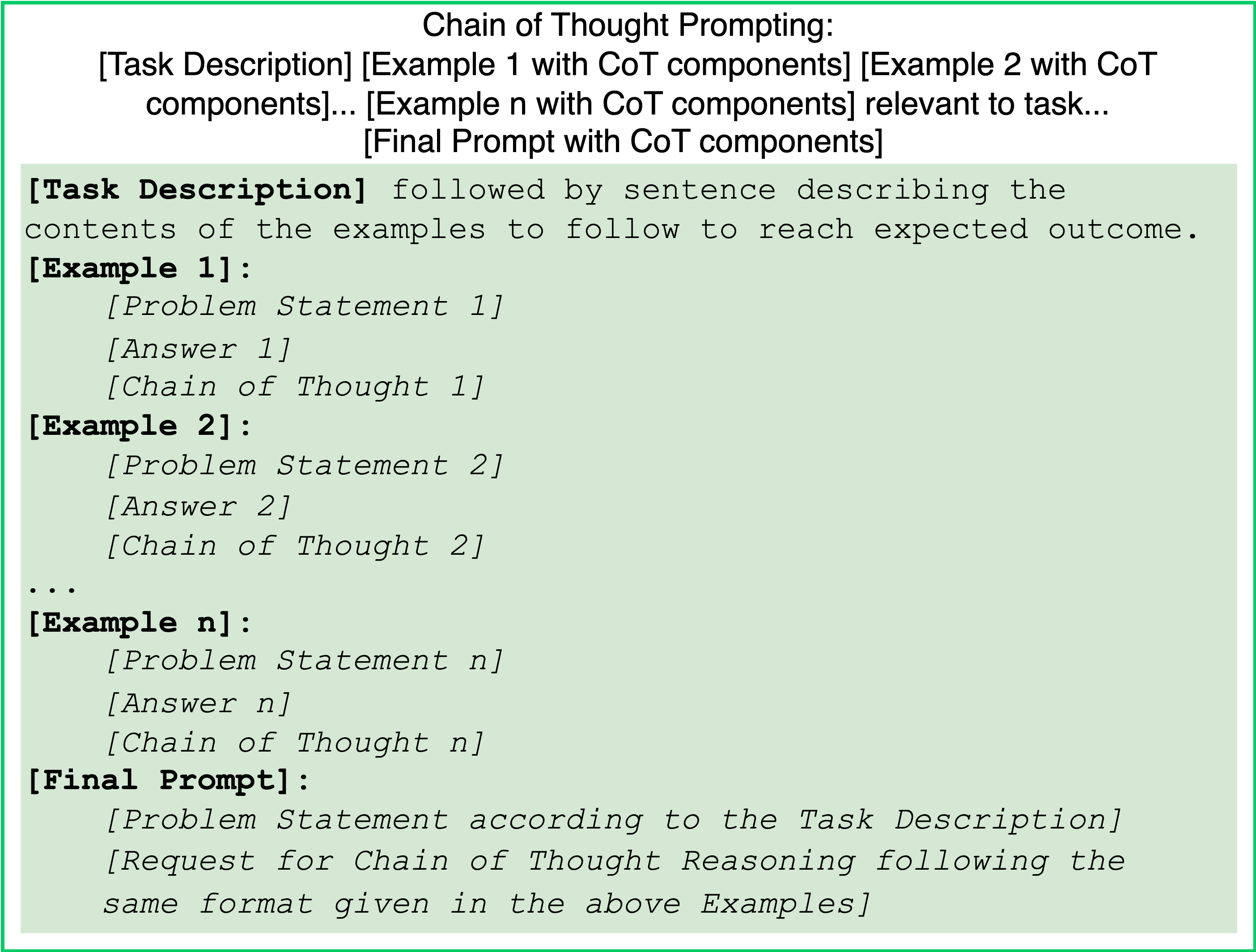}
\caption{Chain-of-Thought (CoT) Prompting Syntax}
\label{fig_cot_app}
\end{figure}

\subsection{Analogical Prompting}
Analogical prompting utilizes analogous examples that closely align with the task's requirements to foster indirect reasoning and creativity. This method enables models to transfer knowledge based on structural similarities between scenarios \cite{yasunaga2023large, lu2021fantastically, wu2022self, guo2024can}.
\begin{figure}[H]
\centering
\includegraphics[width=0.5\textwidth, angle=0]{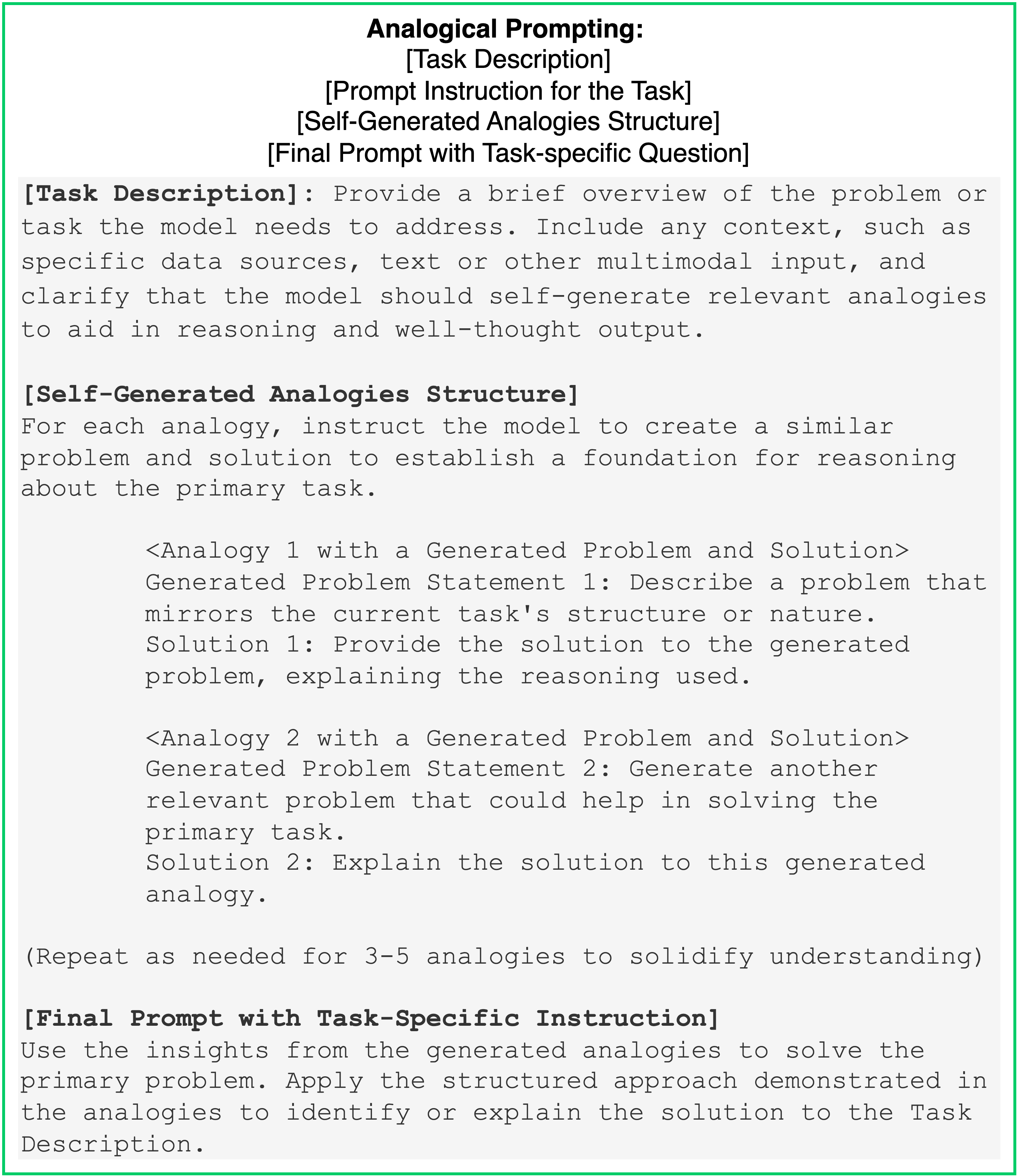}
\caption{Analogical Prompting Syntax}
\label{fig_ana_app}
\end{figure}

\subsection{Generated Knowledge Prompting}
Generated Knowledge Prompting involves prompting the model to generate additional task-relevant background knowledge, which is then used to improve reasoning and decision-making \cite{liu2021generated, liu2023pre}. This technique enriches the input context, leading to improved output accuracy.
\begin{figure}[H]
\centering
\includegraphics[width=0.5\textwidth, angle=0]{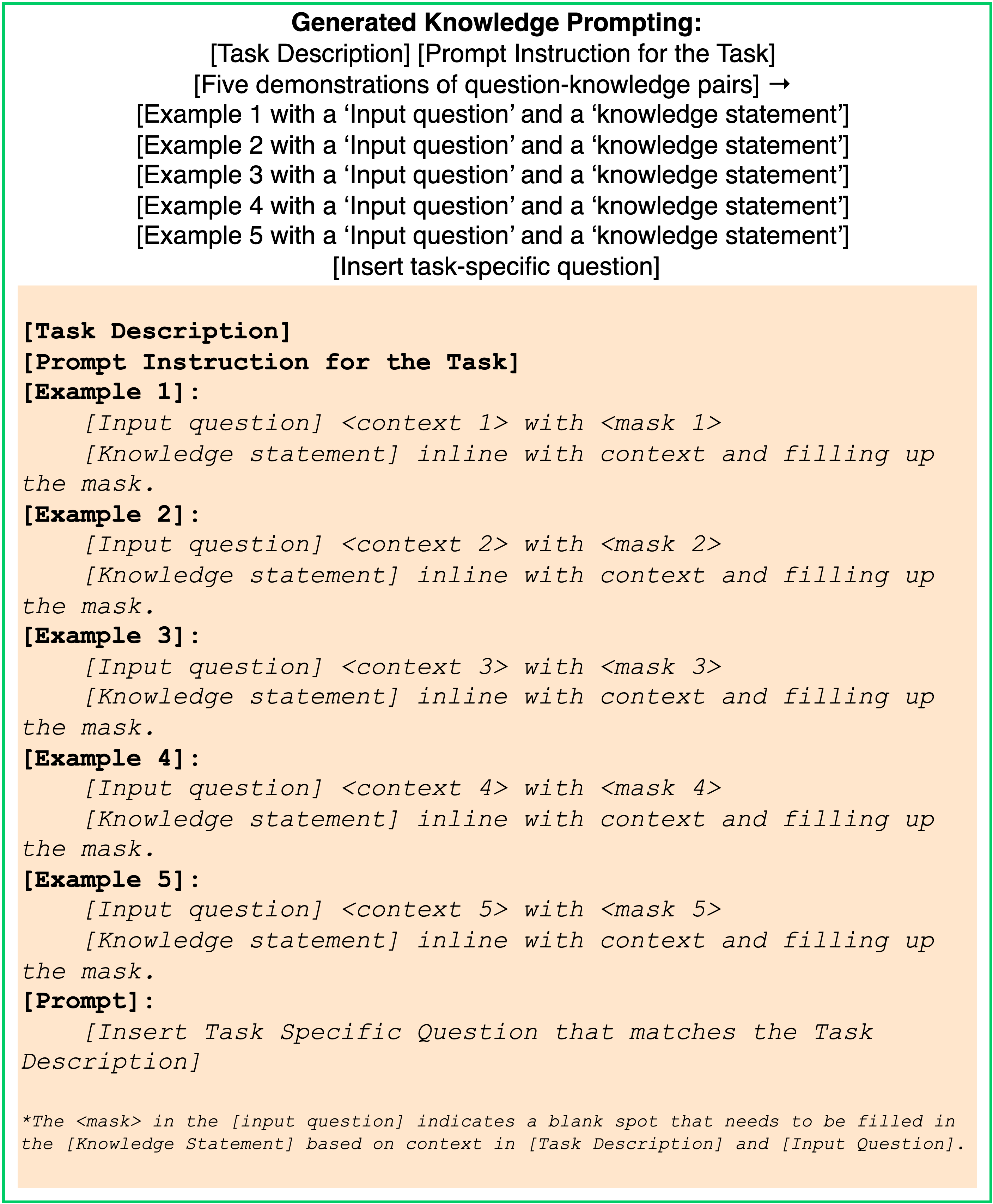}
\caption{Generated Knowledge Prompting Syntax}
\label{fig_genk_app}
\end{figure}

\subsection{Tree-of-Thought (ToT) Prompting}
Tree of Thought Prompting extends the Chain of Thought framework by organizing reasoning into a decision tree. This structure allows the model to explore multiple reasoning paths before converging on a final solution \cite{yao2024tree}, making it particularly effective for exploratory and decision-making tasks.
\begin{figure}[H]
\centering
\includegraphics[width=0.5\textwidth, angle=0]{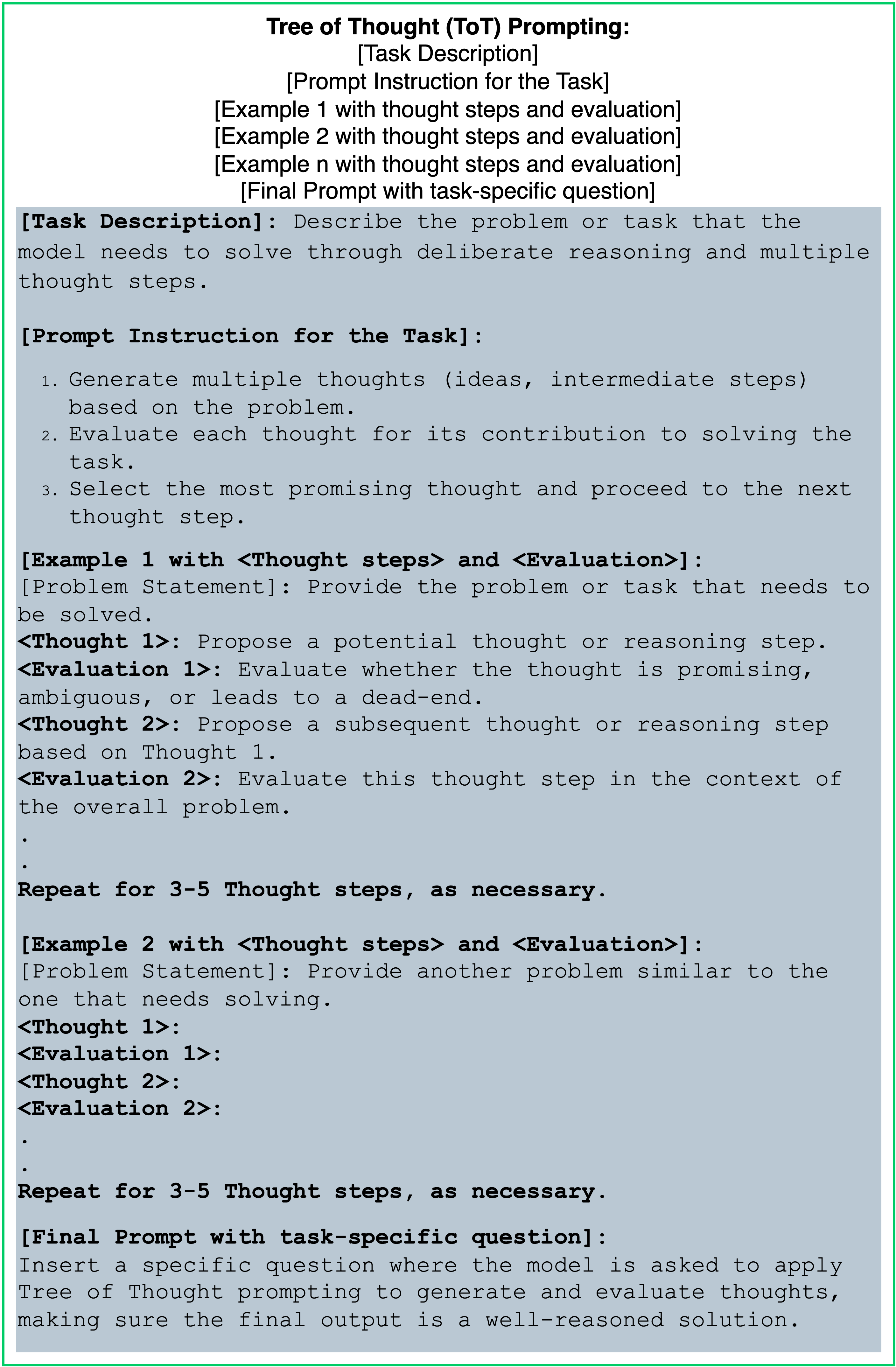}
\caption{Tree-of-Thought Prompting Syntax}
\label{fig_tot_app}
\end{figure}

\section{Task Design Methodology for Evaluation Aspects}
This appendix section provides a detailed account of the task design methodology for each of the four Evaluation Aspects (EA1-EA4). Each subsection outlines the specific objectives, rationale, and design principles used to create the tasks, with a focus on probing different multimodal capabilities of MLLMs. EA1 focuses on reasoning and compositionality, EA2 on multimodal understanding and alignment, EA3 on code generation from visual inputs, and EA4 on integrating contextual knowledge with multimodal cues. Together, these subsections offer a comprehensive view of how the tasks were systematically designed to simulate real-world multimodal scenarios and evaluate model performance across diverse cognitive and functional dimensions.

\subsection{EA1 Task Design Methodology}
\label{apdx_ea1}

The tasks in the study were systematically crafted to evaluate the reasoning and compositionality abilities of MLLMs. Each task was designed with a specific objective and level of complexity, and the accompanying visuals were tailored to challenge the models' multimodal reasoning capabilities. Below, we outline the thought process behind designing the tasks in Evaluation Aspect 1 (Reasoning and Compositionality) and their corresponding visual aids. 

The images for each task were crafted to align with the task objectives while enhancing multimodal reasoning. They provide sufficient context for reasoning without overloading unnecessary details. Every visual element (such as shapes, colors, or character actions) directly contributes to the reasoning challenge. Additionally, the visual style across tasks maintains a professional yet engaging appearance, aiding comprehension and focus.
The design of these tasks reflects a deliberate effort to test different dimensions of reasoning and compositionality in MLLMs. By combining well-structured multimodal inputs with incrementally complex challenges, the study ensures that the evaluation methodology is robust, comprehensive, and replicable.

The design of these tasks reflects a deliberate effort to test different dimensions of reasoning and compositionality in MLLMs. By combining well-structured multimodal inputs with incrementally complex challenges, the study ensures that the evaluation methodology is robust, comprehensive, and replicable.

For Task 1 (Pattern Recognition in Visual Sequences), our objective was to assess the model's ability to identify and extrapolate logical patterns across multiple modalities including numbers, shapes, and colors. The task integrates visual and textual elements, with images depicting sequences of numbers, geometric shapes, and colors. The pattern increases in complexity by combining arithmetic progression, geometric reasoning, and ambiguous color sequences. The sequences provide sufficient data points for models to deduce the logical rule while introducing ambiguity in color patterns to test reasoning flexibility. A carefully designed image illustrates the sequence, ensuring clarity and challenge.
In Task 2 (Logical Deduction from Text and Simplified Diagram), we aimed to evaluate deductive reasoning by combining textual descriptions and a simplified diagram of a real-world scenario. The task requires the model to synthesize text and visual clues (such as a broken window and a football near the window) to deduce the most likely event. A relatable scenario involving siblings and hobbies was chosen to simulate real-life reasoning challenges. Conflicting clues were introduced to evaluate the model's ability to prioritize relevant evidence. A clear, engaging image illustrates the suspects and the scene, making the task visually intuitive.
Task 3 (Mathematical Puzzle with Visual Data) was designed to test numerical reasoning and trend identification by combining tabular and graphical representations of data. Sales figures across four quarters provide a structured data source, while a bar chart complements the table, allowing for cross-referencing between modalities. The task requires models to calculate totals, identify trends, and predict future performance based on observed data. The bar chart and table are designed for easy interpretation while presenting enough complexity to challenge reasoning skills.
Finally, Task 4 (Story Synthesis from Text and Image) evaluates narrative synthesis by requiring models to generate coherent continuations from textual and visual inputs. The task integrates a textual scenario about a science fair and an image of a student with a renewable energy project to provide rich input. The visual representation of characters and their environment helps models establish a narrative context. The task encourages the model to infer plausible events and outcomes based on the given details. The image highlights key elements of the narrative, ensuring alignment with the text.

\subsection{EA2 Task Design Methodology}
\label{apdx_ea2}
The tasks in EA2 (Multimodal Understanding and Alignment) were developed to assess how well MLLMs integrate and align information from diverse modalities, such as images, text, and charts. Each task was designed to evaluate a distinct dimension of multimodal reasoning, such as matching, inference, translation, and consistency detection.
The design approach for EA2 tasks involves real-world multimodal challenges inspired by scenarios where visual and textual data coexist, such as articles with accompanying images and graphs with interpretations. Tasks progress from simpler matching exercises to more abstract reasoning and detailed cross-modal verification, incorporating incremental complexity. Each task requires the model to go beyond basic alignment by explaining its reasoning, testing both understanding and coherence.
The images for EA2 tasks were carefully created to enhance multimodal alignment, providing clear and meaningful data and ensuring that every detail contributes to the task's objective. While maintaining visual richness, the images avoid unnecessary complexity, allowing the focus to remain on reasoning and interpretation. The visuals simulate realistic situations, such as kayaking in nature, abstract art, and unemployment graphs, to make tasks relatable and engaging. Each visual is paired with textual or graphical content in a way that highlights dependencies between modalities.

The objective of Task 1 (Image-Text Matching and Explanation) is to evaluate the model's ability to match visual scenes with corresponding textual descriptions and provide reasoning for each match. The task integrates visual and textual modalities, requiring the model to align content based on key characteristics such as kayaking, manufacturing, and cooking. The model must discern details in each image and accurately match them to abstract textual concepts, testing interpretive and alignment skills. By providing explanations for each match, the task ensures the model demonstrates understanding rather than guessing. Clear and detailed images are used to depict diverse scenes (adventure, industry, and culinary arts), making the task visually intuitive while challenging. Explanation prompts ensure the reasoning process is transparent and logical. Task 2 (Inferring Context from Combined Modalities) aims to assess the model's ability to infer additional context by synthesizing textual and visual information. The task combines a descriptive paragraph (e.g., Maria walking on an empty street at dusk) with an image (a clock tower in a quiet city square) to test the model's ability to infer time, setting, and motivations. The model must analyze textual hints (e.g., streetlights and vendors closing) and visual cues (e.g., clock showing 7:30 PM) to arrive at accurate conclusions. The task introduces subtle contextual clues, requiring the model to resolve ambiguity using logical inference. The image is designed to evoke a specific time and atmosphere, enhancing the model's ability to integrate visual and textual data. 
The objective of Task 3 (Cross-Modal Translation) is to evaluate the model's ability to interpret abstract visual art and translate it into coherent literary themes or narratives. The abstract painting (depicting turbulent seas under sunlight) challenges the model to interpret artistic elements (colors, patterns, and mood) and map them to poetic themes. The task emphasizes translating visual impressions into descriptive language, requiring creative reasoning. By interpreting the painting's mood (e.g., chaotic yet hopeful), the model's ability to infer emotional tones from visuals is tested. The abstract painting is detailed and expressive, providing ample cues for interpretation while leaving room for subjective reasoning.
Task 4 (Aligning Data from Charts and Text) tests the model's capability to detect inconsistencies between data presented in a visual chart and textual descriptions. The task involves analyzing a line graph (e.g., unemployment rates) and cross-referencing it with a textual description to identify mismatches. It evaluates the model's ability to spot errors or inconsistencies (e.g., the text claims a decrease to 2\% while the chart shows an increase to 7\% in 2020). The task requires detailed attention to both visual and textual details, ensuring the model's output is precise and evidence-backed. A simple, clean line chart is used to focus on key data points while minimizing distractions.

\subsection{EA3 Task Design Methodology:}
\label{apdx_ea3}
The tasks in EA3 are designed to evaluate the ability of MLLMs to generate accurate and functional code from multimodal inputs. Each task assesses the model's capacity to interpret visual instructions, generate context-specific code, and execute logical steps. These include interpreting tables, flowcharts, and images that describe problems or provide structured data inputs. The model is expected to produce code that performs well-defined operations such as data visualization, arithmetic computation, or sequence generation, transforming abstract prompts into executable logic. The tasks are carefully crafted to reflect a progression in complexity, ranging from basic operations, such as extracting numerical data from an image to more intricate challenges like generating code from flowchart-based instructions. This incremental design facilitates the evaluation of a model’s ability to reason and scale its performance in increasingly demanding coding scenarios.

To support this, the visual components of EA3 tasks were created with clarity, relevance, and realism in mind. Images were designed to be well-organized and unambiguous, directly tied to the task objectives without introducing extraneous elements. They simulate realistic programming scenarios, including the interpretation of structured data or the automation of repetitive tasks. As complexity increases across tasks, so too does the visual intricacy, providing a robust benchmark for assessing the model’s capacity for abstraction and stepwise code generation.

Task 1's objective is to test the model's ability to interpret tabular data in an image and generate Python code to visualize it as a bar chart.  The task involves extracting structured data (e.g., sales over four quarters) from an image. The model must generate correct visualization code, including libraries (e.g., Matplotlib). By providing a structured table, the task ensures a clear yet challenging input for code generation. The task simulates a common data science workflow, aligning with practical applications. The key goal of Task 2 is to assess the model's ability to generate Python code (e.g., Turtle graphics) to draw a shape depicted in an image.  The task requires interpreting an image of a geometric shape (e.g., a star or hexagon). The model must write accurate code that replicates the shape using a specific library. The task evaluates the model’s attention to dimensions and proportions in the visual input. The task mirrors scenarios in educational coding exercises, emphasizing beginner-friendly challenges. Similarly, Task 3 focuses to test the model’s ability to extract numerical information from textual input in an image and compute a sum. The task requires extracting numbers from an image (e.g., a shopping receipt). The model must compute a sum based on the parsed input, testing logical reasoning. The task combines OCR (optical character recognition) capabilities with mathematical operations. The scenario mimics practical use cases, such as expense calculation from invoices. Task 4 is designed to evaluate the model’s ability to convert chart data into a structured Python dictionary.  The task requires extracting information from a bar chart or pie chart. The model must organize extracted data into a Python dictionary with key-value pairs. The task tests the model’s ability to ensure the structured output matches the visual input. This mirrors tasks in data engineering or ETL (Extract, Transform, Load) workflows. Task 5 tests the models' ability to parse text in an image (e.g., itemized receipt) and perform basic arithmetic. The task involves identifying item prices in an image and summing them up.  Models must handle inconsistencies or unclear inputs in the image (e.g., smudged text). The task simulates expense tracking and financial analysis workflows. Task 6 assess the models' ability to interpret a CSV-like structure in an image and convert it into a Python-compatible format. The task requires parsing tabular data with headers and rows.  The model must generate Python code to represent the CSV data as a list of dictionaries or Pandas DataFrame. Combines OCR capabilities with data engineering skills. This task reflects real-world scenarios in data preprocessing. Task7 evaluate the models' ability to interpret step-by-step algorithm instructions in an image and generate functional code. The task requires the model to translate a flowchart or textual steps into Python code. The model must follow a structured logical process (e.g., loops, recursion) to compute the sequence. This task emphasizes algorithm design and educational coding scenarios. Task 8 is designed to test the model’s ability to follow decision-making logic in a flowchart and implement it as a Python function.  The task requires interpreting decision nodes and paths in the flowchart. The model must translate the flowchart logic into executable code. The task evaluates whether the model can adhere to predefined logical structures.  Flowchart-based programming tasks are widely used in both educational and industrial contexts.

\subsection{EA4 Task Design Methodology}
\label{apdx_ea4}
The EA4 task set simulates a wide range of real-world challenges that demand the integration of multimodal inputs with contextual knowledge. Designed to rigorously test MLLMs' ability to reason, retrieve, and synthesize across domains, these tasks combine clear visual inputs with meaningful textual prompts to evaluate model performance in complex knowledge-driven scenarios. Each task in EA4 requires the model to retrieve relevant contextual knowledge, either from external sources or embedded within the input and make sense of the given scenario. This involves the ability to combine textual and visual content, such as maps, historical images, scientific charts, or cultural references, and generate informed outputs. The tasks are crafted to reflect authentic applications in domains such as history, science, and fact-checking, often requiring the model to resolve ambiguity by reasoning through incomplete or conflicting information.

To support this, the visuals accompanying EA4 tasks are carefully designed for contextual relevance. Each image aligns closely with the task narrative, whether depicting a historical landmark, cultural artifact, or scientific figure and provides just enough detail to encourage reasoning without introducing unnecessary visual complexity. These visuals are grounded in real-world scenarios, helping simulate tasks such as identifying locations, analyzing diagrams, or verifying facts. Moreover, they are deliberately constructed to complement the associated textual inputs, promoting effective multimodal interaction and integrated reasoning across modalities.

Task 1 evaluates the models' ability to identify a historical monument and explain its significance using external knowledge. The task requires the model to identify a landmark (e.g., a clock tower) from its image; where the model must retrieve relevant historical or cultural information about the landmark. The task tests the ability to synthesize visual data and external knowledge into a coherent explanation. Task 2 tests the model’s ability to analyze scientific data from a visual chart and textual explanation, integrating both to draw conclusions.  The task requires interpreting trends and relationships in a graph (e.g., unemployment rates). Hence the model must combine visual insights with textual explanations to infer implications. This task mimics scenarios in research or data journalism. Task 3 assess the models' ability to analyze medical images (e.g., X-rays) and integrate domain-specific knowledge to provide recommendations. The task involves identifying abnormalities or patterns in a medical image. Hence models must draw on medical knowledge to recommend next steps (e.g., tests, treatments). The task introduces subtle visual clues, requiring careful analysis, that reflects applications in AI-powered healthcare diagnostics. Tasks 4 is designed to evaluate the models' ability to interpret cultural artifacts using visual and textual context. The task involves identifying a cultural artifact (e.g., sculpture or painting) from its image. With this, the model must explain the artifact’s significance, historical background, and cultural relevance, by encouraging nuanced reasoning and interpretation. Task 5 aims to test the models' ability to analyze a map and textual description to discuss historical events. The task requires interpreting map data (e.g., trade routes, battlefields). The model must combine geographical insights with historical narratives. It tests the ability to hypothesize based on multimodal inputs, that reflects challenges in historical research or geographic analysis. Task 6 assess the ability to cross-reference data from a chart (e.g., energy trends) with textual information in an article, that combines quantitative data with qualitative reasoning. The task involves verifying claims in the article against chart data, testing the model’s ability to detect inconsistencies or validate arguments; reflects tasks in journalism or policy analysis.
Task 7 evaluate the models' ability to verify claims from headlines using multimodal inputs (e.g., images, encyclopedia excerpts). This requires cross-referencing headlines with visual and textual evidence. The model must retrieve and synthesize external information, while examining the critical thinking and consistency across modalities. Task 8 assess the model’s ability to analyze artwork in relation to its historical and cultural background. The task involves identifying key elements of the artwork and explaining their historical significance, by combining visual understanding with contextual knowledge. This design ensures that EA4 assesses knowledge-intensive multimodal capabilities in modern large models to move beyond surface-level understanding toward deeper, multi-faceted reasoning.

Detailed task descriptions and corresponding expected outputs for each evaluation aspect are provided in the supplementary material accompanying this paper.

\section{Detailed Evaluation Criteria}
\label{apdx_evalcriteria}

For transparency and reproducibility, this section provides the detailed criteria used to evaluate model outputs across the four key metrics: Accuracy, Relevancy, Conciseness, and Hallucination.

\begin{table}[H]
\caption{Detailed evaluation criteria for assessing model outputs. These criteria supplement the empirical thresholds provided in Table~\ref{tab:evaluation_thresholds} and ensure a nuanced and consistent assessment of model performance.}
\label{tab:detailed_evalcriteria}
\centering
\renewcommand{\arraystretch}{1.1}
\setlength{\tabcolsep}{8pt}
\begin{tabularx}{\textwidth}{l l X}
\toprule
\textbf{Criterion} & \textbf{Category} & \textbf{Description} \\
\midrule
\textbf{Accuracy} & Correct & The response addresses all components of the task correctly. \\
                  & Partially Correct & The response addresses only some parts of the task. \\
                  & Incorrect & The response fails to correctly address the task; any error in multi-part tasks renders it incorrect. \\
\midrule
\textbf{Relevancy} & Relevant & The response is fully aligned with the task and context. \\
                  & Partially Relevant & The response contains some relevant information but is incomplete. \\
                  & Irrelevant & The response is unrelated to the task or context. \\
\midrule
\textbf{Conciseness} & Under-Explained & The response lacks sufficient detail to explain its reasoning. \\
                   & To the Point & The response is clear, concise, and appropriately detailed. \\
                   & Over-Explained & The response includes redundant or unnecessary elaboration. \\
\midrule
\textbf{Hallucination} & Yes & The response includes irrelevant, repetitive, or random content not pertinent to the task. \\
                     & No & The response remains focused and free of irrelevant content. \\
\bottomrule
\end{tabularx}
\end{table}

\section{Experimental Setup}
\label{experimental_setup}

The experiment was conducted on two separate servers, each with distinct hardware specifications. Connect1 server was used to run the experiment using Python-based implementations, leveraging its optimized computational capabilities for executing MLLM workloads. The Connect3 server was utilized to run the experiment with Ollama for getting the results of Llama3.2 Vision model. The model is queried from Connect1 server through API call. 

\begin{table}[H]
\centering
\caption{Connect1 Server Specification}
\begin{tabular}{|ll|}
\hline
\multicolumn{2}{|c|}{\textbf{CPU Information}}                                                    \\ \hline
\multicolumn{1}{|l|}{\textbf{Technical Specification}} & \textbf{Intel Cascadelake SP processor}  \\ \hline
\multicolumn{1}{|l|}{Processor}                        & Intel(R) Xeon(R) Gold 6252 CPU @ 2.10GHz \\ \hline
\multicolumn{1}{|l|}{OS}                               & Ubuntu 23.04                             \\ \hline
\multicolumn{1}{|l|}{Micro-architecture}               & Cascadelake                              \\ \hline
\multicolumn{1}{|l|}{Thread(s) per core}               & 2                                        \\ \hline
\multicolumn{1}{|l|}{Cores per socket}                 & 24                                       \\ \hline
\multicolumn{1}{|l|}{Socket(s)}                        & 2                                        \\ \hline
\multicolumn{1}{|l|}{NUMA node(s)}                     & 2                                        \\ \hline
\multicolumn{1}{|l|}{L1d cache}                        & 1.5 MiB                                  \\ \hline
\multicolumn{1}{|l|}{L1I cache}                        & 1.5 MiB                                  \\ \hline
\multicolumn{1}{|l|}{L2 cache}                         & 48 MiB                                   \\ \hline
\multicolumn{1}{|l|}{L3 cache}                         & 71.5 MiB                                 \\ \hline
\multicolumn{1}{|l|}{Main memory}                      & 256 GB                                   \\ \hline
\multicolumn{2}{|c|}{\textbf{GPU Information}}                                                    \\ \hline
\multicolumn{1}{|l|}{GPU Model}                        & NVIDIA RTX A6000                         \\ \hline
\multicolumn{1}{|l|}{Memory}                           & 48 GB                                    \\ \hline
\multicolumn{1}{|l|}{Compute Capability}               & 8.6                                      \\ \hline
\end{tabular}
\end{table}

\begin{table}[H]
\centering
\caption{Connect3 Server Specification}
\begin{tabular}{|ll|}
\hline
\multicolumn{2}{|c|}{\textbf{CPU Information}}                                                     \\ \hline
\multicolumn{1}{|l|}{\textbf{Technical Specification}} & \textbf{Intel Cascadelake SP processor}   \\ \hline
\multicolumn{1}{|l|}{Processor}                        & Intel(R) Xeon(R) Gold 5220R CPU @ 2.20GHz \\ \hline
\multicolumn{1}{|l|}{OS}                               & Ubuntu 23.04                              \\ \hline
\multicolumn{1}{|l|}{Micro-architecture}               & Cascadelake                               \\ \hline
\multicolumn{1}{|l|}{Thread(s) per core}               & 2                                         \\ \hline
\multicolumn{1}{|l|}{Cores per socket}                 & 24                                        \\ \hline
\multicolumn{1}{|l|}{Socket(s)}                        & 2                                         \\ \hline
\multicolumn{1}{|l|}{NUMA node(s)}                     & 2                                         \\ \hline
\multicolumn{1}{|l|}{L1d cache}                        & 1.5 MiB                                   \\ \hline
\multicolumn{1}{|l|}{L1I cache}                        & 1.5 MiB                                   \\ \hline
\multicolumn{1}{|l|}{L2 cache}                         & 48 MiB                                    \\ \hline
\multicolumn{1}{|l|}{L3 cache}                         & 71.5 MiB                                  \\ \hline
\multicolumn{1}{|l|}{Main memory}                      & 256 GB                                    \\ \hline
\multicolumn{2}{|c|}{\textbf{GPU Information}}                                                     \\ \hline
\multicolumn{1}{|l|}{GPU Model}                        & Persistence-M                             \\ \hline
\multicolumn{1}{|l|}{Memory}                           & 24 GB                                     \\ \hline
\multicolumn{1}{|l|}{Compute Capability}               & 8                                         \\ \hline
\end{tabular}
\end{table}

\clearpage

\renewcommand{\thesection}{S\arabic{section}} 
\setcounter{section}{0}

\section*{Supplementary Material}

\addcontentsline{toc}{section}{Supplementary Material}

\section{Evaluation Aspect 1 (EA1): Reasoning and Compositionality Tasks}
\label{supp_EA1}
\subsection{Task 1: Pattern Recognition in Visual Sequences}

\textbf{Task Description}
You are presented with a sequence of images depicting a series of numbers and shapes following a specific pattern:
\begin{itemize}
    \item Image 1: A card showing the number "2" with a red triangle.
    \item Image 2: A card showing the number "4" with a blue square.
    \item Image 3: A card showing the number "6" with a green pentagon.
    \item Image 4: A card showing the number "8" with a yellow heptagon.
\end{itemize}

Question:
Based on the pattern observed in the sequence, identify the next card in the series, specifying both the number and the shape (including its color).

\textbf{Expected Output:} \\
Reasoning
\newline
Number Sequence Analysis:
\begin{itemize}
    \item The numbers increase by 2: 2, 4, 6, 8
    \item Next number: 8 $+$ 2 $=$ 10
\end{itemize}

Shape Sequence Analysis:
\begin{itemize}
    \item Number of sides: 3 (triangle), 4 (square), 5 (pentagon), 7 (heptagon)
    \item The sides increase by 1, but skip the hexagon (6 sides) and go from pentagon (5) to heptagon (7).
    \item The sequence of sides: 3, 4, 5, 7
    \item Pattern in sides: Increase by 1, but skip even numbers after 5.
\end{itemize}

Hypothesising the Next Shape:
\begin{itemize}
    \item Following the pattern, after 7 sides (heptagon), skip the next even number of sides (8 sides, octagon) and go to the next odd number of sides.
    \item Next odd number after 7 is 9.
    \item Shape with 9 sides: Nonagon
\end{itemize}
Color Sequence Analysis:
\begin{itemize}
    \item Colors: Red, Blue, Green, Yellow
    \item Possible pattern: Standard color wheel sequence or arbitrary.
    \item Assuming the sequence follows the rainbow order (Red, Orange, Yellow, Green, Blue, Indigo, Violet), but the colors do not match.
    \item Alternatively, assign the next color as Purple to introduce a new color.
\end{itemize}
Final Answer: 
The next card is a number "10" paired with a purple nonagon (9-sided polygon).

\begin{figure}[H]
    \centering
    \includegraphics[width=1.0\textwidth]{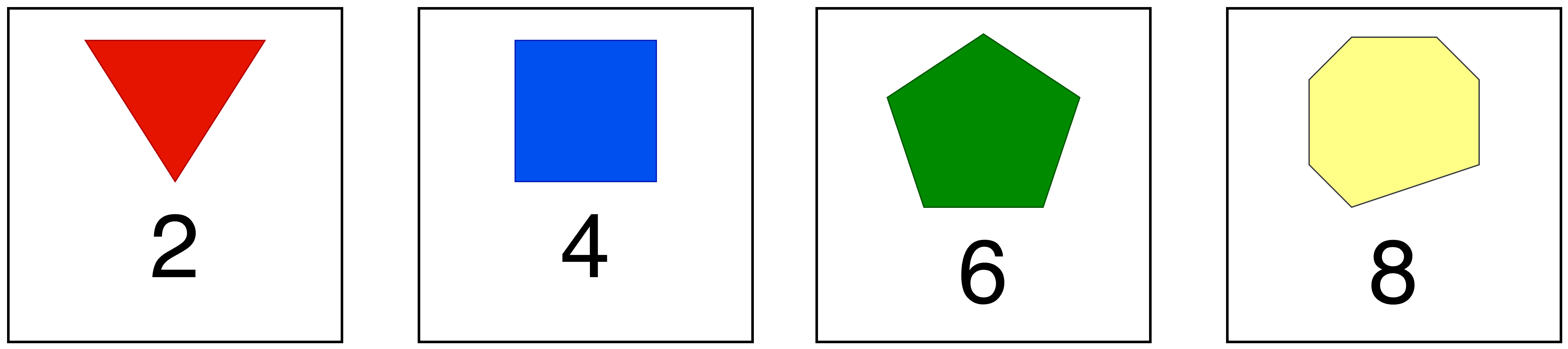}
    \caption{Input Figure for Task 1 for Evaluation Aspect 1 (Reasoning)}
    \label{fig:EA1_T1}
\end{figure}

\subsection{Task 2: Logical Deduction from Text and Simplified Diagram}
\textbf{Task Description:}
\begin{itemize}
    \item Diagram Provided: A simple diagram showing:
    \item Alice: Standing with paint stains on her clothes and a canvas nearby.
    \item Bob: Holding a football.
    \item Carol: Holding a violin case and looking at a broken window.
\end{itemize}
Additional Clues in the Diagram:
\begin{itemize}
    \item Bob’s football is on the ground next to a broken window.
    \item Alice’s hands have paint smudges.
    \item Carol seems surprised and is looking at the window.
\end{itemize}
Textual Information: 
“Alice, Bob, and Carol are siblings with different hobbies:
\begin{itemize}
    \item Alice loves painting.
    \item Bob enjoys sports.
    \item Carol is a musician who plays the violin.
\end{itemize}
Yesterday, one of them accidentally broke a window.”

Question: Based on the text and the diagram, deduce who is most likely to have broken the window and explain your reasoning.

Detailed Context and Hints:
Potential Suspects: Alice, Bob, Carol
Clues:
\begin{itemize}
    \item Bob is holding a football, with one lying near the broken window.
    \item Alice has paint stains, suggesting she was painting.
    \item Carol is looking at the window, suggesting she noticed it but didn’t cause it.
\end{itemize}

\textbf{Expected Output:} \\
Reasoning:
\begin{itemize}
    \item Alice: Painting and focused on her work, with no indication of her near the window.
    \item Bob: Football suggests he may have been playing nearby, and the window could have been accidentally broken by the ball.
    \item Carol: Gazing at the window likely means she discovered the broken window rather than caused it.
\end{itemize}
Final Answer: Bob is most likely to have broken the window while playing with the football, indicated by the ball lying near the broken window.

\begin{figure}[H]
    \centering
    \includegraphics[width=1.0\textwidth]{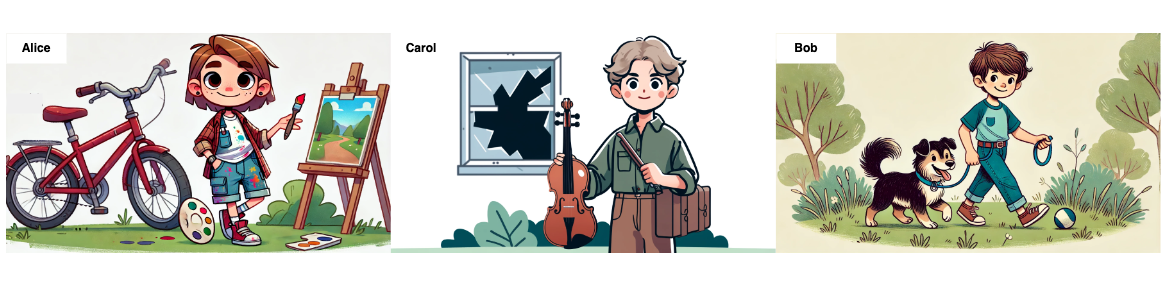}
    \caption{Input Figure for Task 2 for Evaluation Aspect 1 (Reasoning)}
    \label{fig:EA1_T2}
\end{figure}

\subsection{Task 3: Mathematical Puzzle with Visual Data}
\textbf{Task Description:} \\
Table Provided: A table showing the sales data of three products over four quarters.
\begin{table}[h]
    \centering
    \begin{tabular}{|c|c|c|c|}
        \hline
        \textbf{Quarter} & \textbf{Product X} & \textbf{Product Y} & \textbf{Product Z} \\
        \hline
        Q1 & \$10,000 & \$15,000 & \$20,000 \\
        Q2 & \$20,000 & \$25,000 & \$15,000 \\
        Q3 & \$30,000 & \$20,000 & \$25,000 \\
        Q4 & \$25,000 & \$30,000 & \$35,000 \\
        \hline
    \end{tabular}
    \caption{Quarterly Sales Data for Products X, Y, and Z}
    \label{tab:quarterly_sales}
\end{table}

Bar Chart Provided: A bar chart visually representing the same data.

Questions:
Based on the data provided:
\begin{enumerate}
    \item Which product had the highest total sales over the year?
    \item Identify any trends or patterns in sales for each product.
    \item Predict which product is likely to perform best in Q1 of the next year, and explain your reasoning.
\end{enumerate}

\textbf{Expected Output: } \\
Calculate Total Sales:
Product X:
\begin{itemize}
    \item Q1: \$10,000
    \item Q2: \$20,000
    \item Q3: \$30,000
    \item Q4: \$25,000
\end{itemize}
Total: \$85,000 \\
Product Y:
\begin{itemize}
    \item Q1: \$15,000
    \item Q2: \$25,000
    \item Q3: \$20,000
    \item Q4: \$30,000
\end{itemize}
Total: \$90,000 \\
Product Z:
\begin{itemize}
    \item Q1: \$20,000
    \item Q2: \$15,000
    \item Q3: \$25,000
    \item Q4: \$35,000
\end{itemize}
Total: \$95,000

Highest Total Sales: Product Z with \$95,000. \\
Identify Trends: \\
Product X:
\begin{itemize}
    \item Steady increase from Q1 to Q3.
    \item Slight decrease in Q4.
\end{itemize}
Product Y:
\begin{itemize}
    \item Increase from Q1 to Q2.
    \item Decrease in Q3.
    \item Increase again in Q4.
\end{itemize}
Product Z:
\begin{itemize}
    \item Decrease from Q1 to Q2.
    \item Increase in Q3 and significant increase in Q4.
\end{itemize}

Predicting Q1 Next Year Performance:
\begin{itemize}
    \item Product X: Had a slight decrease in Q4 after consistent growth; may stabilize or decrease.
    \item Product Y: Shows volatility but ended with a high in Q4; potential for good performance.
    \item Product Z: Significant growth in Q4; momentum likely to carry into Q1 next year.
\end{itemize}
Final Prediction:
Product Z is likely to perform best in Q1 of the next year due to its upward sales momentum in Q3 and Q4.

\begin{figure}[H]
    \centering
    \includegraphics[width=1.0\textwidth]{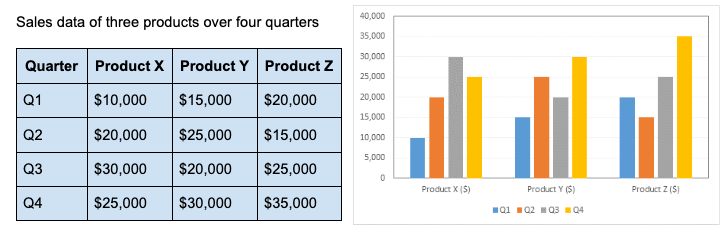}
    \caption{Input Figure for Task 3 for Evaluation Aspect 1 (Reasoning)}
    \label{fig:EA1_T3}
\end{figure}

\subsection{Task 4: Story Synthesis from Text and Image}
\textbf{Task Description:} \\
Text Fragment:
\textit{"During the annual science fair, students from various schools presented their innovative projects. Among them, a young boy named Liam stood nervously beside his exhibit."} \\
Image Provided:
A photo showing a boy next to a display titled "Renewable Energy Solutions," featuring a small wind turbine model and solar panels. Judges are seen approaching with clipboards.

\begin{figure}[H]
    \centering
    \includegraphics[width=0.5\textwidth]{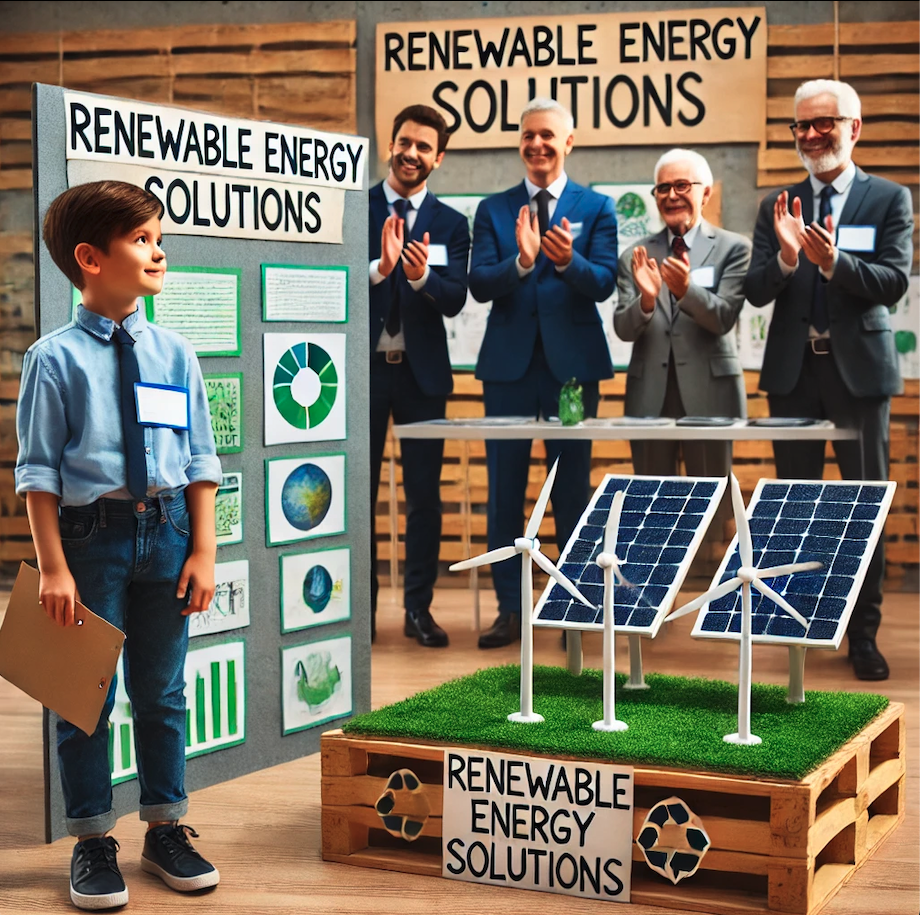}
    \caption{Input Figure for Task 4 for Evaluation Aspect 1 (Reasoning)}
    \label{fig:EA1_T4}
\end{figure}

Detailed Context:
Characters: Liam, judges, other students. \\
Elements to Include:
\begin{itemize}
    \item Liam's project on renewable energy.
    \item Interaction with judges.
    \item The outcome of the science fair.
\end{itemize}
Question:
Compose a narrative that continues the story, incorporating elements from the image, and conclude with the outcome of Liam's participation in the science fair.

\textbf{Expected Output:} \\
Narrative Continuation:
As the judges approached, Liam took a deep breath, recalling all the hours he spent perfecting his renewable energy project. His exhibit showcased a hybrid model combining wind and solar energy to power small communities. The judges listened attentively as he explained how his model could provide sustainable energy solutions in remote areas.
Impressed by his depth of knowledge and passion, they asked insightful questions, which Liam answered confidently. Throughout the day, many visitors stopped by, intrigued by his innovative approach.
At the award ceremony, anticipation filled the auditorium. When the announcer declared, "The first prize goes to Liam for his outstanding project on Renewable Energy Solutions," cheers erupted. Overwhelmed with joy, Liam stepped onto the stage to accept his trophy, knowing that his hard work had paid off and eager to continue his journey in sustainable technology.

\section{Evaluation Aspect 2 (EA2): Multimodal Understanding and Alignment Tasks}
\label{supp_EA2}
\subsection{Task 1: Image-Text Matching and Explanation}
\textbf{Task Description:} \\
You are provided with three images and three text descriptions. The goal is to match each image with the correct text description and explain the reasoning behind each match. \\
Images Provided:
\begin{itemize}
    \item Image A: A photograph of a person kayaking down a rapid river surrounded by forested mountains.
    \item Image B: An illustration of a robotic arm assembling a car in a factory setting.
    \item Image C: A picture of a chef tossing vegetables in a flaming pan in a professional kitchen.
\end{itemize}

\begin{figure}[H]
    \centering
    \includegraphics[width=0.6\textwidth]{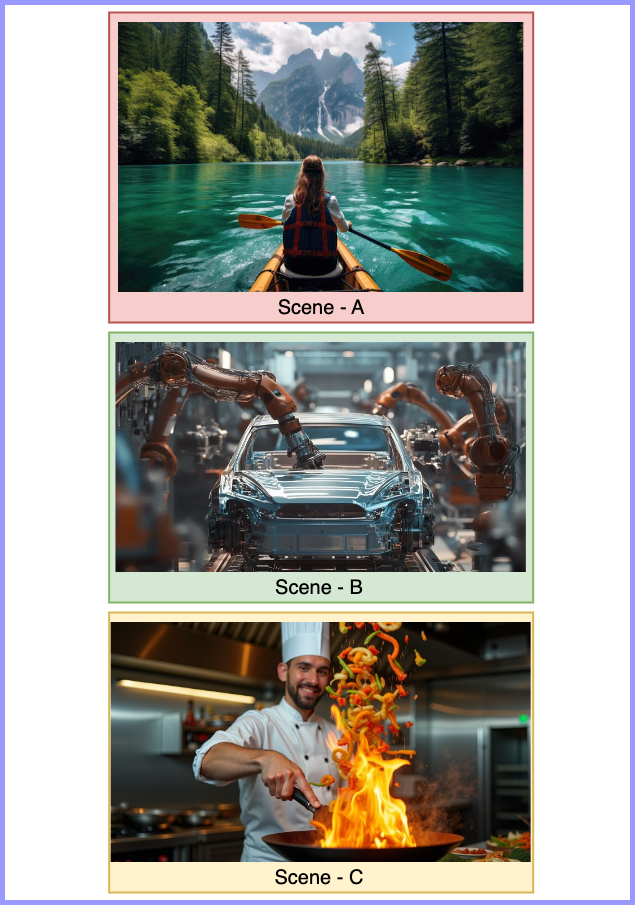}
    \caption{Input Figure for Task 1 for Evaluation Aspect 2}
    \label{fig:EA2_T1}
\end{figure}

Text Descriptions:
\begin{itemize}
    \item \textit{Description 1: "An industrial setting where automation plays a key role in manufacturing vehicles."}
    \item \textit{Description 2: "An adventurous individual navigating through natural landscapes, showcasing extreme sports."}
    \item \textit{Description 3: "Culinary arts in action, capturing the dynamic environment of a busy restaurant."}
\end{itemize}
Question:
Match each image with its corresponding text description and provide a detailed explanation for each pairing.

\textbf{Expected Output:}
Matching:
\begin{itemize}
    \item Image A matches with Description 2.
    \item Image B matches with Description 1.
    \item Image C matches with Description 3.
\end{itemize}
Explanation:
\begin{itemize}
    \item Image A and Description 2: Image A shows a person kayaking, which is an extreme sport involving navigating through rapids. Description 2 mentions "an adventurous individual navigating through natural landscapes, showcasing extreme sports," which aligns with the image content.
    \item Image B and Description 1: Image B depicts a robotic arm assembling a car, indicating automation in manufacturing. Description 1 refers to "an industrial setting where automation plays a key role in manufacturing vehicles," directly matching the image.
    \item Image C and Description 3: Image C shows a chef cooking in a professional kitchen with flames, highlighting the dynamic nature of culinary arts. Description 3 mentions "culinary arts in action, capturing the dynamic environment of a busy restaurant," which corresponds with the image.
\end{itemize}

\subsection{Task 2: Inferring Context from Combined Modalities}
\textbf{Task Description:}\\
You are presented with a short paragraph and an accompanying image. The task is to infer additional context about the situation by integrating information from both modalities. 

Text Provided:
\textit{"Maria checked her watch as she hurried down the quiet street. Most vendors had already closed their stalls, and the streetlights cast a soft glow on the empty cobblestones. She clutched a wrapped package tightly under her arm."} \\

Image Provided:
An image showing a street scene at dusk with shops closing, streetlights illuminating the area, and a clock tower showing the time as 7:30 PM.

\begin{figure}[H]
    \centering
    \includegraphics[width=0.6\textwidth]{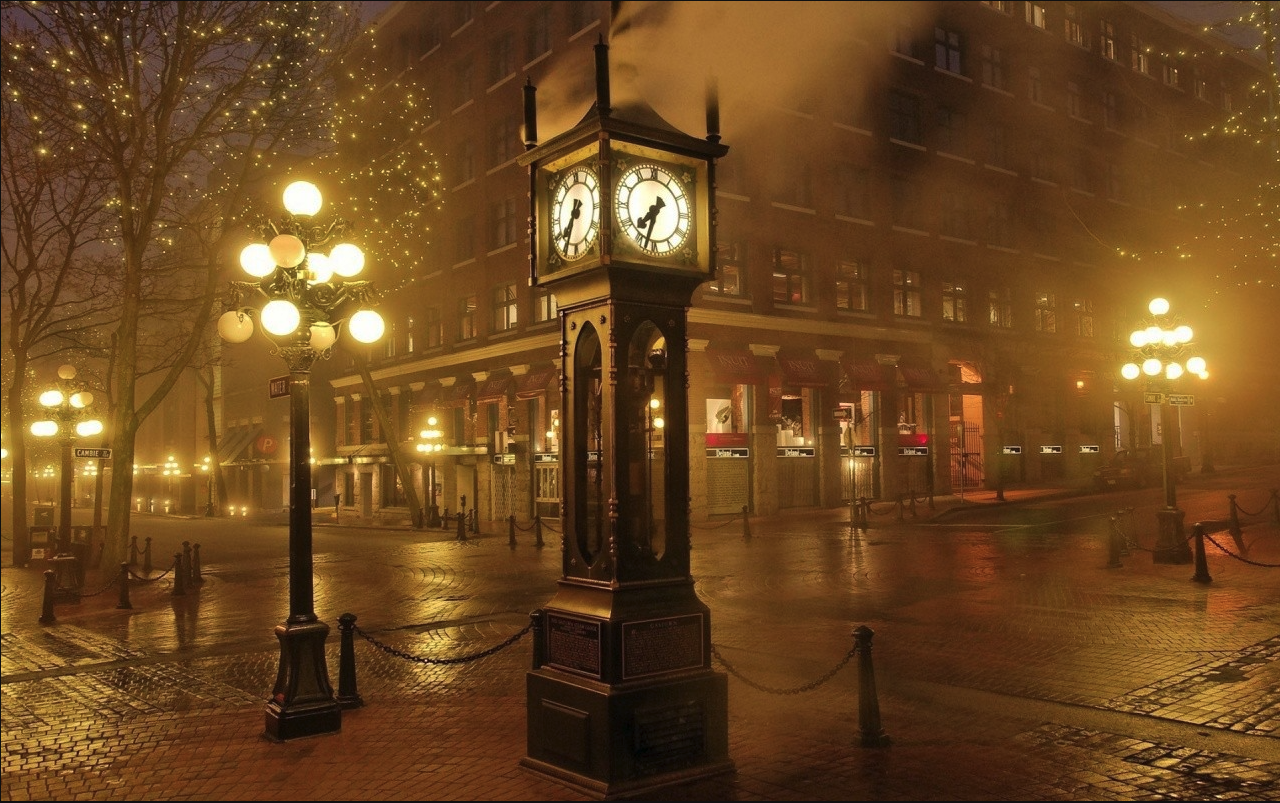}
    \caption{Input Figure for Task 2 for Evaluation Aspect 2}
    \label{fig:EA2_T2}
\end{figure}

Question:
Based on the text and the image, answer the following questions:
\begin{enumerate}
    \item What time of day is it, and how do you know?
    \item Why might Maria be in a hurry?
    \item What is the likely setting (e.g., city, town, village)?
\end{enumerate}

\textbf{Expected Output:}\\
\begin{enumerate}
    \item Time of Day: Answer: It is evening, around 7:30 PM. \textit{Explanation:} The text mentions "streetlights began to flicker on" and vendors closing stalls, suggesting it's getting dark. The clock in the image shows 7:30 PM, confirming the time. The empty streets indicate it’s late enough for most people to have already left the area.
    \item Why Maria Might Be in a Hurry: Answer: Maria might be trying to reach a destination before it closes or deliver the package by a certain time. \textit{Explanation:} She is checking her watch and hurrying, which implies urgency. The wrapped package suggests she’s carrying something important that she needs to deliver promptly, possibly before businesses fully close.
    \item Likely Setting: Answer: A quieter part of a city or town in the evening. \textit{Explanation:} Although the text describes a bustling scene, the empty streets in the image suggest a quieter time, likely after peak hours. The setting could be a commercial area winding down for the night, with shops and streetlights adding to the urban feel.
\end{enumerate}

\subsection{Task 3: Cross-Modal Translation}
\textbf{Task Description:}
You are provided with an abstract painting and a poem. The task is to determine if the poem could be a literary interpretation of the painting and explain the reasoning.

Image Provided:
An abstract painting featuring swirling colors of blue and green with splashes of bright yellow and subtle hints of white. The overall impression is of a turbulent sea under a sunny sky.

\begin{figure}[H]
    \centering
    \includegraphics[width=0.6\textwidth]{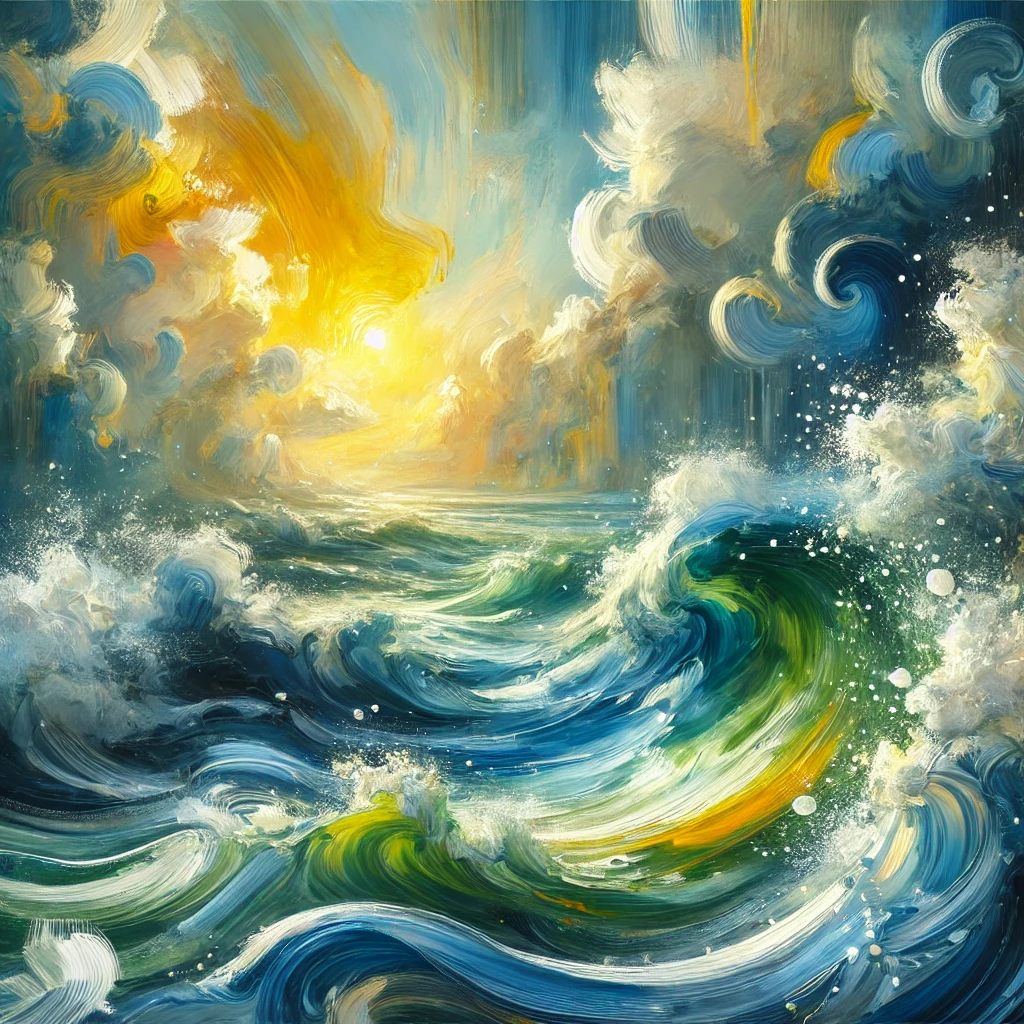}
    \caption{Input Figure for Task 4 for Evaluation Aspect 2}
    \label{fig:EA2_T4}
\end{figure}

Poem Provided:
\textit{"Whirls of azure embrace the golden gleam, Waves dance beneath the sun's radiant beam. Whispers of foam kiss the horizon's line, A symphony of hues in chaotic design."}

Question:
Assess whether the poem aligns with the painting's visual elements and themes. Provide a detailed explanation supporting your assessment.

\textbf{Expected Output:}\\
\textbf{Assessment: }Yes, the poem aligns with the painting's visual elements and themes.

Explanation:
\\
\textit{Colors and Imagery:}
The painting features blues and greens ("Whirls of azure"), representing the sea.
Bright yellow splashes correspond to the "golden gleam" and "sun's radiant beam" in the poem.
Hints of white could represent "whispers of foam." 
\\
\textit{Themes:}
Both the painting and the poem convey a sense of movement and turbulence ("swirling colors," "waves dance," "chaotic design").
The poem's reference to a "symphony of hues" reflects the painting's rich color palette.
\\
\textit{Overall Impression:}
The painting gives an impression of a turbulent sea under sunlight, which aligns with the poem's depiction of waves and interaction with the sun.
\\
Final Answer:
The poem is a literary interpretation of the painting, capturing its visual elements and themes through descriptive language.

\subsection{Task 4: Aligning Data from Charts and Text}
\textbf{Task Description:}\\
You are provided with a line chart and a paragraph describing economic trends. The task is to identify inconsistencies between the chart and the text.

Line Chart Provided:
A graph displaying the unemployment rate over five years (2016-2020):
\begin{itemize}
    \item 2016: 6\%
    \item 2017: 5\%
    \item 2018: 4\%
    \item 2019: 3\%
    \item 2020: 7\%
\end{itemize}

Text Provided:
\textit{"Over the past five years, the country has seen a consistent decline in unemployment rates, reaching an all-time low of 2\% in 2020. This steady improvement reflects the robust economic policies implemented since 2016."}

Question:
Identify and explain any inconsistencies between the information presented in the chart and the text.

\textbf{Expected Output:}\\
Inconsistency Identified:
\begin{itemize}
    \item The text states that unemployment rates have consistently declined over the past five years, reaching an all-time low of 2\% in 2020.
    \item The chart shows that in 2020, the unemployment rate actually increased to 7\%, not decreased to 2\%.
\end{itemize}
Explanation:
According to the chart:
\begin{itemize}
    \item Unemployment rates decreased from 6\% in 2016 to 3\% in 2019.
    \item In 2020, there was a significant increase to 7\%, possibly due to unforeseen circumstances (e.g., economic downturn, global events).
    \item The text contradicts the chart by claiming a decrease to 2\% in 2020, indicating either outdated information or an error in reporting.
\end{itemize}
Conclusion:
There is a clear discrepancy between the chart and the text regarding the unemployment rate in 2020. The chart shows an increase to 7\%, while the text incorrectly states a decrease to 2\%.


\section{Evaluation Aspect 3 (EA3): Complex Code Generation and Execution Tasks}
\label{supp_EA3}

\subsection{Task 1: Data Visualization from an Image of a Table}
\textbf{Task Description:}
You are provided with an image of a simple table containing the names of students and their corresponding test scores.
Image Provided:
An image displaying the following table:
\begin{table}[h]
    \centering
    \begin{tabular}{|c|c|}
        \hline
        \textbf{Name} & \textbf{Score} \\
        \hline
        Alice  & 85  \\
        Bob    & 90  \\
        Carol  & 78  \\
        David  & 92  \\
        Eve    & 88  \\
        \hline
    \end{tabular}
    \caption{Scores of Different Individuals}
    \label{tab:scores}
\end{table}

\begin{figure}[H]
    \centering
    \includegraphics[width=0.5\textwidth]{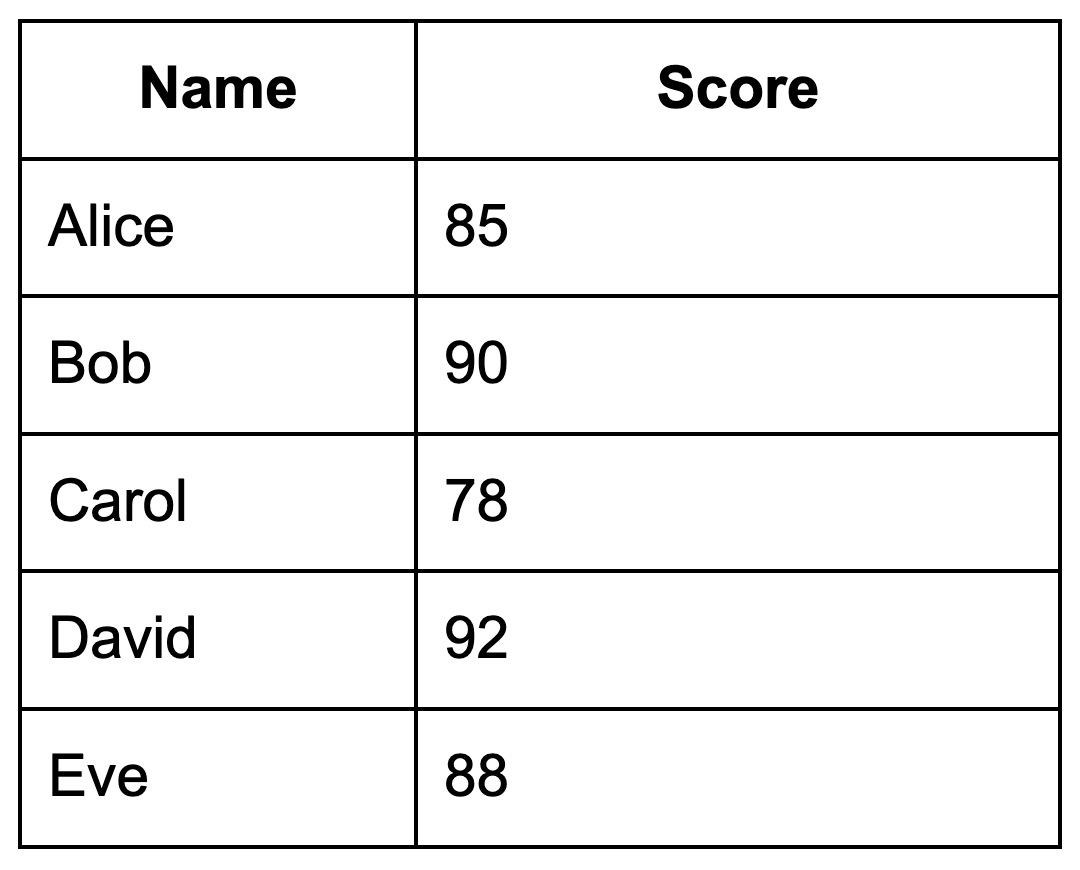}
    \caption{Input Figure for Task 1 for Evaluation Aspect 3}
    \label{fig:EA3_T1}
\end{figure}

Question:
Write a Python script that:
\begin{enumerate}
    \item Stores the data from the table into appropriate data structures.
    \item Plots a bar chart showing each student's name and their corresponding score.
\end{enumerate}

\textbf{Expected Output:}

\begin{lstlisting}[language=Python]
# Import necessary libraries
import matplotlib.pyplot as plt

# Store the data from the table
names = ['Alice', 'Bob', 'Carol', 'David', 'Eve']
scores = [85, 90, 78, 92, 88]

# Plot the bar chart
plt.bar(names, scores, color='skyblue')
plt.xlabel('Students')
plt.ylabel('Scores')
plt.title('Test Scores of Students')
plt.show()
\end{lstlisting}

\subsection{Task 2: Drawing a Shape Based on an Image}
\textbf{Task Description:}
Image Provided: An image depicting a simple house shape composed of a square (representing the house) and a triangle on top (representing the roof). 

\begin{figure}[H]
    \centering
    \includegraphics[width=0.6\textwidth]{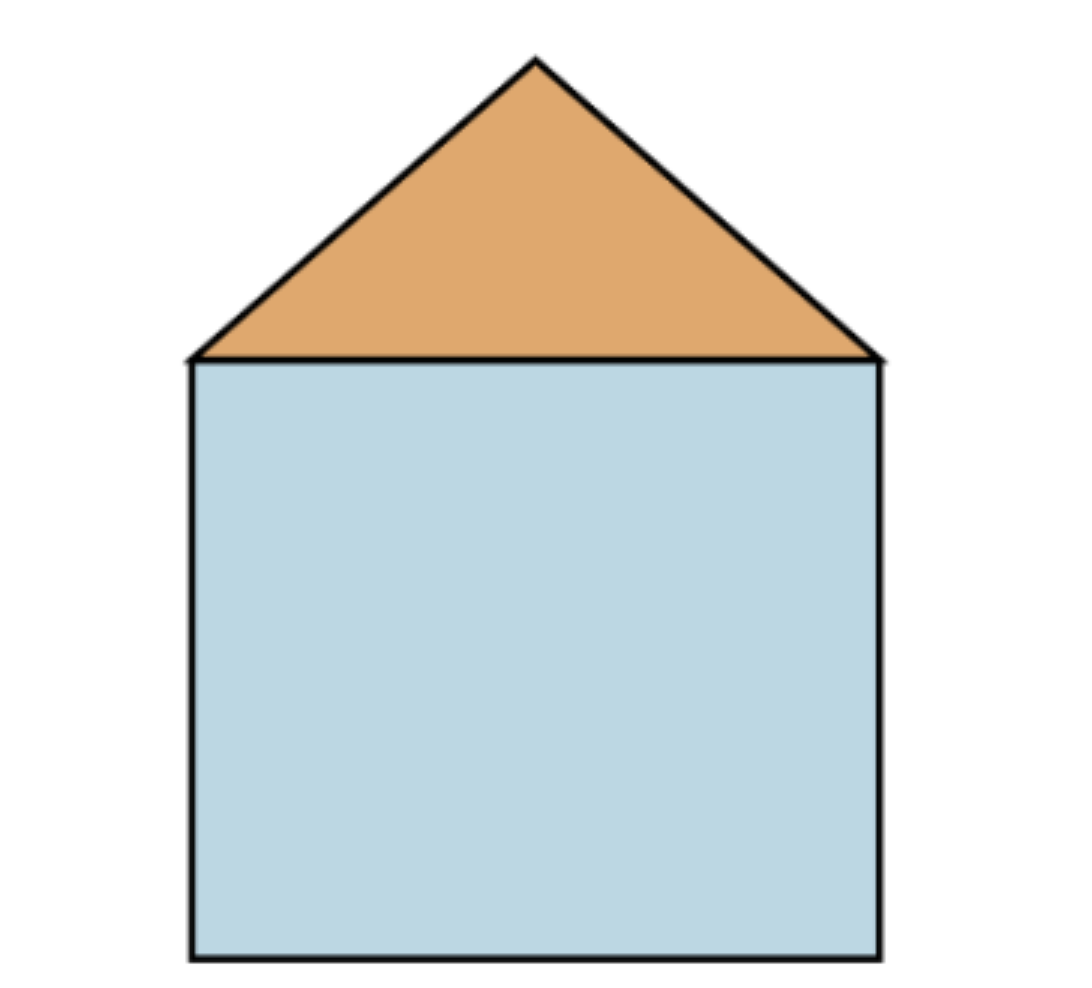}
    \caption{Input Figure for Task 2 for Evaluation Aspect 3}
    \label{fig:EA3_T2}
\end{figure}

Question:
Write a Python script using the turtle module that draws the house as shown in the image.

\textbf{Expected Output:}
\begin{lstlisting}[language=Python]
import turtle

t = turtle.Turtle()

# Draw the square (house base)
for _ in range(4):
    t.forward(100)
    t.left(90)

# Position for the roof
t.left(90)
t.forward(100)
t.right(90)

# Draw the triangle (roof)
t.left(45)
t.forward(70)
t.right(90)
t.forward(70)
t.right(135)
t.forward(100)

turtle.done()
\end{lstlisting}

\subsection{Task 3: Calculating a Sum from Text in an Image }
\textbf{Task Description:}
Image Provided: You are given an image containing the following mathematical instruction:
\textit{"Calculate the sum of all even numbers from 1 to 10."}

\begin{figure}[H]
    \centering
    \includegraphics[width=0.7\textwidth]{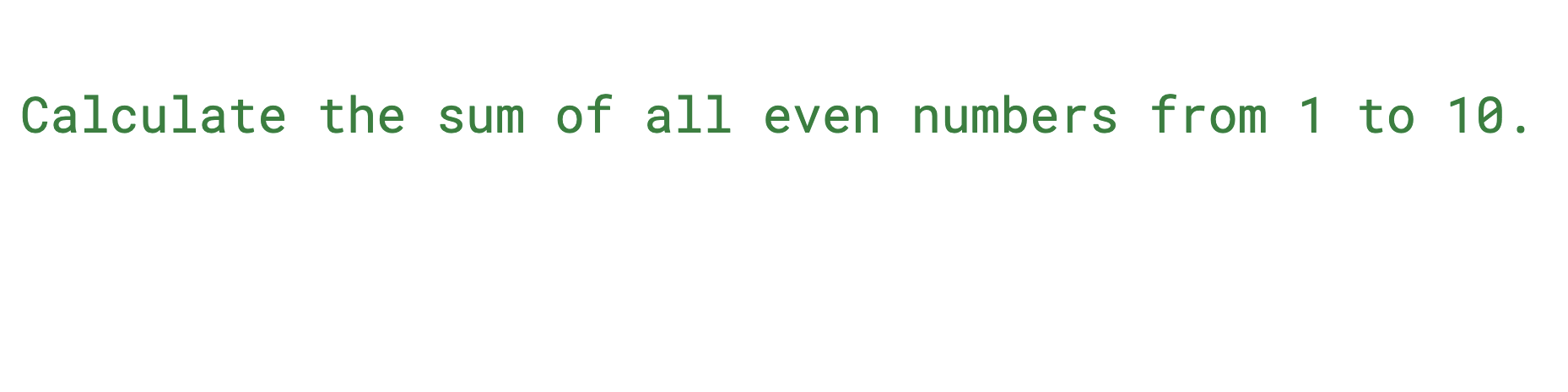}
    \caption{Input Figure for Task 3 for Evaluation Aspect 3}
    \label{fig:EA3_T3}
\end{figure}

Question:
Write a Python script that calculates and prints the sum as instructed in the image.

\textbf{Expected Output:}
\begin{lstlisting}[language=Python]
# Calculate the sum of all even numbers from 1 to 10
total = sum(num for num in range(1, 11) if num % 2 == 0)
print("The sum of all even numbers from 1 to 10 is:", total)
\end{lstlisting}

\subsection{Task 4: Creating a Dictionary from an Image of a Chart}
\textbf{Task Description:}
You are provided with an image of a simple bar chart showing the number of units sold for three products: Product A, Product B, and Product C.
A bar chart illustrating:
\begin{itemize}
    \item Product A: 50 units sold
    \item Product B: 70 units sold
    \item Product C: 40 units sold
\end{itemize}

\begin{figure}[H]
    \centering
    \includegraphics[width=0.6\textwidth]{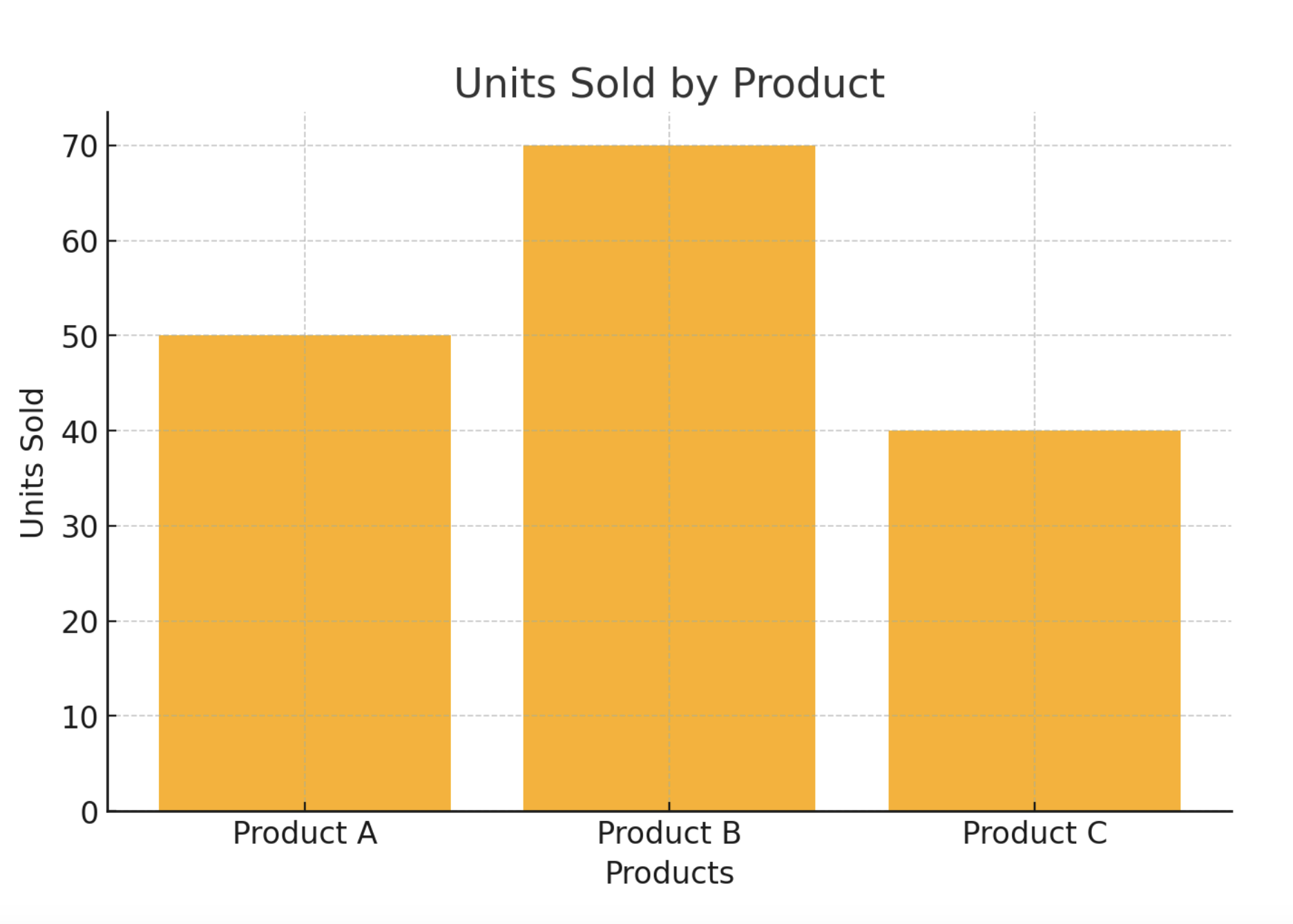}
    \caption{Input Figure for Task 4 for Evaluation Aspect 3}
    \label{fig:EA3_T4}
\end{figure}

Question:
Write a Python script that:
\begin{enumerate}
    \item Creates a dictionary with product names as keys and units sold as values.
    \item Prints the dictionary.
\end{enumerate}

\textbf{Expected Output:}
\begin{lstlisting} [language=Python]
    # Create the dictionary with sales data
sales = {
    'Product A': 50,
    'Product B': 70,
    'Product C': 40
}

# Print the dictionary
print("Sales Data:", sales)
\end{lstlisting}

\subsection{Task 5: Summing Prices from a Shopping List Image }
\textbf{Task Description:}
You are given an image containing a shopping list with items and their prices.

Image Provided: An image displaying the following list:
\begin{itemize}
    \item Apples: \$2
    \item Bananas: \$1
    \item Oranges: \$3
    \item Grapes: \$4
\end{itemize}

\begin{figure}[H]
    \centering
    \includegraphics[width=0.6\textwidth]{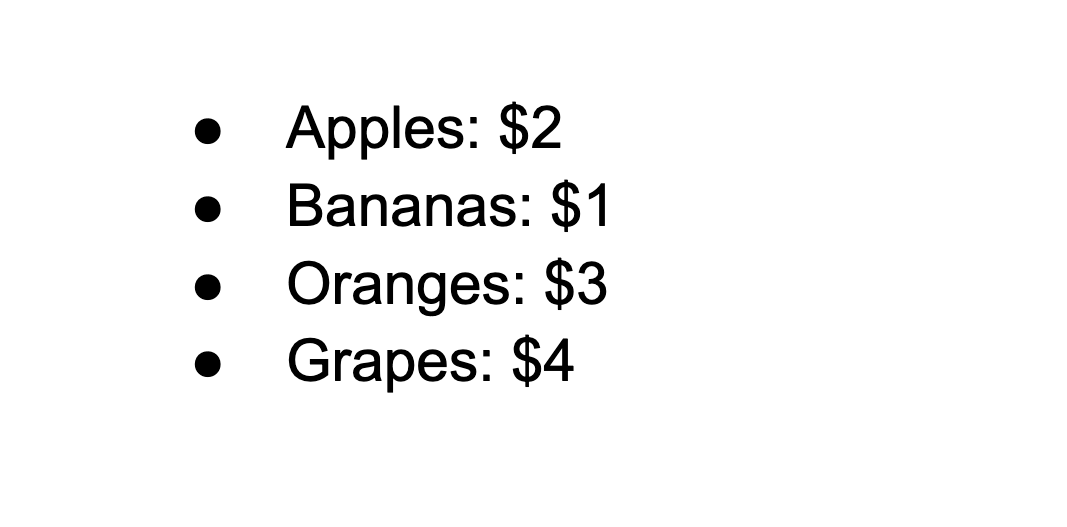}
    \caption{Input Figure for Task 5 for Evaluation Aspect 3}
    \label{fig:EA3_T5}
\end{figure}

Question:
Write a Python script that:
\begin{itemize}
    \item Stores the items and their prices in a dictionary.
    \item Calculates and prints the total cost of all items.
\end{itemize}

\textbf{Expected Output:}
\begin{lstlisting} [language=Python]
# Store items and their prices in a dictionary
shopping_list = {
    'Apples': 2,
    'Bananas': 1,
    'Oranges': 3,
    'Grapes': 4
}

# Calculate the total cost
total_cost = sum(shopping_list.values())

# Print the total cost
print("Total cost of all items is: $", total_cost)
\end{lstlisting}

\subsection{Task 6: Parsing a Simple CSV Structure from an Image }
\textbf{Task Description:}
You are provided with an image showing a CSV-like structure listing employees and their departments.

Image Provided: An image displaying the following data:
\begin{lstlisting}
Name, Department
John Doe, Sales
Jane Smith, Marketing
Alice Johnson, Development
Bob Brown, HR
\end{lstlisting}

\begin{figure}[H]
    \centering
    \includegraphics[width=0.6\textwidth]{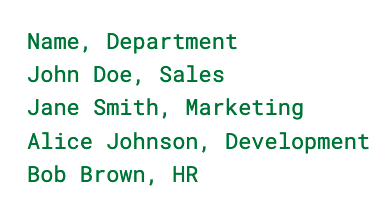}
    \caption{Input Figure for Task 6 for Evaluation Aspect 3}
    \label{fig:EA3_T6}
\end{figure}

Question:
Write a Python script that:
\begin{enumerate}
    \item Parses the data and stores it in a list of dictionaries.
    \item Prints the list.
\end{enumerate}

\textbf{Expected Output:}
\begin{lstlisting}[language=Python]
    # Parse the data
data = [
    {'Name': 'John Doe', 'Department': 'Sales'},
    {'Name': 'Jane Smith', 'Department': 'Marketing'},
    {'Name': 'Alice Johnson', 'Department': 'Development'},
    {'Name': 'Bob Brown', 'Department': 'HR'}
]

# Print the list of dictionaries
print("Employee Data:")
for entry in data:
    print(entry)
\end{lstlisting}

\subsection{Task 7: Generating Fibonacci Sequence Based on Image Instruction}
\textbf{Task Description:}
Provided Image: An image contains the following instruction:
\textit{"Write a program to generate the first 10 numbers of the Fibonacci sequence."}

\begin{figure}[H]
    \centering
    \includegraphics[width=0.6\textwidth]{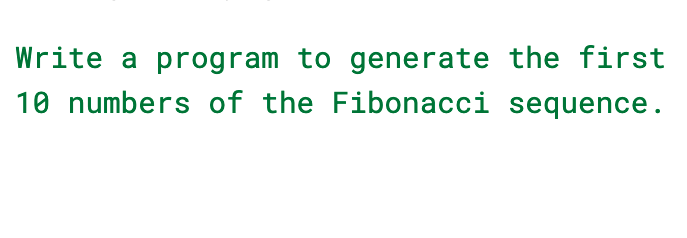}
    \caption{Input Figure for Task 7 for Evaluation Aspect 3}
    \label{fig:EA3_T7}
\end{figure}

Question:
Write a Python script that fulfills the instruction provided in the image.

\textbf{Expected Output:}
\begin{lstlisting}[language=Python]
# Initialize the first two numbers of the Fibonacci sequence
fib_sequence = [0, 1]

# Generate the next 8 numbers
for i in range(2, 10):
    next_number = fib_sequence[i-1] + fib_sequence[i-2]
    fib_sequence.append(next_number)

# Print the first 10 numbers of the Fibonacci sequence
print("First 10 numbers of the Fibonacci sequence:")
print(fib_sequence)   
\end{lstlisting}

\subsection{Task 8: Responding to a Flowchart Image }
\textbf{Task Description:}
You are given an image of a simple flowchart that outlines the steps of a decision-making process for checking if a number is prime.

Image Provided:
A flowchart with the following steps:
\begin{lstlisting}
Start
Input a number n
If n <= 1, print "Not prime" and end
For i from 2 to n - 1:
If n mod i == 0, print "Not prime" and end
Print "Prime"
End  
\end{lstlisting}

\begin{figure}[H]
    \centering
    \includegraphics[width=0.6\textwidth]{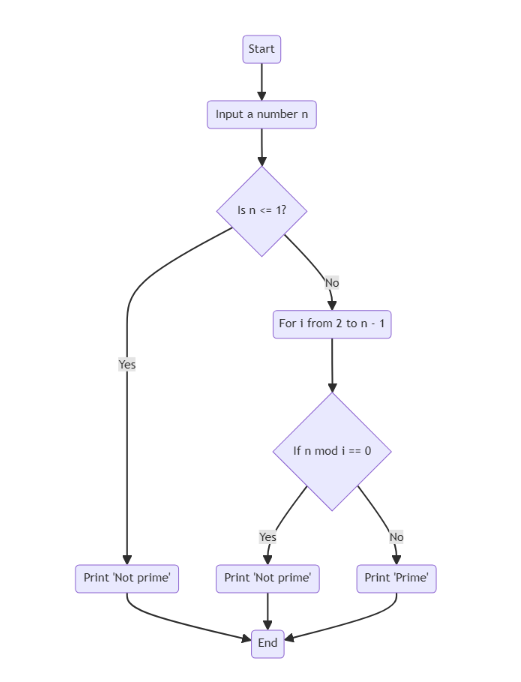}
    \caption{Input Figure for Task 8 for Evaluation Aspect 3}
    \label{fig:EA3_T8}
\end{figure}

Question:
Write a Python function $is$\textunderscore$prime(n)$ that implements the logic from the flowchart and returns $True$ if $n$ is a prime number $n$ and $False$ otherwise.

\textbf{Expected Output:}
\begin{lstlisting}[language=Python]
def is_prime(n):
    if n <= 1:
        return False
    for i in range(2, n):
        if n % i == 0:
            return False
    return True

# Example usage
number = int(input("Enter a number: "))
if is_prime(number):
    print(number, "is a prime number.")
else:
    print(number, "is not a prime number.")    
\end{lstlisting}


\section{Evaluation Aspect 4 (EA4): Knowledge Retrieval and Integration Tasks}
\label{supp_EA4}

\subsection{Task 1: Historical Monument Identification and Explanation}
\textbf{Task Description:}
You are provided with an image of a historical monument and asked to answer questions that require integrating visual information with external knowledge.

Image Provided: An image of the Eiffel Tower in Paris, France.
\begin{figure}[H]
    \centering
    \includegraphics[width=0.6\textwidth]{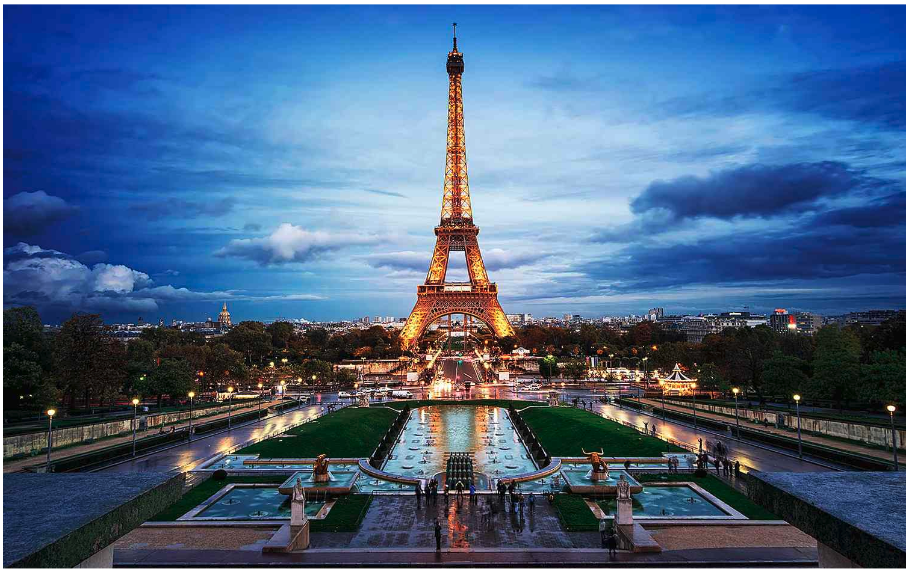}
    \caption{Input Figure for Task 1 for Evaluation Aspect 4}
    \label{fig:EA4_T1}
\end{figure}

Question:
\begin{enumerate}
    \item Identify the monument shown in the image.
    \item Provide a brief history of this monument, including the year it was completed and its original purpose.
    \item Explain why it has become a significant cultural symbol.
\end{enumerate}

\textbf{Expected Output:}
Identification: The monument is the Eiffel Tower.
Brief History: Year Completed: 1889.
Original Purpose: Built as the entrance arch for the 1889 Exposition Universelle (World's Fair) to celebrate the 100th anniversary of the French Revolution.
Designer: Gustave Eiffel and his engineering company.
Cultural Significance:
Initially met with criticism from some artists and intellectuals.
Over time, it became a global icon of France and an enduring symbol of Paris.
Represents architectural innovation and industrial advancement.
Attracts millions of visitors annually, contributing to tourism and cultural heritage.

\subsection{Task 2: Scientific Data Interpretation from Graph and Text}
\textbf{Task Description:}
You are provided with a line graph showing global average temperatures over the past century and a short paragraph discussing climate change.
\\
Graph Provided:
A line graph displaying global average temperatures from 1900 to 2000.
The graph shows a gradual increase in temperature, with a more pronounced rise in the latter half of the century.
\begin{figure}[H]
    \centering
    \includegraphics[width=0.6\textwidth]{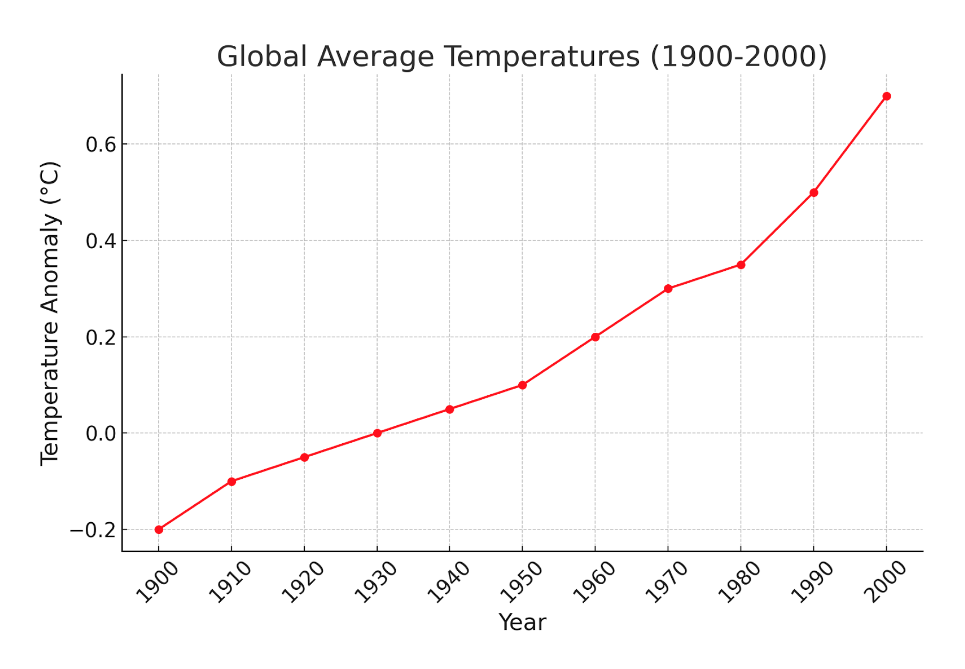}
    \caption{Input Figure for Task 2 for Evaluation Aspect 4}
    \label{fig:EA4_T2}
\end{figure}

Text Provided:
\textit{"Recent studies indicate a significant trend in global warming, particularly in the last 50 years. Scientists attribute this rise to increased greenhouse gas emissions from human activities like burning fossil fuels and deforestation."}

Question:
\begin{enumerate}
    \item Based on the graph, calculate the approximate increase in global average temperature from 1900 to 2000.
    \item Summarize how the data in the graph supports the information provided in the text.
    \item Discuss potential implications of this trend on global ecosystems.
\end{enumerate}

\textbf{Expected Output:}
Approximate Temperature Increase:
\begin{itemize}
    \item 1900 Temperature: Approximately 13.7°C.
    \item 2000 Temperature: Approximately 14.4°C.
    \item Increase: 14.4°C - 13.7°C = 0.7°C.
\end{itemize}
Data Supporting the Text:
The graph shows a steady rise in global temperatures, aligning with the text's mention of global warming.
The sharper increase in the last 50 years corresponds with industrialization and higher greenhouse gas emissions.
Visual evidence from the graph reinforces the scientific studies referenced in the text.
\\
Potential Implications:
\begin{itemize}
    \item Rising Sea Levels: Melting polar ice caps leading to coastal flooding.
    \item Extreme Weather Events: Increased frequency of hurricanes, droughts, and heatwaves.
    \item Ecosystem Disruption: Loss of biodiversity as species struggle to adapt.
    \item Agricultural Impact: Changes in crop viability due to shifting climate zones.
\end{itemize}

\subsection{Task 3: Medical Image Analysis with Knowledge Integration}
\textbf{Task Description:}
You are provided with a chest X-ray image and asked to diagnose a potential medical condition for a lung, integrating visual analysis with medical knowledge.

Image Provided: An X-ray showing a noticeable opacity in the upper lobe of the right lung.
\begin{figure}[H]
    \centering
    \includegraphics[width=0.6\textwidth]{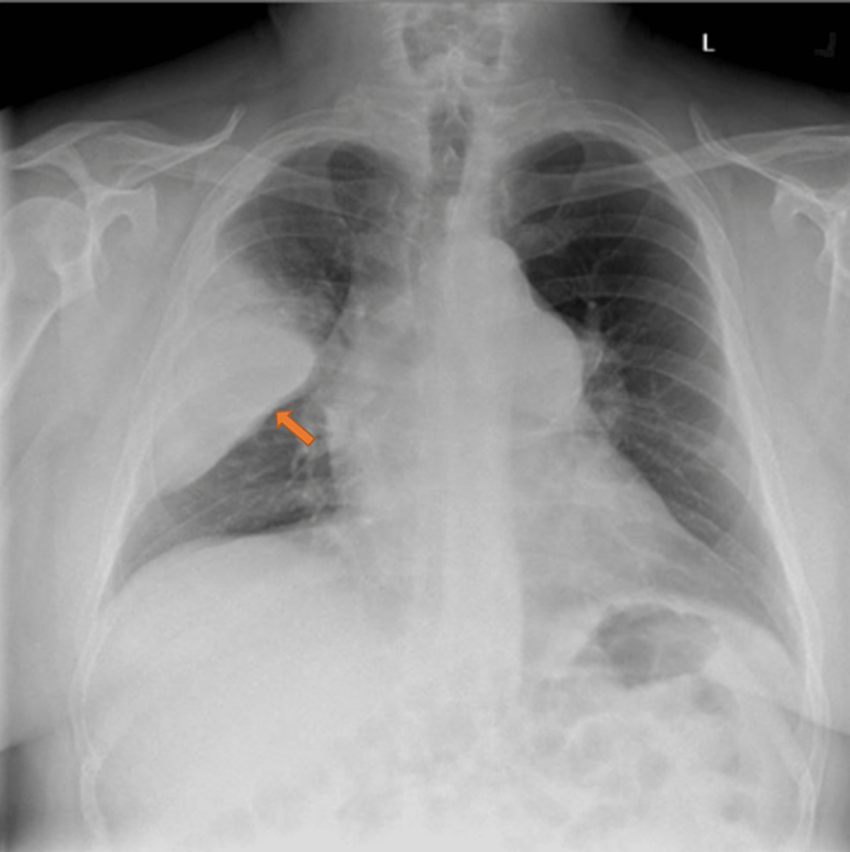}
    \caption{Input Figure for Task 3 for Evaluation Aspect 4}
    \label{fig:EA4_T3}
\end{figure}

Question:
\begin{enumerate}
    \item Identify any abnormalities present in the X-ray.
    \item Suggest possible medical conditions associated with the observed abnormality.
    \item Recommend further tests or procedures to confirm the diagnosis.
\end{enumerate}

\textbf{Expected Output:}
Identified Abnormality:
Presence of an opacity (shadow) in the upper lobe of the right lung.
\\
Possible Medical Conditions:
\begin{enumerate}
    \item Pulmonary Tuberculosis (TB): Commonly affects upper lobes.
    \item Lung Cancer: A mass or nodule could indicate a tumor.
    \item Pneumonia: Infection causing localized consolidation.
    \item Pulmonary Fibrosis: Scarring leading to opacity.
\end{enumerate}
Recommended Further Tests:
\begin{enumerate}
    \item CT Scan of the Chest: Provides detailed imaging to assess the lesion.
    \item Sputum Analysis: Checks for infectious agents like TB bacteria.
    \item Biopsy: Obtaining tissue samples for histopathological examination.
    \item Blood Tests: Inflammatory markers, tumor markers, complete blood count.
\end{enumerate}

\subsection{Task 4: Cultural Artifact Interpretation}
\textbf{Task Description:}
You are given an image of an ancient artifact along with some background information.
\\
Image Provided:
A photograph of the Rosetta Stone, a slab inscribed with text in three scripts.
\begin{figure}[H]
    \centering
    \includegraphics[width=0.6\textwidth]{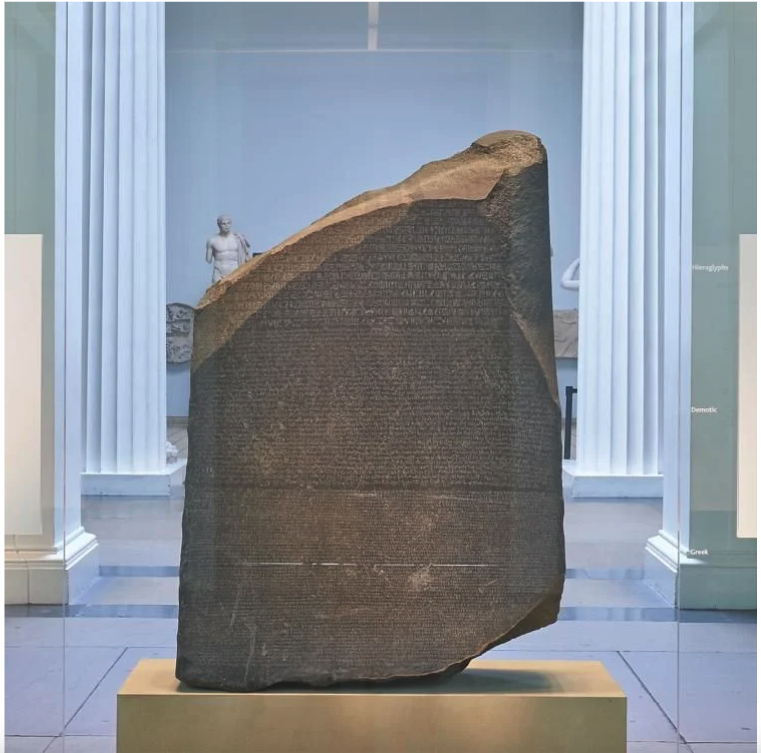}
    \caption{Input Figure for Task 4 for Evaluation Aspect 4}
    \label{fig:EA4_T4}
\end{figure}

Background Information:
The artifact was discovered in 1799 and has been crucial in understanding ancient languages.
\\
Questions:
\begin{enumerate}
    \item Describe the significance of the artifact shown in the image.
    \item Explain how it contributed to the field of linguistics and the study of ancient civilizations.
    \item Identify the languages or scripts present on the artifact.
\end{enumerate}

\textbf{Expected Output:}
Significance of the Artifact:
The Rosetta Stone is significant because it was key to deciphering Egyptian hieroglyphs.\\
Contribution to Linguistics and Ancient Studies:
\begin{itemize}
    \item Provided a bilingual (actually trilingual) inscription that enabled scholars to compare hieroglyphs with known languages.
    \item Allowed Jean-François Champollion to decode hieroglyphs in 1822.
    \item Opened up vast knowledge about ancient Egyptian history, culture, and language.
\end{itemize}
Languages or Scripts Present:
\begin{itemize}
    \item Hieroglyphic Script: Used for important or religious documents.
    \item Demotic Script: Common script for daily purposes in ancient Egypt.
    \item Ancient Greek: The administrative language at the time; Greek was well-understood by scholars.
\end{itemize}

\subsection{Task 5: Integrating Knowledge from a Map and Text Description}
\textbf{Task Description:}
You are provided with a map highlighting key locations during World War II and a text describing a historical event.

Map Provided:
A map of Europe showing locations such as Normandy, Berlin, and Moscow.
\begin{figure}[H]
    \centering
    \includegraphics[width=0.6\textwidth]{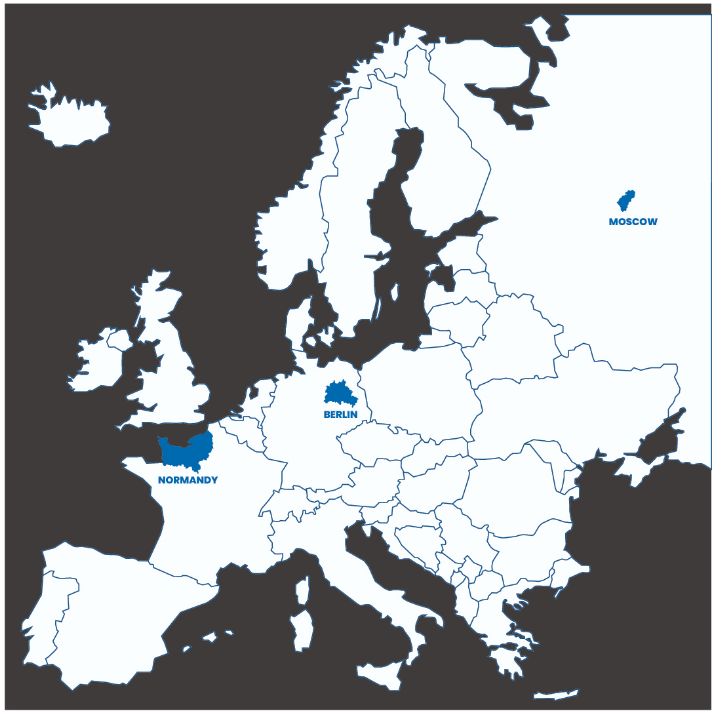}
    \caption{Input Figure for Task 5 for Evaluation Aspect 4}
    \label{fig:EA4_T5}
\end{figure}

Text Provided:
\textit{"Operation Overlord was a pivotal event during World War II, marking the start of the Allied invasion of German-occupied Western Europe. The operation commenced on June 6, 1944, with landings on the Normandy beaches."}

Questions:
\begin{enumerate}
    \item Locate the area where Operation Overlord took place using the map.
    \item Explain the strategic importance of this operation in the context of World War II.
    \item Discuss the outcome and its impact on the war's progression.
\end{enumerate}

\textbf{Expected Output:}
Location:
Operation Overlord took place in Normandy, on the northern coast of France.

Strategic Importance:
\begin{itemize}
    \item Opened a Western Front against Nazi Germany.
    \item Forced Germany to divert resources from the Eastern Front against the Soviet Union.
    \item Enabled the liberation of Western European countries from Nazi control.
\end{itemize}
Outcome and Impact:
\begin{itemize}
    \item Successful establishment of a beachhead by Allied forces.
    \item Led to the liberation of Paris and eventual defeat of Nazi Germany.
    \item Accelerated the end of the war in Europe, culminating in Germany's surrender in May 1945.
\end{itemize}

\subsection{Task 6: Integrating Information from a Chart and Article }
\textbf{Task Description:}
You are given a pie chart showing global energy consumption by source and an article discussing renewable energy trends. \\

Pie Chart Provided:
Distribution of global energy consumption:
\begin{itemize}
    \item Oil: 33\%
    \item Coal: 27\%
    \item Natural Gas: 24\%
    \item Renewables: 10\%
    \item Nuclear: 6\%
\end{itemize}

\begin{figure}[H]
    \centering
    \includegraphics[width=0.6\textwidth]{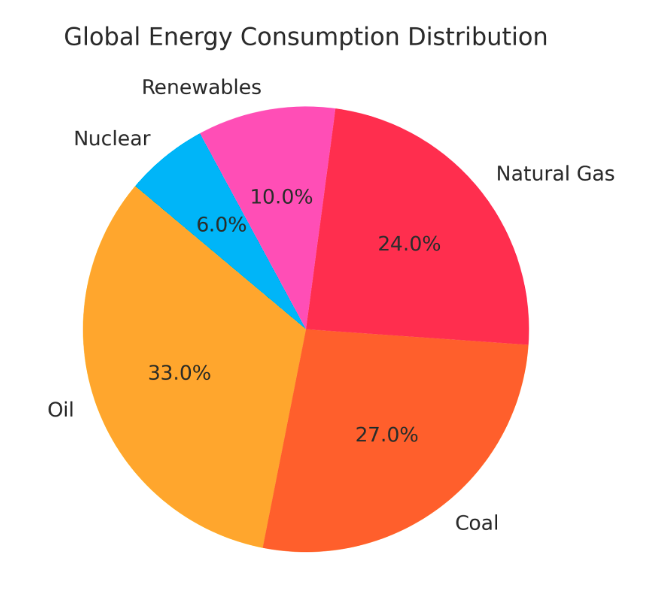}
    \caption{Input Figure for Task 6 for Evaluation Aspect 4}
    \label{fig:EA4_T6}
\end{figure}

Article Excerpt:
\textit{"Despite significant investments, renewable energy sources account for a small portion of global energy consumption. However, the shift towards renewables is accelerating due to environmental concerns and technological advancements."}

Question:
\begin{enumerate}
    \item Calculate the total percentage of energy consumption from non-renewable sources based on the chart.
    \item Summarize the main points of the article regarding renewable energy trends.
    \item Discuss challenges and benefits associated with increasing renewable energy usage.
\end{enumerate}

\textbf{Expected Output:}
Percentage from Non-Renewable Sources:
Non-renewable sources: Oil (33\%) + Coal (27\%) + Natural Gas (24\%) + Nuclear (6\%) = 90\%.

Article Summary:
Renewables currently make up a small share (10\%) of energy consumption.
Investments in renewables are growing.
Environmental concerns and technology are driving a shift towards renewable energy.
Challenges and Benefits: \\

Challenges:
\begin{itemize}
    \item High initial costs for infrastructure.
    \item Intermittency issues (e.g., solar and wind depend on weather conditions).
    \item Need for improved energy storage solutions.
    \item Transitioning from established fossil fuel industries.
\end{itemize}

Benefits:
\begin{itemize}
    \item Reduces greenhouse gas emissions and combats climate change.
    \item Provides sustainable, inexhaustible energy sources.
    \item Enhances energy security by diversifying supply.
    \item Stimulates economic growth through new industries and job creation.
\end{itemize}

\subsection{Task 7: Multimodal Fact Checking}
\textbf{Task Description:}
You are provided with an image of a newspaper headline and an excerpt from a reputable online encyclopedia.

Image Provided:
A photograph of a newspaper with the headline:
\textit{"Discovery of a New Planet: Astronomers Find Planet X Beyond Pluto!"}

\begin{figure}[H]
    \centering
    \includegraphics[width=0.6\textwidth]{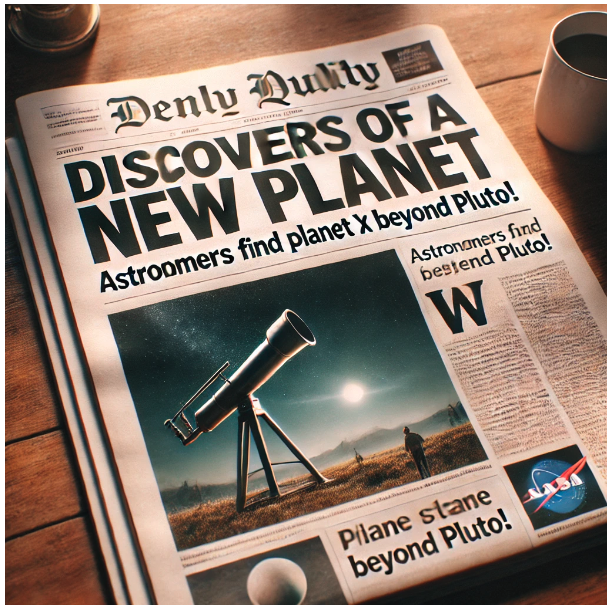}
    \caption{Input Figure for Task 7 for Evaluation Aspect 4}
    \label{fig:EA4_T7}
\end{figure}

Encyclopedia Excerpt:
\textit{"As of now, there are eight recognized planets in the Solar System. Pluto was reclassified as a dwarf planet in 2006. While there have been hypotheses about a 'Planet Nine' or 'Planet X', no such planet has been confirmed."}
\\
Question:
\begin{enumerate}
    \item Assess the accuracy of the newspaper headline based on the encyclopedia excerpt.
    \item Explain any discrepancies between the two sources.
    \item Provide a reasoned conclusion about the existence of "Planet X."
\end{enumerate}

\textbf{Expected Output:}
Assessment of Accuracy:
The newspaper headline claims the discovery of a new planet, "Planet X," beyond Pluto.
According to the encyclopedia, no such planet has been confirmed. \\

Discrepancies:
\begin{itemize}
    \item The newspaper reports a confirmed discovery, whereas the encyclopedia mentions only hypotheses without confirmation.
    \item Possible that the newspaper is reporting speculative or unverified information.
\end{itemize}

Reasoned Conclusion:
\begin{itemize}
    \item Based on current reputable sources, "Planet X" has not been officially discovered.
    \item The headline may be sensationalized or based on preliminary findings not yet validated by the scientific community.
    \item Until confirmed by multiple observations and peer-reviewed studies, the existence of "Planet X" remains unproven.
\end{itemize}

\subsection{Task 8: Integrating Visual Art and Historical Context}
\textbf{Task Description:}
You are given an image of a famous painting and asked to analyze it in the context of its historical background.
\\
Image Provided: A picture of Vincent van Gogh's "Starry Night."

\begin{figure}[H]
    \centering
    \includegraphics[width=0.6\textwidth]{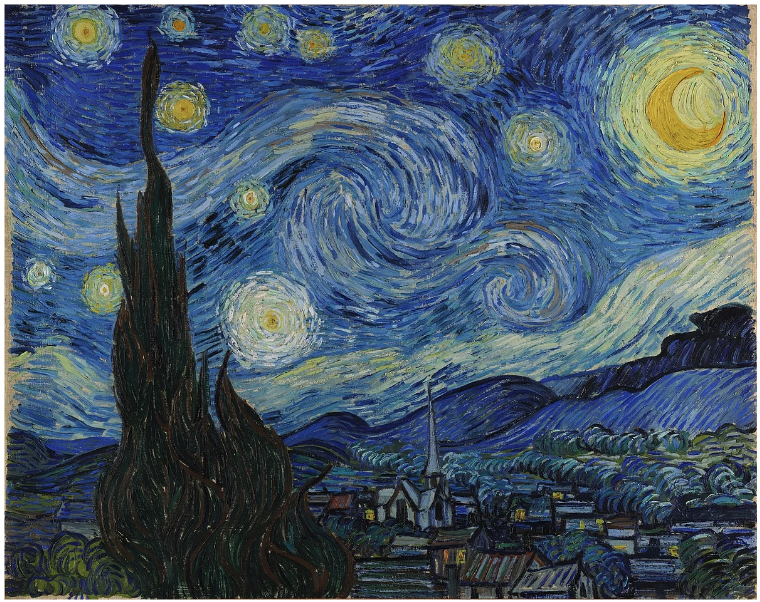}
    \caption{Input Figure for Task 8 for Evaluation Aspect 4}
    \label{fig:EA4_T8}
\end{figure}

Question:
\begin{enumerate}
    \item Identify the painting and its artist.
    \item Discuss the historical and personal context in which it was created.
    \item Explain how the painting reflects the characteristics of the art movement it is associated with.
\end{enumerate}

\textbf{Expected Output:}
Identification:
The painting is "Starry Night" by Vincent van Gogh.\\

Historical and Personal Context:
\begin{itemize}
    \item Created in 1889 while van Gogh was in the Saint-Paul-de-Mausole asylum in Saint-Rémy-de-Provence, France.
    \item Reflects his emotional turmoil and struggles with mental health.
    \item Painted from memory during the day, depicting the view from his asylum room at night.
\end{itemize}
Art Movement Characteristics: 
Associated with Post-Impressionism. \\

Characteristics Reflected:
\begin{itemize}
    \item Use of bold colors and expressive, swirling brushstrokes.
    \item Emphasis on emotional and psychological content over realistic representation.
    \item Depicts the artist's inner feelings and subjective experience of the world.
\end{itemize}

\end{document}